\documentclass{article} % For LaTeX2e
\usepackage{iclr2023_conference,times}

\usepackage{amsmath,amsfonts,bm}

\def\figref#1{Fig.~\ref{#1}}
\def\Figref#1{Fig.~\ref{#1}}

\def\eqref#1{Eq.~\ref{#1}}
\def\Eqref#1{Eq.~\ref{#1}}
\def\algref#1{algorithm~\ref{#1}}
\def\1{\bm{1}}

\def\ru{{\textnormal{u}}}

\def\rx{{\textnormal{x}}}

\def\rvd{{\mathbf{d}}}

\def\rvu{{\mathbf{i}}}

\def\rvm{{\mathbf{m}}}

\def\rvp{{\mathbf{p}}}

\def\rvs{{\mathbf{s}}}

\def\rvu{{\mathbf{u}}}
\def\rvv{{\mathbf{v}}}
\def\rvw{{\mathbf{w}}}
\def\rvx{{\mathbf{x}}}
\def\rvy{{\mathbf{y}}}
\def\rvz{{\mathbf{z}}}

\def\rmI{{\mathbf{I}}}

\def\vtheta{{\bm{\theta}}}

\def\vu{{\bm{u}}}
\def\vv{{\bm{v}}}
\def\vw{{\bm{w}}}
\def\vx{{\bm{x}}}

\def\vz{{\bm{z}}}

\def\mA{{\bm{A}}}

\def\mC{{\bm{C}}}
\def\mD{{\bm{D}}}

\def\mF{{\bm{F}}}

\def\mH{{\bm{H}}}
\def\mI{{\bm{I}}}

\def\mP{{\bm{P}}}

\def\mR{{\bm{R}}}
\def\mS{{\bm{S}}}

\def\mV{{\bm{V}}}
\def\mW{{\bm{W}}}

\def\mLambda{{\bm{\Lambda}}}

\DeclareMathAlphabet{\mathsfit}{\encodingdefault}{\sfdefault}{m}{sl}
\SetMathAlphabet{\mathsfit}{bold}{\encodingdefault}{\sfdefault}{bx}{n}

\newcommand{\E}{\mathbb{E}}

\newcommand{\KL}{D_{\mathrm{KL}}}
\newcommand{\Var}{\mathrm{Var}}

\DeclareMathOperator*{\argmax}{arg\,max}
\DeclareMathOperator*{\argmin}{arg\,min}

\DeclareMathOperator{\sign}{sign}

\newtheorem{thm}{Theorem}
\newtheorem{mydef}{Definition}

\def\iid{i.i.d.~}
\def\eg{\textit{e.g.}}
\def\ie{\textit{i.e.}}

\usepackage{graphicx}
\usepackage{subfiles}
\usepackage[ruled,vlined,linesnumbered]{algorithm2e}
\usepackage{mathtools}
\usepackage{array, multirow}
\usepackage{xcolor}
\usepackage{colortbl}
\usepackage{wrapfig}
\usepackage{amssymb}
\usepackage{pifont}

\newcommand{\xmark}{\ding{55}}
\definecolor{mypink1}{rgb}{0.858, 0.188, 0.478}

\definecolor{baselinecolor}{gray}{.9}
\newcommand{\best}[1]{\cellcolor{baselinecolor}{#1}}

\def\appref#1{App.~\ref{#1}}
\def \tabref#1{Table~\ref{#1}}

\usepackage[utf8]{inputenc} % allow utf-8 input
\usepackage[T1]{fontenc}    % use 8-bit T1 fonts
\usepackage{booktabs}       % professional-quality tables
\usepackage{nicefrac}       % compact symbols for 1/2, etc.
\usepackage{microtype}      % microtypography

\usepackage{hyperref}
\usepackage{url}

\title{Differentiable Gaussianization Layers for \\Inverse Problems Regularized by Deep Generative Models}

\author{Dongzhuo Li \\
ExxonMobil Technology \& Engineering Company\\
\texttt{dongzhuo.li@exxonmobil.com}
}

\iclrfinalcopy % Uncomment for camera-ready version, but NOT for submission.
\begin{document}

\maketitle

\begin{abstract}
Deep generative models such as GANs, normalizing flows, and diffusion models are powerful regularizers for inverse problems. They exhibit great potential for helping reduce ill-posedness and attain high-quality results. However, the latent tensors of such deep generative models can fall out of the desired high-dimensional standard Gaussian distribution during inversion, particularly in the presence of data noise and inaccurate forward models, leading to low-fidelity solutions. To address this issue, we propose to reparameterize and Gaussianize the latent tensors using novel differentiable data-dependent layers wherein custom operators are defined by solving optimization problems. These proposed layers constrain inverse problems to obtain high-fidelity in-distribution solutions. We validate our technique on three inversion tasks: compressive-sensing MRI, image deblurring, and eikonal tomography (a nonlinear PDE-constrained inverse problem) using two representative deep generative models: StyleGAN2 and Glow. Our approach achieves state-of-the-art performance in terms of accuracy and consistency.
\end{abstract}

\section{Introduction}
Inverse problems play a crucial role in many scientific fields and everyday applications. For example, astrophysicists use radio electromagnetic data to image galaxies and black holes~\citep{hogbom1974aperture,akiyama2019first}. Geoscientists rely on seismic recordings to reveal the internal structures of Earth~\citep{tarantola1984inversion,tromp2005seismic,virieux2009overview}. Biomedical engineers and doctors use X-ray projections, ultrasound measurements, and magnetic resonance data to reconstruct images of human tissues and organs~\citep{lauterbur1973image,gemmeke20073d,lustig2007sparse}. Therefore, developing effective solutions for inverse problems is of great importance in advancing scientific endeavors and improving our daily lives.

Solving an inverse problem starts with the definition of a forward mapping from parameters \(\rvm\) to data \(\rvd\), which we formally write as
\begin{equation}
	\rvd = f(\rvm) + \bm{\epsilon},
\end{equation}
where \(f\) stands for a forward model that usually describes some physical process, \(\bm{\epsilon}\) denotes noise, \(\rvd\) the observed data, and \(\rvm\) the parameters to be estimated. The forward model can be either linear or nonlinear and either explicit or implicitly defined by solving partial differential equations (PDEs). This study considers three representative inverse problems: \textbf{Compressive Sensing MRI}, \textbf{Deblurring}, and \textbf{Eikonal (traveltime) Tomography}, which have important applications in medical science, geoscience, and astronomy. The details of each problem and its forward model are in \appref{append:background_forward_models}.

The forward problem maps \(\rvm\) to \(\rvd\), while the inverse problem estimates \(\rvm\) given \(\rvd\). Unfortunately, inverse problems are generally under-determined with infinitely many compatible solutions and intrinsically ill-posed because of the nature of the physical system. Worse still, the observed data are usually noisy, and the assumed forward model might be inaccurate, exacerbating the ill-posedness. These challenges require using regularization to inject \textit{a priori} knowledge into inversion processes to obtain plausible and high-fidelity results. Therefore, an inverse problem is usually posed as an optimization problem:
\begin{equation}
  \argmin_{\rvm} ~(1/2) \left\|\rvd - f \left(\rvm\right) \right\|^2_2 + \mathcal{R}(\rvm),
  \label{eqn:inv_reg}
\end{equation}
where \(\mathcal{R}(\rvm)\) is the regularization term. Beyond traditional regularization methods such as the Tikhonov regularization and Total Variation (TV) regularization, deep generative models (DGM), such as VAEs~\citep{kingma2013auto}, GANs~\citep{goodfellow2014generative}, and normalizing flows~\citep{dinh2014nice,dinh2016density,kingma2016improving,papamakarios2017masked,marinescu2020bayesian}, have shown great potential for regularizing inverse problems~\citep{bora2017compressed,van2018compressed,hand2018phase,ongie2020deep,asim2020invertible,mosser2020stochastic,li2021multiparameter,siahkoohi2021preconditioned,pmlr-v139-whang21a,cheng2022inout,daras2021intermediate,daras2022score}. Such deep generative models directly learn from training data distributions and are a powerful and versatile prior. They map latent vectors \(\rvz\) to outputs \(\rvm\) distributed according to an \textit{a priori} distribution: \( \rvm = g(\rvz) \sim p_{\texttt{target}}, \rvz \sim \mathcal{N}(\mathbf{0}, \rmI) \), for example. The framework of DGM-regularized inversion~\citep{bora2017compressed} is
\begin{equation}
  \argmin_{\rvz} ~(1/2) \left\|\rvd - f \circ g\left(\rvz\right) \right\|^2_2 + \mathcal{R^\prime}(\rvz),
\label{eqn-augmented_inversion}
\end{equation}
where the deep generative model \(g\), whose layers are frozen, reparameterizes the original variable \(\rvm\), acting as a hard constraint. Instead of optimizing for \(\rvm\), we now estimate the latent variable \(\rvz\) and retrieve the inverted \(\rvm\) by forward mappings. Since the latent distribution is usually a standard Gaussian, the new (optional) regularization term \(\mathcal{R^\prime}(\rvz)\) can be chosen as \(\beta \|\rvz\|^2_2\) for GANs and VAEs, where \(\beta\) is a weighting factor. See \appref{append:glow_inv_formulation} for more details on a similar formulation for normalizing flows. Since the optimal \(\beta\) depends on the problem and data, tuning \(\beta\) is highly subjective and costly. 

However, this formulation of DGM-regularized inversion still leads to unsatisfactory results if the data are noisy or the forward model is inaccurate, as shown in \figref{fig:figs_intro}, even if we fine-tune the weighting parameter \(\beta\).
\begin{figure}[t!]
	\centering
	\includegraphics[width=0.8\textwidth]{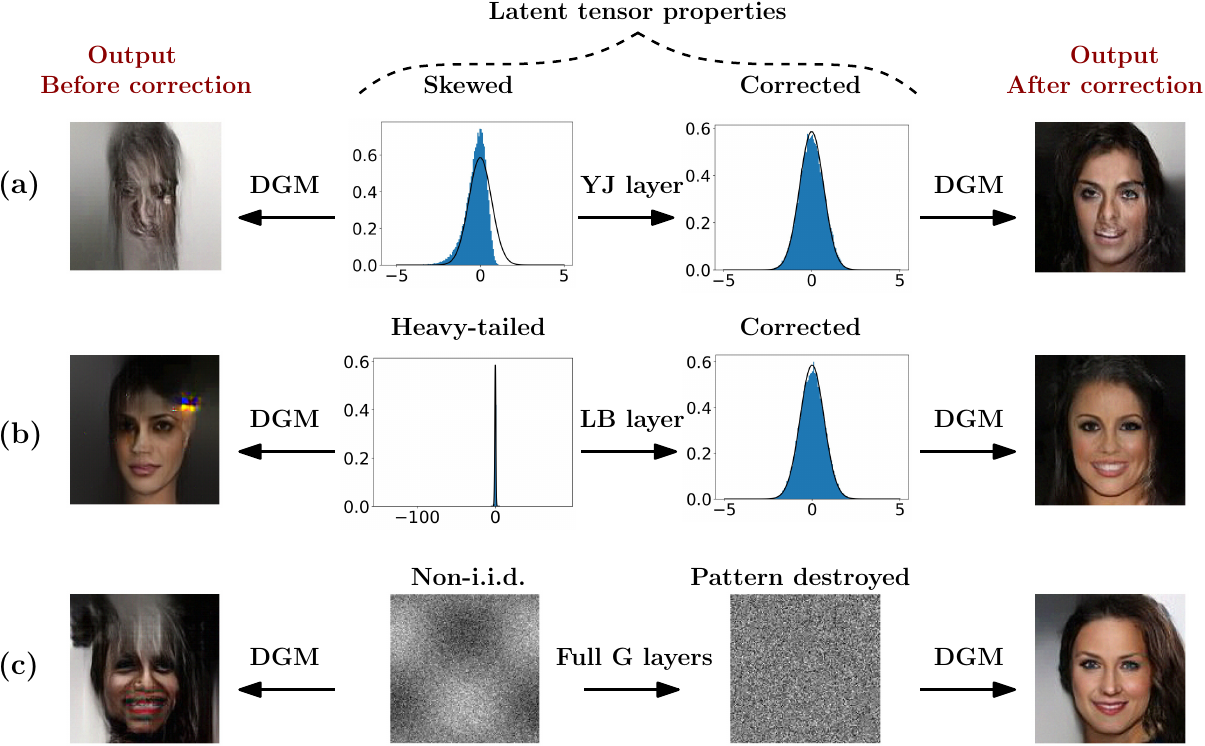}
	\caption{Comparison of images generated by a deep generative model (DGM), Glow, using latent tensors that deviate from a spherical Gaussian distribution (left) and those after corresponding corrections (right). The visual effects highlight the necessity of keeping the latent tensor within such a distribution during inversion. The second column shows the characteristics of deviated latent tensors: (a) histogram: \iid components but the distribution is skewed; (b) histogram: \iid components but the distribution is heavy-tailed; (c) latent tensor image: non-\iid entries. The first column shows the corresponding outputs of a Glow network. The third column shows latent tensors corrected by (a) the Yeo-Johnson layer (YJ), (b) the Lambert $W\times F_X$ layer (LB), and (c) the full set of our Gaussianization layers (G layers). Those corrected latent tensors map to realistic images shown in the fourth column. All latent tensors have a norm of 0.7$\sqrt{\text{vec dim}}$ because of the Gaussian Annulus Theorem (\appref{append:gaussian_typical_set}) and the fact that Glow works best with a temperature smaller than one (see \figref{fig:various_radii}). Additional examples for \textbf{StyleGAN2} and \textbf{Stable Diffusion}~\citep{Rombach_2022_CVPR} can be found in \appref{append:add_latent_deviations}.}
	\label{fig:effects_deviate_glow}
\end{figure}

To analyze this problem, first recall that a well-trained DGM has a latent space (usually) defined on a standard Gaussian distribution. In other words, a DGM either only sees standard Gaussian latent tensors during training (\eg, GANs) or learns to establish 1-1 mappings between training examples and typical samples from the standard Gaussian distribution (\eg, normalizing flows, \appref{append:KL_duality}). As a result, the generator may map out-of-distribution latent vectors to unrealistic results. We show in \figref{fig:effects_deviate_glow} the visual effects of several types of deviations of latent vectors from a spherical Gaussian (with a temperature of 0.7). It can be seen that (1) the latent tensor should have \iid entries, and (2) these entries should be distributed as a 1D standard Gaussian in order to generate plausible images. Since the traditional DGM-regularized inversion lacks such Gaussianity constraint, we conjecture\footnote{Out-of-distribution/typicality tests are very challenging for high-dimensional data~\citep{rabanser2019failing,nalisnick2018do}. Also, since latent tensors are Gaussian/Gaussian-like, it is hard to conduct dimension reduction for such tests.} that the latent tensor deviates from the desired high-dimensional standard Gaussian distribution during inversion, leading to poor results. Our observations and reasoning motivate us to propose a set of differentiable Gaussianization layers to reparameterize and Gaussianize the latent vectors of deep generative models (\eg, StyleGAN2~\citep{karras2020analyzing} and Glow~\citep{kingma2018glow}) for inverse problems.  The code is available at \url{https://github.com/lidongzh/DGM-Inv-Gaussianization}.

\section{Method}
\label{sec:method}
\subsection{Overview - reparameterization using Gaussianization layers}
\label{sec:reparam_gaussian_layers}
Our solution is based on a necessary condition of the standard Gaussian prior on latent tensors. Let us define a \textit{partition} of a latent tensor \(\rvz \in \mathbb{R}^n\) as the collection of non-overlapping patches \(P(\rvz) = \{\rvz_{i}\}_{i = 1,\cdots,N}\), where the patches \(\rvz_i \in \mathbb{R}^D\) are of the same dimension and can be assembled as \(\rvz\), \ie, \(n = N\times D\). If the latent tensor is a sample from the standard Gaussian, denoted as \(\rvz \sim \mathcal{N}(\bm{0}, \rmI)\), then for all \(\rvz_i\) from any partition \(P(\rvz)\), we have \(\rvz_i \sim \mathcal{N}(\bm{0}, \rmI)\).

Note that \(\rvz\) is a symbolic representation of the latent tensor. In a specific DGM, \(\rvz\) can be either a 2D/3D tensor or a list of such tensors corresponding to a multi-scale architecture (\appref{append:reparam_schemes}). Even though there are numerous partition schemes, such as random grouping of components, we choose the simplest: partitioning \(\rvz\) based on neighboring components, which works well in practice. 

To constrain \(\rvz_i \sim \mathcal{N}(\bm{0}, \rmI)\) during inversion, we reparameterize \(\rvz_i\) by constructing a mapping \(h: \rvv_i \rightarrow \rvz_i\), such that \(\rvz_i \sim \mathcal{N}(\bm{0}, \rmI)\). The new variables \(\rvv_i\) are of the same dimension as \(\rvz_i\). Suppose that we have constructed the patch-level mapping \(h\), we can obtain a mapping \(h^\dagger\) at the tensor level, such that \(\rvz = h^\dagger(\rvv)\), where \(\rvv\) is assembled from \(\rvv_i\) in the same way as \(\rvz\) from \(\rvz_i\). For example, \(h^\dagger = \mathtt{diag}(h, \cdots, h)\) if patches are extracted from neighboring components and are concatenated into a vectorized \(\rvv\). Hence, the original DGM-regularized inversion~\ref{eqn-augmented_inversion} becomes
   \begin{equation}
      \argmin_{\rvv} ~(1/2) \left\|\rvd - f \circ g \circ   h^\dagger(\rvv)   \right\|^2_2.
	  \label{eqn:reparameterized_augmented_inversion}
   \end{equation}
Since we are imposing a constraint through reparameterization, there is no need to include the regularization term \(\mathcal{R^\prime}(\rvz)\). The new formulation is still an unconstrained optimization problem, enabling us to use highly efficient unconstrained optimizers, such as L-BFGS~\citep{nocedal2006numerical} and ADAM~\citep{kingma2015adam}.

The remaining critical piece is to construct \(h\), and it leads to our Gaussianization layers. First, we translate the constraint of \(\rvz_i = h(\rvv_i) \sim \mathcal{N}(\bm{0}, \rmI)\) into the following optimization problem: 
\begin{equation}
  \argmin_{h} \KL\left( p_{Z}\left( h\left(\rvv_i\right) \right)\Vert \mathcal{N}(\bm{0}, \rmI) \right),
\end{equation}
where \(p_Z\) is the probability density function (PDF) of \(\rvz_i\). Second, we adopt the framework proposed by precursor works of normalizing flows on Gaussianization~\citep{chen2000gaussianization,laparra2011iterative} to solve this optimization problem. The KL-divergence can be decomposed as the sum of the multi-information \(I(\rvz_i)\) and the marginal negentropy \(J_m(\rvz_i)\)~\citep{chen2000gaussianization,meng2020gaussianization}:
   \begin{equation}
   		\KL \left(p_Z\left(\rvz_i \right) \Vert \, \mathcal{N}\left(\bm{0}, \mathbf{I}\right)\right) = I(\rvz_i) + J_m(\rvz_i),
		\label{eqn:KL_div}
   \end{equation}
where
   \begin{equation}
   		I(\rvz_i) = \KL \left( p_Z\left( \rvz_i \right) \Big\Vert \prod_j^D p_j(z_i^{(j)}) \right), \text{ and  } J_m(\rvz_i) = \sum_{j=1}^D \KL \left( p_j(z_i^{(j)} ) \Big\Vert \mathcal{N}(0, 1) \right).
       \label{eqn:1D_gaussianization}
   	\end{equation}
Here \(z_i^{(j)}\) denotes the \(j\)-th component of patch vector \(\rvz_i\), and \(p_j\) stands for the marginal PDF for that component. The multi-information \(I(\rvz_i)\) quantifies the independence of the components of \(\rvz_i\), while the marginal negentropy \(J_m(\rvz_i)\) describes how close each component is to a 1D standard Gaussian. With this decomposition, the optimization procedure depends on the facts: (1) the KL divergence and a standard Gaussian in \eqref{eqn:KL_div} are invariant to an orthogonal transformation, and (2) the multi-information term is invariant to a component-wise invertible differentiable transformation (\appref{append:KL-MI-proofs}). As a result, we perform Gaussianization in two steps:
  
1. Minimize the multi-information \(I(\rvz_i) \). This is done by an orthogonal transformation that keeps the overall KL divergence the same but increases the negentropy \(J_m(\rvz_i)\). We achieve this by using our independent component analysis (ICA) layer. ICA is the optimal choice since it maximizes non-Gaussianity so that the subsequent marginal Gaussianization step removes it and results in a large decrease in \(\KL\left( p_{Z}\left( h\left(\rvv_i\right) \right)\Vert \mathcal{N}(\bm{0}, \rmI) \right)\).

2. Minimize the marginal negentropy \(J_m(\rvz_i)\) by component-wise operations that perform 1D Gaussianization of marginal distributions \(p_j, {}_{j=1,\cdots,D}\). The multi-information does not change under component-wise invertible operations. Therefore, the overall KL divergence between \(\rvz_i\) and the Gaussian distribution decreases.

The Gaussianization steps are well-aligned with the motivating example (\figref{fig:effects_deviate_glow}). To constrain DGM outputs to be plausible, one should make components within latent tensor patches independent (or destroy the patterns) (\figref{fig:effects_deviate_glow}(c)) and shape the 1D distribution as close as possible to Gaussian (\figref{fig:effects_deviate_glow}(a)(b)).

We will see that \(h\) is parameterized by an orthogonal matrix and two scalar parameters in 1D Gaussianization layers. Unlike conventional neural network layers, the input-data-dependent Gaussianization layers are not defined by learning from a dataset ahead of time but by solving certain optimization problems on the job (\figref{fig:opt_layer}). Special care should be taken to implement the gradient computation correctly and ensure that they pass the finite-difference convergence test (\appref{append:opt_layer_backward}). As an overview, the composition of our proposed layers is:
\begin{equation*}
  {\rvv \boldsymbol{\rightarrow} \textbf{Whitening} \boldsymbol{\rightarrow} \textbf{ICA} \boldsymbol{\rightarrow} \textbf{Yeo-Johnson} \boldsymbol{\rightarrow} \textbf{Lambert }W\times F_X \boldsymbol{\rightarrow} \textbf{Standardization} \boldsymbol{\rightarrow} \rvz},
\end{equation*}
where whitening and ICA belong to the first step and the rest belong to the second step. We will discuss in \appref{append:additional_discussions} some possible simplifications of the layers in practice after our ablation studies.

\begin{figure}[htpb]
	\centering
			\includegraphics[width=0.8\textwidth]{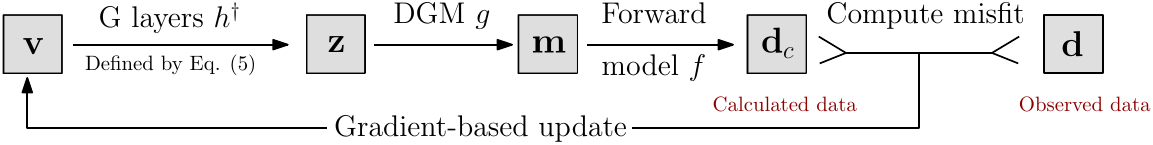}
	\caption{Illustration of the proposed inversion process. Gradient computation in the Gaussianization layers is enabled by the implicit function theorem and automatic differentiation (\appref{append:opt_layer_backward}). We use the L-BFGS optimizer~\citep{nocedal2006numerical} to update \(\rvv\).}
	\label{fig:inversion_illustration}
\end{figure}

The overall proposed inversion process (one iteration) is illustrated in \figref{fig:inversion_illustration}.

\subsection{Reducing multi-information -- ICA layer}
\label{sec:ica_layer}
The orthogonal matrix \(\mW\) is constructed by the independent component analysis (ICA). The input is the patch vectors \(\{\rvv_{i}\vert {}_{i = 1,\cdots,N}\}\). The ICA algorithm first computes the input-dependent orthogonal matrix \(\mW\) and then computes \(\rvp_i = \mW^\top \rvv_i, {}_{i = 1,\cdots,N}\) as the output. The orthogonal matrix \(\mW\) makes the entries of each \(\rvp_i\) independent. 
    
We use the FastICA algorithm~\citep{hyvarinen1999fast,hyvarinen2000independent}, which employs a fixed-point algorithm to maximize a contrast function \(\Phi\) (\eg, the logcosh function), for our ICA layer to reduce multi-information.

The FastICA algorithm typically requires that the data are pre-whitened. We adopt the ZCA whitening method or the iterative whitening method introduced in \citet{hyvarinen1999fast} (\appref{append:G_layers_details}).
With the whitened data, we compute \(\mW\) using a damped fixed-point iteration scheme:
\begin{equation}
  \mW = \frac{1}{N} \left[\alpha \mV \phi\left(\mW^\top \mV\right)^\top - \mW \mathtt{diag}\left( \phi'\left( \mW^\top \mV \right) \mathbf{1} \right)\right],
  \label{eqn:mat_form_ica}
\end{equation}
where \(\mathbf{1}\) is an all-one vector, the column vectors of \(\mV\) are \(\{\rvv_{i}\vert {}_{i = 1,\cdots,N}\}\), \(\phi(\cdot) = \Phi'(\cdot)\), \(\alpha\in (0,1)\), and we use \(\alpha=0.8\) throughout our experiments. To save computation time, we only perform a maximum of 10 iterations. The details of the whole algorithm can be found in \appref{append:G_layers_details}.

We set the initial \(\mW\) as an identity matrix. If the input vectors are already standard Gaussian (in-distribution), the computed \(\mW\) will still be an identity matrix, which maps the input to the same output. In practice, the empirical distribution from finite samples is not a standard Gaussian, so \(\mW\) is not an identity matrix but another orthogonal matrix, which still maps standard Gaussian input vectors to standard Gaussian vectors as output. For this reason, we sample starting \(\{\rvv_{i}\vert {}_{i = 1,\cdots,N}\}\) from the standard Gaussian distribution to start inversions with plausible outputs.

\subsection{Reducing marginal negentropy}
For 1D Gaussianization, we choose a combination of the Yeo-Johnson transformation that reduces skewness and the Lambert \(W\times F_X\) transformation that reduces heavy-tailedness. Both are layers based on optimization problems with only one parameter, which is cheap to compute and is easy to back-propagate the gradient. \Eqref{eqn:1D_gaussianization} requires us to perform such 1D transformations for each component. In other words, we need to solve the same optimization problem for \(D\) times, which imposes a substantial computational burden. Instead, we empirically find it acceptable to share the same optimization-generated parameter across all components. In other words, we perform only a single 1D Gaussianization, treating all entry values in the latent vector as the data simultaneously.

\begin{wrapfigure}{R}{0.48\textwidth}
% \begin{figure}[htpb]
	\centering
		\setlength{\tabcolsep}{0.1cm}
		\begin{tabular}{ll}
			\small{(a)} & \small{(b)} \\
			\includegraphics[width=0.22\textwidth]{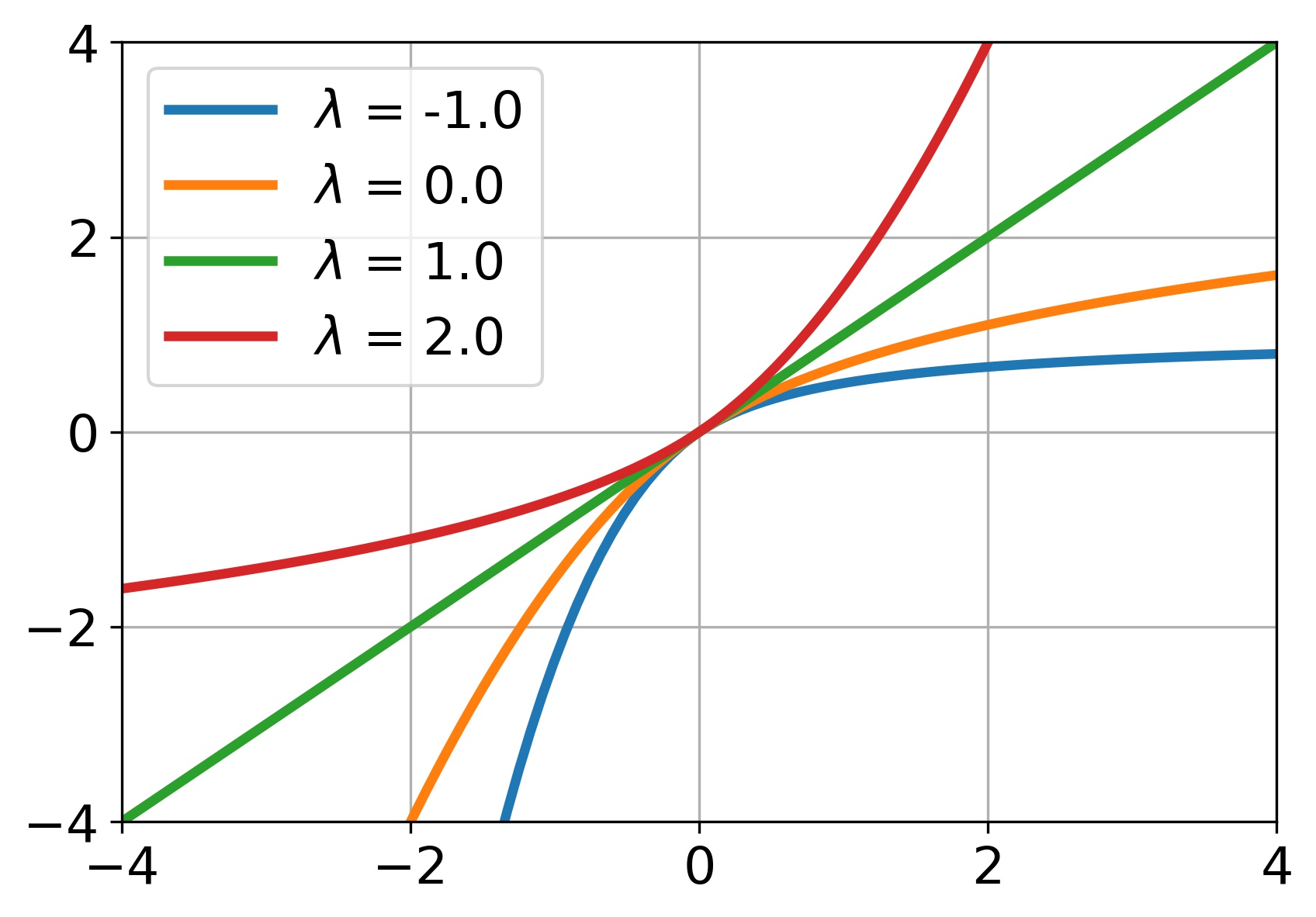} & \includegraphics[width=0.22\textwidth]{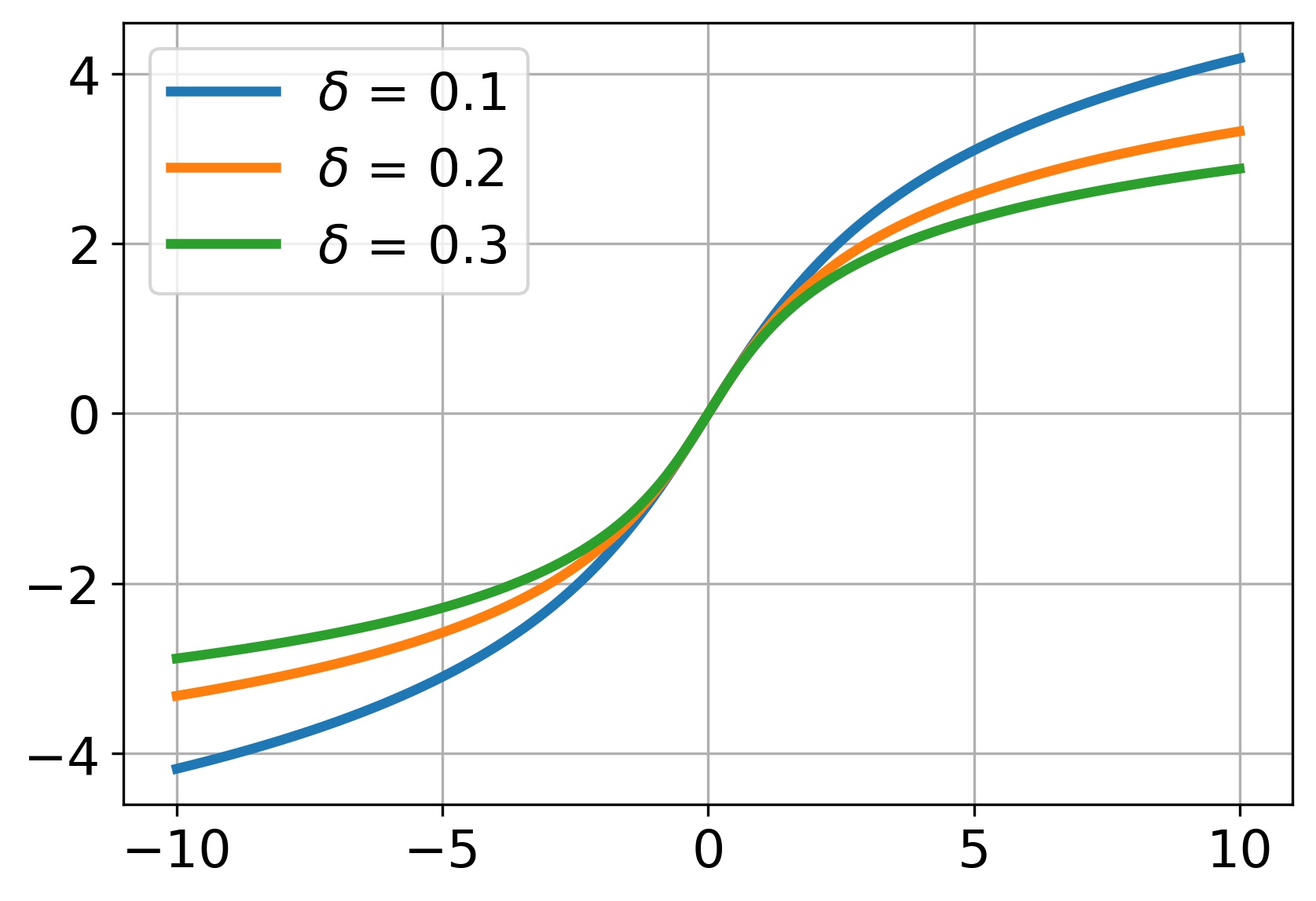}
		\end{tabular}
	\caption{The nonlinear activation functions from (a) the power transformation (Yeo-Johnson) layer and (b) the Lambert \(W \times F_X\) layer.}
	\label{fig:lambert_yeojohnson_activation}
% \end{figure}
\end{wrapfigure}

\paragraph{Power transformation layer}
We propose to use the power transformation or Yeo-Johnson transformation~\citep{yeo2000new} to reduce the skewness of distributions. As shown in \figref{fig:lambert_yeojohnson_activation}(a), the form of the Yeo-Johnson activation function depends on parameter \(\lambda\). If \(\lambda = 1\), the mapping is an identity mapping. If \(\lambda > 1\), the activation function is convex, compressing the left tail and extending the right tail, reducing the left-skewness. If \(\lambda < 1\), the activation function is concave, which oppositely reduces the right-skewness.
We refer the readers to \appref{append:G_layers_details} for details.

\paragraph{Lambert \(W\times F_X\) layer}
Due to noise and inaccurate forward models, we observe that the distribution of latent vector values tends to be shaped as a heavy-tailed distribution during the inversion process. To reduce the heavy-tailedness, we adopt the Lambert \(W\times F_X\) method detailed in~\citet{goerg2015lambert}. 

We use the parameterized Lambert \(W\times F_X\) distribution family to approximate a heavy-tailed input and solve an optimization to estimate an optimal parameter \(\delta\) (\appref{append:G_layers_details}), with which the inverse transformation maps the heavy-tailed distribution towards a Gaussian.

\Figref{fig:lambert_yeojohnson_activation}(b) shows that the Lambert \(W\times F_X\) layer acts as a nonlinear squashing function. As \(\delta\) increases, it compresses more the large values and reduces the heavy-tailedness. Intuitively, the Lambert \(W\times F_X\) layer can also be interpreted as an intelligent way of imposing constraints on the range of values instead of a simple box constraint. We refer the readers to \appref{append:G_layers_details} for more details about the optimization problem and implementation.

\paragraph{Standardization with temperature} Since the Lambert \(W\times F_X\) layer output may not necessarily have a zero mean and a unit (or a prescribed) variance, we standardize the output using
\begin{equation}
	\rvz = \left(\rvx - \mathtt{mean}\left(\rvx\right)\right)/\mathtt{std}(\rvx) *\gamma,
\end{equation}
where \(\mathtt{mean}\) and \(\mathtt{std}\) represent the mean and the standard deviation, respectively, of the elements of the vector \(\rvx\), and \(\gamma\) is the temperature parameter suggested in \citet{kingma2018glow}.

\section{Related work}
\label{sec:related_work}
\paragraph{End-to-end NNs for inverse problems} There are numerous end-to-end neural networks designed for inverse problems, using CNNs~\citep{chen2017low,jin2017deep,sriram2020end}, GANs~\citep{mardani2018deep,lugmayr2020srflow,wang2018esrgan}, invertible networks~\citep{ardizzone2018analyzing}, and diffusion models~\citep{kawar2022denoising}. The general idea is simple: train a neural network that directly maps observed data to estimated parameters. Even though such methods seem effective in a few applications, DGM-regularized inversion with our Gaussianization layers is preferable for the following reasons. First, the forward modeling can be so expensive computationally that it is infeasible to collect a decent training datasets for some applications. For example, one large-scale fluid mechanics or wave propagation simulation can take hours, if not days. Second, the relationship between parameters and data can be highly nonlinear, and multiple solutions may exist. An end-to-end network may map data to interpolated solutions that are not realistic. In comparison, our method can start from different initializations or even employ sampling techniques to address this issue. Third, the configuration of data collection can change from experiment to experiment. It is impractical, for example, to re-train the network each time we change the number and locations of sensors. In contrast, not only can our method deal with this situation, but it can even use the same DGM for different forward models, as we can see in the compressive sensing MRI and eikonal tomography examples. While almost all end-to-end methods are only applied to linear inverse problems, our Gaussianization layers are also effective in nonlinear problems.

\paragraph{Other techniques to improve DGM-regularized inversion}
In high-dimensional space, the probability mass of a standard Gaussian distribution concentrates within the so-called Gaussian typical set (\appref{append:gaussian_typical_set}). To be in the Gaussian typical set, one necessary but not sufficient condition is to be within an annulus area around a high-dimensional sphere (\appref{append:gaussian_typical_set}). Utilizing this geometric property, DGM-regularized inversion methods like \citet{bojanowski2017optimizing} and \citet{liang2021flow} force updated latent vectors to stay on the sphere. This strategy is closely related to spherical interpolation~\citep{white2016sampling}. We call this strategy the \textbf{spherical constraint} for inversion. In the original StyleGAN2 paper, the authors also noticed that in image projection tasks, the noise maps tend to have leakage from signals~\citep{karras2020analyzing} -- the same phenomenon we discussed. They proposed a multi-scale noise regularization term (\textbf{NoiseRlg}) to penalize spatial correlation in noise maps. We extend the same technique to our inverse problems for comparison. Note that we use a whitening layer before the ICA layer. The \textbf{whitening layer} can be used alone, similar to \citet{huang2018decorrelated} and \citet{siarohin2018whitening}, whose performance will be reported in ablation studies. Also, \citet{wulff2020improving} observed that for StyleGAN2, the Leaky ReLU function can ``Gaussianize" latent vectors in the W space. \textbf{CSGM-w}~\citep{kelkar2021prior} utilizes this \textit{a priori} knowledge to improve DGM-regularized compressive sensing problems. To further compare with the Gaussianizaion layers, we in addition propose an alternative idea that reparameterizes latent vectors using learnable orthogonal matrices (Cayley parameterization) and fixed latent vectors, which is closely related to the work of orthogonal over-parameterized training~\citep{liu2021orthogonal} (\appref{append:ortho_param} ). Recently, score-based generative models have started to show promise for solving inverse problems~\citep{jalal2021robust,song2021solving}. However, they have been mainly applied to linear inverse problems and challenged by noisy data~\citep{kawar2021snips}. Besides, scored-based methods might not work for certain physics-based inverse problems, since Gaussian noise parameters may break the physics simulation engine\footnote{For example, in elastic waveform inversion, initializing material properties using Gaussian noise may create unrealistic scenarios where the P-wave speed is lower than the S-wave speed at some spatial locations.}. 

\section{Experiments}
We consider three representative inversion problems for testing: compressive sensing MRI, image deblurring, and eikonal traveltime tomography. For MRI and eikonal tomography, we used synthetic brain images as inversion targets and used the pre-trained StyleGAN2 weights from \citet{kelkar2021prior} (trained on data from the databases of fastMRI~\citep{zbontar2018fastmri,knoll2020fastmri}, TCIA-GBM~\citep{scarpace2016radiology}, and OASIS-3~\citep{lamontagne2019oasis}) for regularization. We used the test split of the CelebA-HQ dataset~\citep{karras2017progressive} for deblurring, and the DGM is a Glow network trained on the training split. We refer readers to \appref{append:datasets_training} for details on datasets and training. 

We tested each parameter configuration in each inversion on 100 images (25 in the eikonal tomography due to its expensive forward modeling). Since the deep generative models are highly nonlinear, the results may get stuck in local minima. Thus, we started inversion using three different randomly initialized latent tensors for each of the 100 or 25 images, picked the best value among the three for each metric, and reported the mean and standard deviation of those metrics, except for CSGM-w, TV, and NoiseRlg, where the initialization is fixed. The metrics we used were PSNR, SSIM~\citep{wang2004image}, and an additional LPIPS~\citep{zhang2018perceptual} for the CelebA-HQ data. We used the LBFGS~\citep{nocedal2006numerical} optimizer in all experiments except TV, noise regularization, and CSGM-w, which use FISTA~\citep{beck2009fast} or ADAM~\citep{kingma2015adam}. The temperature was set to 1.0 for StyleGAN2 and 0.7 for Glow.
\subsection{Compressive sensing MRI using StyleGAN2}
The mathematical model of compressive sensing MRI is
\begin{equation}
  \rvd = \mA \rvm + \bm{\epsilon},
  \label{eqn:mri_eq}
\end{equation}
where $\mA \in \mathbb{C}^{M\times N}$ is the sensing matrix, which consists of FFT and subsampling in the k-space (frequency domain). \eqref{eqn:mri_eq} is an under-determined system, and we use \(\text{Accl}=N/M\) to denote the acceleration ratio. We also added \iid complex Gaussian noise with a signal-to-noise ratio (SNR) of 20 dB or 10 dB to the measured data. See \appref{append:MRI} for more background information.

\tabref{tab:MRI_results_comparison} compares the results from total variation regularization (TV), noise regularization (NoiseRlg)~\citep{karras2020analyzing}, spherical constraint/reparameterization: \(\vz = \vv / \|\vv\|_2 * \sqrt{\texttt{dim}(\vv)}\), CSGM-w~\citep{kelkar2021prior}, our proposed orthogonal reparameterization (Orthogonal), and our proposed Gaussianization layers (G layers). \Figref{fig:comparison_mri} shows examples of inversion results. In the base case where Accl=8x and SNR=20 dB, the Gaussianization layers give the best scores, and this advantage gets more significant when data SNR decreases to 10 dB. Interestingly, the scores from all methods improve significantly if we make the system better determined (\ie, Accl=2x), and the performance of TV, spherical constraint, and Gaussianization layers become more similar in this scenario. We conclude that our proposed Gaussianization layers are effective and more robust than other methods, particularly in low-SNR scenarios.

\begingroup
\setlength{\tabcolsep}{1.2pt}
\begin{figure}[htpb]
    \centering%
    \newcommand{\sizeI}{0.11}
    \newcommand{\imgTrue}[1]{\includegraphics[width=\sizeI\linewidth]{figures/figures_inv_mri/tv_accl8x_snr20.0/#1/True.jpg}}
    \newcommand{\imgTV}[1]{\includegraphics[width=\sizeI\linewidth]{figures/figures_inv_mri/tv_accl8x_snr20.0/#1/Inv.jpg}}
    \newcommand{\imgSP}[1]{\includegraphics[width=\sizeI\linewidth]{figures/figures_inv_mri/sphere_accl8x_snr20.0/#1/Inv.jpg}}
    \newcommand{\imgCSGM}[1]{\includegraphics[width=\sizeI\linewidth]{figures/figures_inv_mri/csgm_accl8x_snr20.0/#1/Inv.jpg}}
    \newcommand{\imgNoiseRlg}[1]{\includegraphics[width=\sizeI\linewidth]{figures/figures_inv_mri/noisergl_accl8x_snr20.0/#1/Inv.jpg}}
    \newcommand{\imgOrtho}[1]{\includegraphics[width=\sizeI\linewidth]{figures/figures_inv_mri/ortho_accl8x_snr20.0/#1/Inv.jpg}}
    \newcommand{\imgGauss}[1]{\includegraphics[width=\sizeI\linewidth]{figures/figures_inv_mri/gaus_accl8x_snr20.0/#1/Inv.jpg}}
	\begin{tabular}{ccccccc}
	\small{True} & \small{TV} & \small{Spherical} & \small{CSGM-w} & \small{NoiseRlg} & \small{Orthogonal} & \small{G layers}\\
	\imgTrue{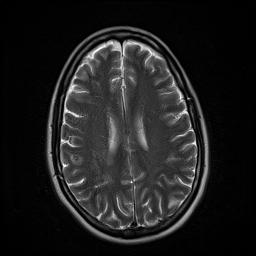} & \imgTV{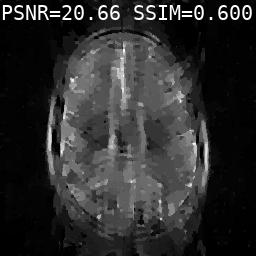} &\imgSP{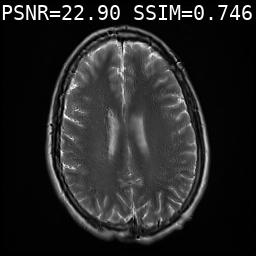}& \imgCSGM{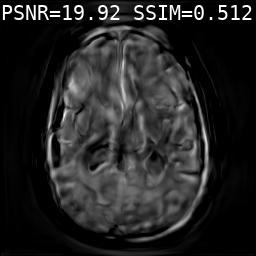} 
        & \imgNoiseRlg{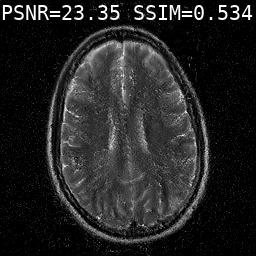} & \imgOrtho{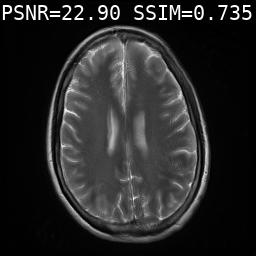} & \imgGauss{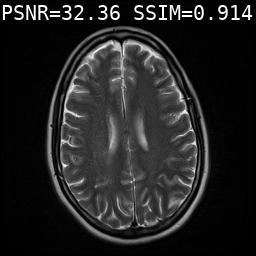} \\
	\imgTrue{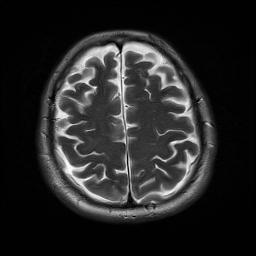} & \imgTV{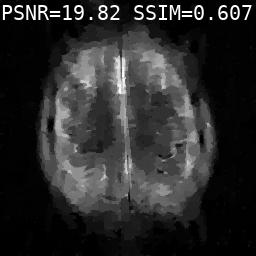} &\imgSP{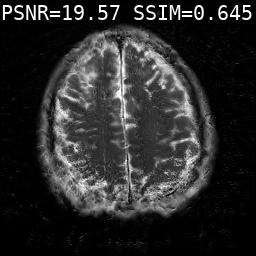}& \imgCSGM{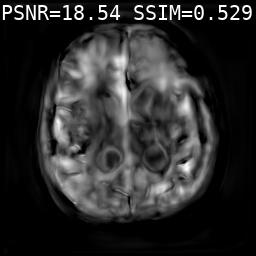} 
        & \imgNoiseRlg{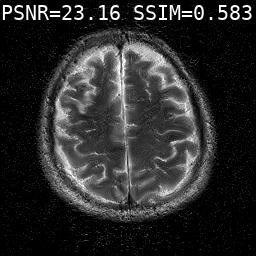} & \imgOrtho{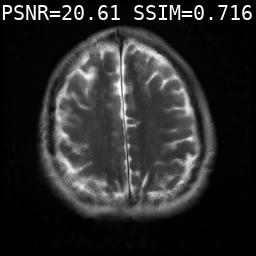} & \imgGauss{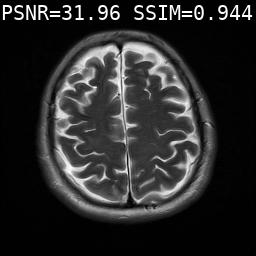} \\
	\imgTrue{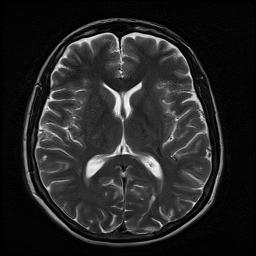} & \imgTV{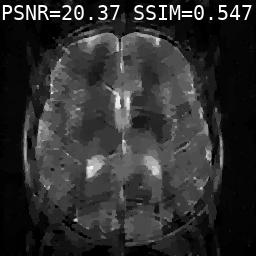} &\imgSP{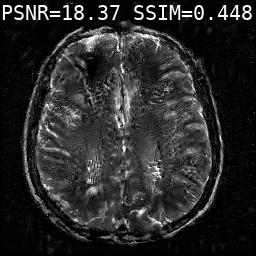}& \imgCSGM{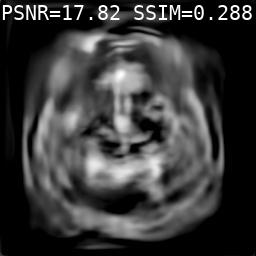} 
        & \imgNoiseRlg{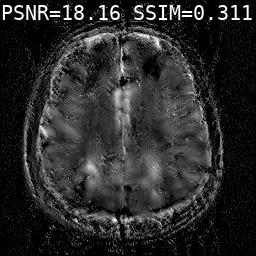} & \imgOrtho{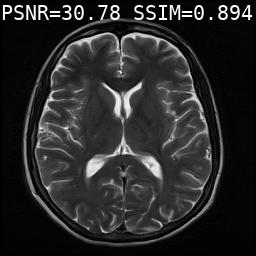} & \imgGauss{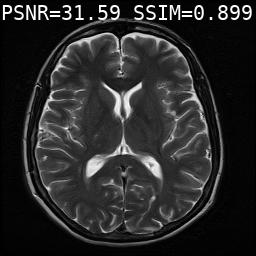} \\
	\end{tabular}
	\caption{Comparison of compressive sensing MRI inversion results (Accl=8x, SNR=20 dB).}
	\label{fig:comparison_mri}
\end{figure}
\endgroup

\begingroup
\setlength{\extrarowheight}{1pt}
\begin{table}[htpb]
\caption{Comparison of compressive sensing MRI results.}
\label{tab:MRI_results_comparison}
\centering
\small
\begin{tabular}{l@{~~~~}c@{~~~~~}c@{~~~~~}c@{~~~~~}c@{~~~~~}c@{~~~~~}c@{~~~~~}}
 \toprule
  \multirow{2}{3em}{Method} 
            & \multicolumn{2}{c}{Accl = 8x, SNR = 20 dB} & \multicolumn{2}{c}{Accl = 8x, SNR = 10 dB}
            & \multicolumn{2}{c}{Accl = 2x, SNR = 20 dB} \\
            \cmidrule(r){2-3} \cmidrule(r){4-5} \cmidrule(r){6-7}
            & PSNR$\uparrow$ & SSIM$\uparrow$ & PSNR$\uparrow$ & SSIM$\uparrow$ & PSNR$\uparrow$ & SSIM$\uparrow$ \\
 \hline
 TV         & 20.20\tiny{$\pm$1.33} & 0.60\tiny{$\pm$0.063}
            & 19.78\tiny{$\pm$1.32} & 0.57\tiny{$\pm$0.058}
            & \best{32.92\tiny{$\pm$1.36}} & \best{0.91\tiny{$\pm$0.024}} \\

 Spherical  & 26.93\tiny{$\pm$6.57} & 0.79\tiny{$\pm$0.159}
            & 22.78\tiny{$\pm$5.38} & 0.61\tiny{$\pm$0.246}
            & 32.90\tiny{$\pm$3.50} & 0.90\tiny{$\pm$0.061} \\

 CSGM-w     & 20.19\tiny{$\pm$2.11} & 0.57\tiny{$\pm$0.096}
            & 19.98\tiny{$\pm$1.94} & 0.55\tiny{$\pm$0.091}
            & 27.53\tiny{$\pm$2.98} & 0.74\tiny{$\pm$0.085} \\

 NoiseRgl   & 21.61\tiny{$\pm$2.27} & 0.50\tiny{$\pm$0.073}
            & 18.09\tiny{$\pm$1.05} & 0.27\tiny{$\pm$0.036}
            & 28.04\tiny{$\pm$1.46} & 0.66\tiny{$\pm$0.037} \\

 Orthogonal & 26.14\tiny{$\pm$5.79} & 0.78\tiny{$\pm$0.134}
            & 25.10\tiny{$\pm$5.08} & 0.77\tiny{$\pm$0.132}
            & 29.29\tiny{$\pm$5.32} & 0.84\tiny{$\pm$0.108} \\

 G layers   & \best{27.99\tiny{$\pm$5.70}} & \best{0.83\tiny{$\pm$0.128}}
            & \best{25.48\tiny{$\pm$4.76}} & \best{0.78\tiny{$\pm$0.149}}
            & 32.41\tiny{$\pm$4.60} & 0.90\tiny{$\pm$0.079} \\
\bottomrule
\end{tabular}
\end{table}
\endgroup 

\paragraph{Ablation study}
\tabref{tab:ablation_mri} summarizes the ablation study on the components of the Gaussianization layers. We kept the standardization layer on for all cases. The conclusions are as follows: 1. The ICA layer plays the most significant role in improving result scores; 2. The whitening/ZCA layer alone is not effective; 3. The Yeo-Johnson (YJ) and the Lambert \(W\times F_X\) (Lambt) layers are more effective when the noise level is higher (\eg, SNR=10 dB vs. 20 dB). Their performance seems data-dependent: when SNR=20 dB, YJ seems more effective, while Lambt gives the best scores when SNR=10 dB. But the overall improvement from these two layers is marginal. In practice, one may only use ICA and one of the 1D Gaussianization layers. Additionally, we tested the effect of patch size on the z (style) vectors (\appref{append:additional_exp}). We find that the largest possible patch size gives the best results.
\begingroup
\setlength{\extrarowheight}{1pt}
\begin{table}[htpb]
\caption{Ablation study of the Gaussianization layers.}
\label{tab:ablation_mri}
\centering
\small
\begin{tabular}{l@{~~}l@{~~~}l@{~~~~}c@{~~~~~}c@{~~~~~}c@{~~~~~}c@{~~~~~}c@{~~~~~}c@{~~~~~}}
 \toprule
  \multicolumn{3}{c}{\multirow{2}{3em}{Method}}
            & \multicolumn{2}{c}{Accl = 8x, SNR = 20 dB} & \multicolumn{2}{c}{Accl = 8x, SNR = 10 dB} \\
            \cmidrule(r){4-5} \cmidrule(r){6-7} 
    &   &   & PSNR$\uparrow$ & SSIM$\uparrow$ & PSNR$\uparrow$ & SSIM$\uparrow$ \\
 \hline
  ICA (\xmark),\,& YJ (\xmark),& Lambt (\xmark)
            & 26.93\tiny{$\pm$6.40} & 0.787\tiny{$\pm$0.153}
            & 22.98\tiny{$\pm$5.54} & 0.604\tiny{$\pm$0.252} \\

  ICA (\xmark),\,& YJ (\checkmark),& Lambt (\xmark)
            & 26.92\tiny{$\pm$6.42} & 0.787\tiny{$\pm$0.156}
            & 23.14\tiny{$\pm$5.44} & 0.622\tiny{$\pm$0.245} \\

  ICA (\xmark),\,& YJ (\xmark),& Lambt (\checkmark)
            & 25.33\tiny{$\pm$5.89} & 0.743\tiny{$\pm$0.163}
            & 23.58\tiny{$\pm$5.19} & 0.695\tiny{$\pm$0.189} \\
  
  ZCA,\,& YJ (\xmark),& Lambt (\xmark)
            & 27.08\tiny{$\pm$6.52} & 0.786\tiny{$\pm$0.157}
            & 22.94\tiny{$\pm$5.50} & 0.623\tiny{$\pm$0.235} \\

  ICA (\checkmark),\,& YJ (\xmark),& Lambt (\xmark)
            & 27.91\tiny{$\pm$5.77} & 0.824\tiny{$\pm$0.129}
            & 25.22\tiny{$\pm$4.76} & 0.770\tiny{$\pm$0.154} \\
  
  ICA (\checkmark),\,& YJ (\checkmark),& Lambt (\xmark)
            & \best{27.99\tiny{$\pm$5.70}} & \best{0.831\tiny{$\pm$0.128}}
            & 25.48\tiny{$\pm$4.76} & \best{0.779\tiny{$\pm$0.149}} \\

  ICA (\checkmark),\,& YJ (\xmark),& Lambt (\checkmark)
            & 27.21\tiny{$\pm$5.74} & 0.816\tiny{$\pm$0.125}
            & \best{25.57\tiny{$\pm$4.98}} & \best{0.779\tiny{$\pm$0.148}} \\

  ICA (\checkmark),\,& YJ (\checkmark),& Lambt (\checkmark)
            & 27.37\tiny{$\pm$5.90} & 0.819\tiny{$\pm$0.135}
            & 25.09\tiny{$\pm$4.91} & 0.771\tiny{$\pm$0.144} \\

\bottomrule
\end{tabular}
\end{table}
\endgroup

\subsection{Image deblurring using Glow}
The mathematical model for deblurring is
\begin{equation}
  \rvd = \mH \ast \rvm + \bm{\epsilon},
\end{equation}
where \(\mH\) is a smoothing filter and \(\ast\) denotes convolution. We used a Gaussian smoothing filter with a standard deviation of 3, and added noise \(\bm{\epsilon}\sim \mathcal{N}(\mathbf{0}, 50^2 \rmI)\) to the observed data. Although the system may not be under-determined, the high-frequency information is lost due to low-pass filtering; hence this is also an ill-posed problem. Though we only tested on facial images, deblurring has wide applications in astronomy and geophysics. See \appref{append:deblurring} for more background information.
% \begin{wrapfigure}{R}{0.5\textwidth}
\begingroup
\setlength{\tabcolsep}{1pt}
\begin{figure}[htpb]
    \centering%
    \newcommand{\sizeI}{0.09}
    \newcommand{\imgTrue}[1]{\includegraphics[width=\sizeI\linewidth]{figures/figures_inv_deblur/gaus_p2/#1/True.jpg}}
    \newcommand{\imgSmooth}[1]{\includegraphics[width=\sizeI\linewidth]{figures/figures_inv_deblur/gaus_p2/#1/Obs.jpg}}
    \newcommand{\imgSph}[1]{\includegraphics[width=\sizeI\linewidth]{figures/figures_inv_deblur/spherical/#1/Inv.jpg}}
    \newcommand{\imgOrt}[1]{\includegraphics[width=\sizeI\linewidth]{figures/figures_inv_deblur/ortho/#1/Inv.jpg}}
    \newcommand{\imgGau}[1]{\includegraphics[width=\sizeI\linewidth]{figures/figures_inv_deblur/gaus_p1/#1/Inv.jpg}}
 \begin{tabular}{l@{~}|@{~}l}

    \begin{tabular}{c c c c c}
    \scriptsize{Original} & \scriptsize{Observed} & \scriptsize{Spherical} & \scriptsize{Orthogonal} & \scriptsize{G layers}\\
    \imgTrue{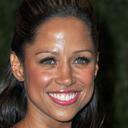} & \imgSmooth{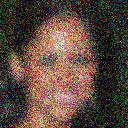} & \imgSph{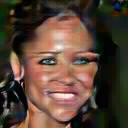} &\imgOrt{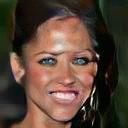} & \imgGau{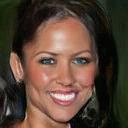} \\
    \imgTrue{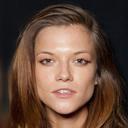} & \imgSmooth{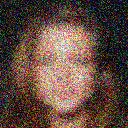} & \imgSph{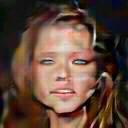} &\imgOrt{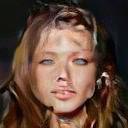} & \imgGau{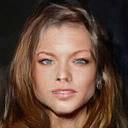} \\
    \end{tabular}
    &
    \begin{tabular}{c c c c c}
    \scriptsize{Original} & \scriptsize{Observed} & \scriptsize{Spherical} & \scriptsize{Orthogonal} & \scriptsize{G layers}\\
    \imgTrue{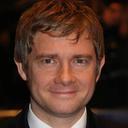} & \imgSmooth{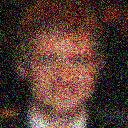} & \imgSph{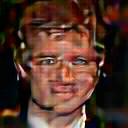} &\imgOrt{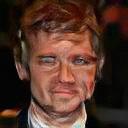} & \imgGau{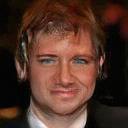}\\
    \imgTrue{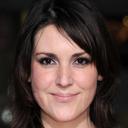} & \imgSmooth{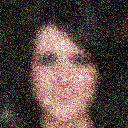} & \imgSph{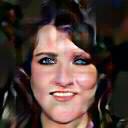} &\imgOrt{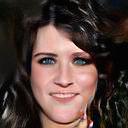} & \imgGau{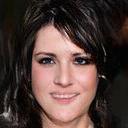}\\
    \end{tabular}

\end{tabular}
	\caption{Comparison of deblurring results from different methods.}
	\label{fig:glow_deblur_celeba-hq}
\end{figure}
\endgroup
% \end{wrapfigure}

% % compare with langevin
% \begingroup
% \setlength{\tabcolsep}{1pt}
% \begin{figure*}[htpb]
%     \centering%
%     \newcommand{\sizeI}{0.12}
%     \newcommand{\imgTrue}[1]{\includegraphics[width=\sizeI\linewidth]{figures/figures_inv_deblur/gaus_p2/#1/True.jpg}}
%     \newcommand{\imgSmooth}[1]{\includegraphics[width=\sizeI\linewidth]{figures/figures_inv_deblur/gaus_p2/#1/Obs.jpg}}
%     \newcommand{\imgLgv}[1]{\includegraphics[width=\sizeI\linewidth]{figures/figures_lgv/#1.jpg}}
%     \newcommand{\imgGau}[1]{\includegraphics[width=\sizeI\linewidth]{figures/figures_inv_deblur/gaus_p1/#1/Inv.jpg}}
%  \begin{tabular}{l@{~}|@{~}l}

%     \begin{tabular}{c c c c}
%     \scriptsize{Original} & \scriptsize{Observed} & \scriptsize{Langevin} & \scriptsize{G layers}\\
%     \imgTrue{186521} & \imgSmooth{186521} & \imgLgv{186521} & \imgGau{186521} \\
%     \imgTrue{185120} & \imgSmooth{185120} & \imgLgv{185120} & \imgGau{185120} \\
%     \end{tabular}
%     &
%     \begin{tabular}{c c c c}
%     \scriptsize{Original} & \scriptsize{Observed} & \scriptsize{Langevin} & \scriptsize{G layers}\\
%     \imgTrue{194792} & \imgSmooth{194792} & \imgLgv{194792} & \imgGau{194792}\\
%     \imgTrue{185989} & \imgSmooth{185989} & \imgLgv{185989} & \imgGau{185989}\\
%     \end{tabular}

% \end{tabular}
% 	\caption{Comparison of deblurring results from annealed Langevin dynamics and Gaussianization layers.}
% \end{figure*}
% \endgroup

\tabref{tab:deblurring_results} and \figref{fig:glow_deblur_celeba-hq} show that the Gaussianization layers are also effective in Glow, better than using the spherical constraint or the orthogonal reparameterization. We also demonstrate the efficacy of Gaussianization layers when the forward model is inaccurate, in which case the induced error in data is not Gaussian (\appref{append:additional_exp}). 

\paragraph{Ablation study}
\tabref{tab:deblurring_results} also shows that the two parameterization schemes (\appref{append:reparam_schemes}) for Glow using the Gaussianization layers have similar performance. Besides, we report the ablation study on the components of the Gaussianizaiton layers for Glow (\appref{append:additional_exp}).

\subsection{Eikonal tomography using StyleGAN2}
In acoustic wave imaging (e.g., ultrasound tomography), we excite waves using sparsely distributed sources one at a time at the boundary of the object. Then we reconstruct its internal structures (the spatial distribution of wave speed) given the first-arrival travel time recorded on the boundary. The following eikonal equation approximately describes the shortest travel time \(T(\vx;\vx_s)\) that the acoustic wave emerging from the source location \(\vx_s\) takes to reach location \(\vx\) inside the target object~\citep{yilmaz2001seismic}:
\begin{equation}
	|\nabla T\left( \vx; \vx_s \right)| = 1/ c\left( \vx \right), \quad T(\vx_s;\vx_s) = 0,
\end{equation}
where \(c(\vx)\) is the wave propagation speed at each location. Both this eikonal PDE and the implicitly defined forward mapping \(c(\vx)\rightarrow T(\vx)\) are nonlinear, and there has been little research on DGM-regularized inverse problems with such nonlinear characteristics. The inverse problem is severely ill-posed, which is equivalent to a curved-ray tomography problem. See \appref{append:eikonal_tomography} for more background information. We added noise to the recorded travel time using the following formula: \(T_{\text{noisy}}(\vx_r;\vx_s) = T(\vx_r;\vx_s) (1 + \epsilon) \), where \(\epsilon \sim \mathcal{N}(0,0.001^2)\) and \(\vx_r\) denotes any receiver location. In other words, longer traveltime corresponds to larger uncertainties.

We show that the Gaussianization layers outperformed other methods in this tomography task in \tabref{tab:eikonal_results} and \figref{fig:comparison_tomo}. Note that this is a statistical conclusion. We also report an example where the spherical constraint works better than the Gaussianization layers (bottom right).

\begin{table}[htpb]
\centering
\makebox[0pt][c]{\parbox{\textwidth}{%
    \begin{minipage}[t]{0.5\textwidth}\centering
      \setlength{\extrarowheight}{1pt}
      \caption{Deblurring results using Glow.}
      \label{tab:deblurring_results}
      \vspace{0.2em}
      \centering
      \small

      \begin{tabular}{l@{~~~}|c@{~~~}c@{~~~}c@{~~~}}
      \toprule
        Method      & LPIPS$\downarrow$ & PSNR$\uparrow$ & SSIM$\uparrow$ \\
      \hline
      
      Spherical  & 0.17\tiny{$\pm$0.06} & 21.78\tiny{$\pm$1.14} & 0.580\tiny{$\pm$0.066} \\

      Orthogonal   & 0.16\tiny{$\pm$0.06} & 21.91\tiny{$\pm$1.31} & 0.583\tiny{$\pm$0.063} \\
      
      G layers P1  & 0.13\tiny{$\pm$0.05} & 22.40\tiny{$\pm$1.34} & 0.583\tiny{$\pm$0.069} \\
      
      G layers P2  & \best{0.13\tiny{$\pm$0.05}} & \best{22.47\tiny{$\pm$1.27}} & \best{0.590\tiny{$\pm$0.064}} \\
      \bottomrule
      \end{tabular}
    
    \end{minipage}
    \hfill
    \begin{minipage}[t]{0.5\textwidth}\centering
      \setlength{\extrarowheight}{1pt}
      \caption{Eikonal tomography using StyleGAN2}
      \label{tab:eikonal_results}
      \vspace{0.2em}
      \centering
      \small
      \begin{tabular}{l@{~~~}|c@{~~~}c@{~~~}}
      \toprule
        Method      & PSNR$\uparrow$ & SSIM$\uparrow$ \\
      \hline
      
      %  Baseline  & 22.3033\tiny{$\pm$4.0239} & 0.6912\tiny{$\pm$0.1485} \\
      TV  & 20.66\tiny{$\pm$1.21} & 0.543\tiny{$\pm$0.073} \\
      Spherical  & 22.19\tiny{$\pm$4.05} & 0.686\tiny{$\pm$0.144} \\
      G layers   & \best{24.80\tiny{$\pm$2.55}} & \best{0.803\tiny{$\pm$0.093}} \\
      \bottomrule
      \end{tabular}
    \end{minipage}
}}
\end{table}

\begingroup
\setlength{\tabcolsep}{1.2pt}
\begin{figure}[htpb]
    \centering%
    \newcommand{\sizeI}{0.11}
    \newcommand{\imgTrue}[1]{\includegraphics[width=\sizeI\linewidth]{figures/figures_inv_mri/tv_accl8x_snr20.0/#1/True.jpg}}
    \newcommand{\imgTV}[1]{\includegraphics[width=\sizeI\linewidth]{figures/figures_inv_tomo/tv_noisestd0.001_stride36_tv_beta10.0/#1/Inv.jpg}}
    \newcommand{\imgSP}[1]{\includegraphics[width=\sizeI\linewidth]{figures/figures_inv_tomo/sphere_noisestd0.001_stride36/#1/Inv.jpg}}
    \newcommand{\imgGauss}[1]{\includegraphics[width=\sizeI\linewidth]{figures/figures_inv_tomo/gaus_noisestd0.001_stride36_1_0_1_1/#1/Inv.jpg}}
  \begin{tabular}{c | c}

	\begin{tabular}{cccc}
	\small{True} & \small{TV} & \small{Spherical} & \small{G layers}\\
	\imgTrue{mri_104} & \imgTV{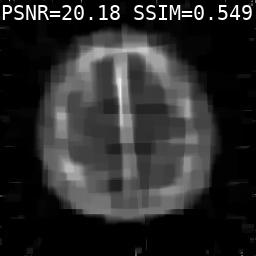} &\imgSP{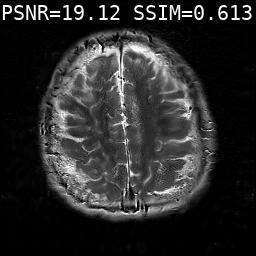} & \imgGauss{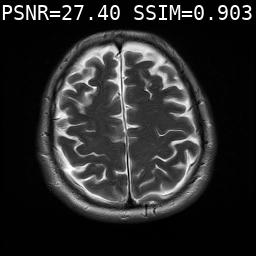} \\
	\imgTrue{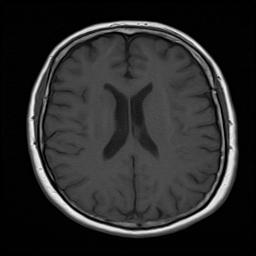} & \imgTV{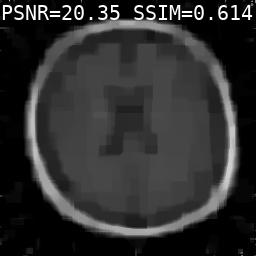} &\imgSP{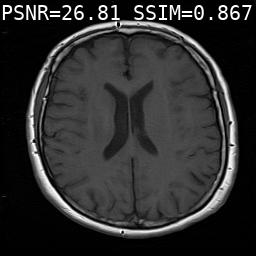} & \imgGauss{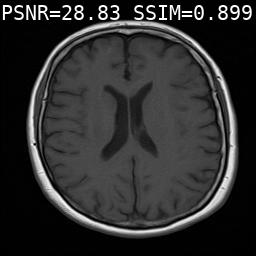} \\
	\end{tabular} 
  &
	\begin{tabular}{cccc}
	\small{True} & \small{TV} & \small{Spherical} & \small{G layers}\\
	\imgTrue{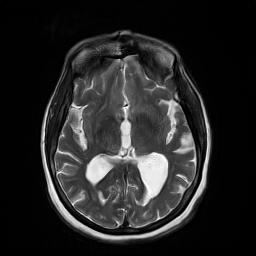} & \imgTV{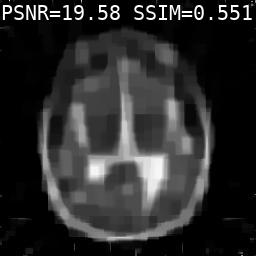} &\imgSP{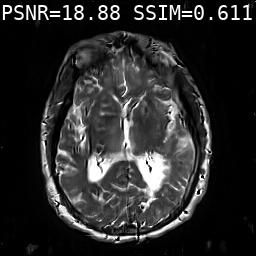} & \imgGauss{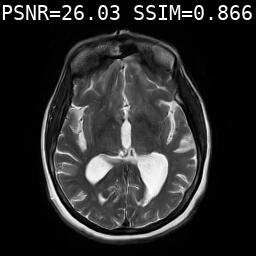} \\
	\imgTrue{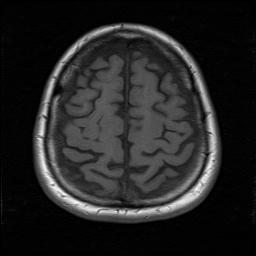} & \imgTV{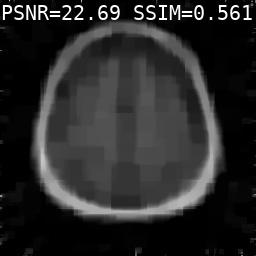} &\imgSP{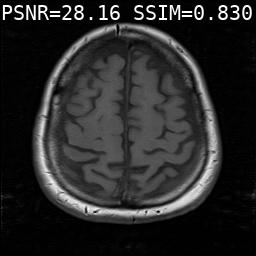} & \imgGauss{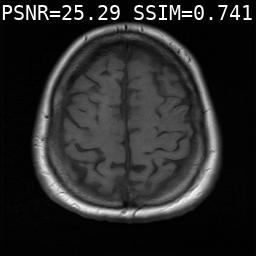} \\
	\end{tabular} \\

\end{tabular}

	\caption{Comparison of eikonal tomography results from different methods.}
	\label{fig:comparison_tomo}
\end{figure}
\endgroup

\section{Discussion and conclusions}
\label{sec:conclusions}
We provide insights on the experiment results and further discuss this work's limitations, computational cost, and broader impact in \appref{append:additional_discussions}.

In summary, we have identified a critical problem in DGM-regularized inversion: the latent tensor can deviate from a typical example from the desired high-dimensional standard Gaussian distribution, leading to unsatisfactory inversion results. To address the problem, we have introduced the differentiable Gaussianization layers that reparameterize and Gaussianize latent tensors so that the inversion results remain plausible. In general, our method has achieved the best scores in terms of mean values and standard deviations compared with other methods, demonstrating our method's advantages and high performance in terms of accuracy and consistency. (Regarding the standard deviation of scores, we only compare with other methods with competitive mean scores, such as orthogonal reparameterization and spherical constraint.) Our proposed layers are plug-and-play, require minimal parameter tuning, and can be applied to various deep generative models and inverse problems.

\section*{Acknowledgements}
The author would like to thank Huseyin Denli, Ashutosh Tewari, Myun-Seok Cheon, Di Du, Stuart Harwood, Yu Fan, and Qiuzi Li for helpful discussions and the anonymous reviewers for their constructive feedback, which greatly improved the paper.

\bibliography{refs}

\begin{thebibliography}{85}
\providecommand{\natexlab}[1]{#1}
\providecommand{\url}[1]{\texttt{#1}}
\expandafter\ifx\csname urlstyle\endcsname\relax
  \providecommand{\doi}[1]{doi: #1}\else
  \providecommand{\doi}{doi: \begingroup \urlstyle{rm}\Url}\fi

\bibitem[Akiyama et~al.(2019)Akiyama, Alberdi, Alef, Asada, Azulay, Baczko,
  Ball, Balokovi{\'c}, Barrett, Bintley, et~al.]{akiyama2019first}
Kazunori Akiyama, Antxon Alberdi, Walter Alef, Keiichi Asada, Rebecca Azulay,
  Anne-Kathrin Baczko, David Ball, Mislav Balokovi{\'c}, John Barrett, Dan
  Bintley, et~al.
\newblock First m87 event horizon telescope results. iv. imaging the central
  supermassive black hole.
\newblock \emph{The Astrophysical Journal Letters}, 875\penalty0 (1):\penalty0
  L4, 2019.

\bibitem[Ardizzone et~al.(2018)Ardizzone, Kruse, Wirkert, Rahner, Pellegrini,
  Klessen, Maier-Hein, Rother, and K{\"o}the]{ardizzone2018analyzing}
Lynton Ardizzone, Jakob Kruse, Sebastian Wirkert, Daniel Rahner, Eric~W
  Pellegrini, Ralf~S Klessen, Lena Maier-Hein, Carsten Rother, and Ullrich
  K{\"o}the.
\newblock Analyzing inverse problems with invertible neural networks.
\newblock \emph{arXiv preprint arXiv:1808.04730}, 2018.

\bibitem[Asim et~al.(2020)Asim, Daniels, Leong, Ahmed, and
  Hand]{asim2020invertible}
Muhammad Asim, Max Daniels, Oscar Leong, Ali Ahmed, and Paul Hand.
\newblock Invertible generative models for inverse problems: mitigating
  representation error and dataset bias.
\newblock In \emph{International Conference on Machine Learning}, pp.\
  399--409. PMLR, 2020.

\bibitem[Beck \& Teboulle(2009)Beck and Teboulle]{beck2009fast}
Amir Beck and Marc Teboulle.
\newblock A fast iterative shrinkage-thresholding algorithm for linear inverse
  problems.
\newblock \emph{SIAM journal on imaging sciences}, 2\penalty0 (1):\penalty0
  183--202, 2009.

\bibitem[Blum et~al.(2020)Blum, Hopcroft, and Kannan]{blum2020foundations}
Avrim Blum, John Hopcroft, and Ravindran Kannan.
\newblock \emph{Foundations of data science}.
\newblock Cambridge University Press, 2020.

\bibitem[Bojanowski et~al.(2017)Bojanowski, Joulin, Lopez-Paz, and
  Szlam]{bojanowski2017optimizing}
Piotr Bojanowski, Armand Joulin, David Lopez-Paz, and Arthur Szlam.
\newblock Optimizing the latent space of generative networks.
\newblock \emph{arXiv preprint arXiv:1707.05776}, 2017.

\bibitem[Bora et~al.(2017)Bora, Jalal, Price, and Dimakis]{bora2017compressed}
Ashish Bora, Ajil Jalal, Eric Price, and Alexandros~G Dimakis.
\newblock Compressed sensing using generative models.
\newblock In \emph{International Conference on Machine Learning}, pp.\
  537--546. PMLR, 2017.

\bibitem[Brent(2013)]{brent2013algorithms}
Richard~P Brent.
\newblock \emph{Algorithms for minimization without derivatives}.
\newblock Courier Corporation, 2013.

\bibitem[Chen et~al.(2017)Chen, Zhang, Kalra, Lin, Chen, Liao, Zhou, and
  Wang]{chen2017low}
Hu~Chen, Yi~Zhang, Mannudeep~K Kalra, Feng Lin, Yang Chen, Peixi Liao, Jiliu
  Zhou, and Ge~Wang.
\newblock Low-dose ct with a residual encoder-decoder convolutional neural
  network.
\newblock \emph{IEEE transactions on medical imaging}, 36\penalty0
  (12):\penalty0 2524--2535, 2017.

\bibitem[Chen \& Gopinath(2000)Chen and Gopinath]{chen2000gaussianization}
Scott Chen and Ramesh Gopinath.
\newblock Gaussianization.
\newblock \emph{Advances in neural information processing systems},
  13:\penalty0 423--429, 2000.

\bibitem[Cheng et~al.(2022)Cheng, Lin, Lee, Ren, Tulyakov, and
  Yang]{cheng2022inout}
Yen-Chi Cheng, Chieh~Hubert Lin, Hsin-Ying Lee, Jian Ren, Sergey Tulyakov, and
  Ming-Hsuan Yang.
\newblock Inout: Diverse image outpainting via gan inversion.
\newblock In \emph{Proceedings of the IEEE/CVF Conference on Computer Vision
  and Pattern Recognition}, pp.\  11431--11440, 2022.

\bibitem[Corless et~al.(1996)Corless, Gonnet, Hare, Jeffrey, and
  Knuth]{corless1996lambertw}
Robert~M Corless, Gaston~H Gonnet, David~EG Hare, David~J Jeffrey, and Donald~E
  Knuth.
\newblock On the lambertw function.
\newblock \emph{Advances in Computational mathematics}, 5\penalty0
  (1):\penalty0 329--359, 1996.

\bibitem[Cover \& Thomas(2012)Cover and Thomas]{cover2012elements}
Thomas~M Cover and Joy~A Thomas.
\newblock \emph{Elements of Information Theory}.
\newblock Wiley, 2012.
\newblock ISBN 9781118585771.
\newblock URL \url{https://books.google.com/books?id=VWq5GG6ycxMC}.

\bibitem[Daras et~al.(2021)Daras, Dean, Jalal, and
  Dimakis]{daras2021intermediate}
Giannis Daras, Joseph Dean, Ajil Jalal, and Alexandros~G Dimakis.
\newblock Intermediate layer optimization for inverse problems using deep
  generative models.
\newblock \emph{arXiv preprint arXiv:2102.07364}, 2021.

\bibitem[Daras et~al.(2022)Daras, Dagan, Dimakis, and
  Daskalakis]{daras2022score}
Giannis Daras, Yuval Dagan, Alexandros~G Dimakis, and Constantinos Daskalakis.
\newblock Score-guided intermediate layer optimization: Fast langevin mixing
  for inverse problem.
\newblock \emph{arXiv preprint arXiv:2206.09104}, 2022.

\bibitem[Dinh et~al.(2014)Dinh, Krueger, and Bengio]{dinh2014nice}
Laurent Dinh, David Krueger, and Yoshua Bengio.
\newblock Nice: Non-linear independent components estimation.
\newblock \emph{arXiv preprint arXiv:1410.8516}, 2014.

\bibitem[Dinh et~al.(2016)Dinh, Sohl-Dickstein, and Bengio]{dinh2016density}
Laurent Dinh, Jascha Sohl-Dickstein, and Samy Bengio.
\newblock Density estimation using real nvp.
\newblock \emph{arXiv preprint arXiv:1605.08803}, 2016.

\bibitem[Gemmeke \& Ruiter(2007)Gemmeke and Ruiter]{gemmeke20073d}
H~Gemmeke and NV~Ruiter.
\newblock 3d ultrasound computer tomography for medical imaging.
\newblock \emph{Nuclear Instruments and Methods in Physics Research Section A:
  Accelerators, Spectrometers, Detectors and Associated Equipment},
  580\penalty0 (2):\penalty0 1057--1065, 2007.

\bibitem[Goerg(2015)]{goerg2015lambert}
Georg~M Goerg.
\newblock The lambert way to gaussianize heavy-tailed data with the inverse of
  tukey’sh transformation as a special case.
\newblock \emph{The Scientific World Journal}, 2015, 2015.

\bibitem[Goodfellow et~al.(2014)Goodfellow, Pouget-Abadie, Mirza, Xu,
  Warde-Farley, Ozair, Courville, and Bengio]{goodfellow2014generative}
Ian~J Goodfellow, Jean Pouget-Abadie, Mehdi Mirza, Bing Xu, David Warde-Farley,
  Sherjil Ozair, Aaron Courville, and Yoshua Bengio.
\newblock Generative adversarial networks.
\newblock \emph{arXiv preprint arXiv:1406.2661}, 2014.

\bibitem[Gu et~al.(2020)Gu, Shen, and Zhou]{gu2020image}
Jinjin Gu, Yujun Shen, and Bolei Zhou.
\newblock Image processing using multi-code gan prior.
\newblock In \emph{Proceedings of the IEEE/CVF conference on computer vision
  and pattern recognition}, pp.\  3012--3021, 2020.

\bibitem[Hand et~al.(2018)Hand, Leong, and Voroninski]{hand2018phase}
Paul Hand, Oscar Leong, and Vladislav Voroninski.
\newblock Phase retrieval under a generative prior.
\newblock \emph{arXiv preprint arXiv:1807.04261}, 2018.

\bibitem[H{\"o}gbom(1974)]{hogbom1974aperture}
JA~H{\"o}gbom.
\newblock Aperture synthesis with a non-regular distribution of interferometer
  baselines.
\newblock \emph{Astronomy and Astrophysics Supplement Series}, 15:\penalty0
  417, 1974.

\bibitem[Huang et~al.(2018)Huang, Yang, Lang, and Deng]{huang2018decorrelated}
Lei Huang, Dawei Yang, Bo~Lang, and Jia Deng.
\newblock Decorrelated batch normalization.
\newblock In \emph{Proceedings of the IEEE Conference on Computer Vision and
  Pattern Recognition}, pp.\  791--800, 2018.

\bibitem[Hyvarinen(1999)]{hyvarinen1999fast}
Aapo Hyvarinen.
\newblock Fast and robust fixed-point algorithms for independent component
  analysis.
\newblock \emph{IEEE transactions on Neural Networks}, 10\penalty0
  (3):\penalty0 626--634, 1999.

\bibitem[Hyv{\"a}rinen(1999)]{hyvarinen1999fixed}
Aapo Hyv{\"a}rinen.
\newblock The fixed-point algorithm and maximum likelihood estimation for
  independent component analysis.
\newblock \emph{Neural Processing Letters}, 10\penalty0 (1):\penalty0 1--5,
  1999.

\bibitem[Hyv{\"a}rinen \& Oja(2000)Hyv{\"a}rinen and
  Oja]{hyvarinen2000independent}
Aapo Hyv{\"a}rinen and Erkki Oja.
\newblock Independent component analysis: algorithms and applications.
\newblock \emph{Neural networks}, 13\penalty0 (4-5):\penalty0 411--430, 2000.

\bibitem[Jalal et~al.(2021)Jalal, Arvinte, Daras, Price, Dimakis, and
  Tamir]{jalal2021robust}
Ajil Jalal, Marius Arvinte, Giannis Daras, Eric Price, Alexandros~G Dimakis,
  and Jon Tamir.
\newblock Robust compressed sensing mri with deep generative priors.
\newblock \emph{Advances in Neural Information Processing Systems},
  34:\penalty0 14938--14954, 2021.

\bibitem[Jin et~al.(2017)Jin, McCann, Froustey, and Unser]{jin2017deep}
Kyong~Hwan Jin, Michael~T McCann, Emmanuel Froustey, and Michael Unser.
\newblock Deep convolutional neural network for inverse problems in imaging.
\newblock \emph{IEEE Transactions on Image Processing}, 26\penalty0
  (9):\penalty0 4509--4522, 2017.

\bibitem[Karras et~al.(2018)Karras, Aila, Laine, and
  Lehtinen]{karras2017progressive}
Tero Karras, Timo Aila, Samuli Laine, and Jaakko Lehtinen.
\newblock Progressive growing of gans for improved quality, stability, and
  variation.
\newblock 2018.

\bibitem[Karras et~al.(2020)Karras, Laine, Aittala, Hellsten, Lehtinen, and
  Aila]{karras2020analyzing}
Tero Karras, Samuli Laine, Miika Aittala, Janne Hellsten, Jaakko Lehtinen, and
  Timo Aila.
\newblock Analyzing and improving the image quality of stylegan.
\newblock In \emph{Proceedings of the IEEE/CVF conference on computer vision
  and pattern recognition}, pp.\  8110--8119, 2020.

\bibitem[Kawar et~al.(2021)Kawar, Vaksman, and Elad]{kawar2021snips}
Bahjat Kawar, Gregory Vaksman, and Michael Elad.
\newblock Snips: Solving noisy inverse problems stochastically.
\newblock \emph{Advances in Neural Information Processing Systems},
  34:\penalty0 21757--21769, 2021.

\bibitem[Kawar et~al.(2022)Kawar, Elad, Ermon, and Song]{kawar2022denoising}
Bahjat Kawar, Michael Elad, Stefano Ermon, and Jiaming Song.
\newblock Denoising diffusion restoration models.
\newblock \emph{arXiv preprint arXiv:2201.11793}, 2022.

\bibitem[Kelkar \& Anastasio(2021)Kelkar and Anastasio]{kelkar2021prior}
Varun~A Kelkar and Mark Anastasio.
\newblock Prior image-constrained reconstruction using style-based generative
  models.
\newblock In \emph{International Conference on Machine Learning}, pp.\
  5367--5377. PMLR, 2021.

\bibitem[Kingma \& Ba(2015)Kingma and Ba]{kingma2015adam}
Diederik Kingma and Jimmy Ba.
\newblock Adam, a method for stochastic optimization.
\newblock 2015.

\bibitem[Kingma \& Dhariwal(2018)Kingma and Dhariwal]{kingma2018glow}
Diederik~P Kingma and Prafulla Dhariwal.
\newblock Glow: Generative flow with invertible 1x1 convolutions.
\newblock \emph{arXiv preprint arXiv:1807.03039}, 2018.

\bibitem[Kingma \& Welling(2013)Kingma and Welling]{kingma2013auto}
Diederik~P Kingma and Max Welling.
\newblock Auto-encoding variational bayes.
\newblock \emph{arXiv preprint arXiv:1312.6114}, 2013.

\bibitem[Kingma et~al.(2016)Kingma, Salimans, Jozefowicz, Chen, Sutskever, and
  Welling]{kingma2016improving}
Diederik~P Kingma, Tim Salimans, Rafal Jozefowicz, Xi~Chen, Ilya Sutskever, and
  Max Welling.
\newblock Improving variational inference with inverse autoregressive flow.
\newblock \emph{arXiv preprint arXiv:1606.04934}, 2016.

\bibitem[Knoll et~al.(2020)Knoll, Zbontar, Sriram, Muckley, Bruno, Defazio,
  Parente, Geras, Katsnelson, Chandarana, et~al.]{knoll2020fastmri}
Florian Knoll, Jure Zbontar, Anuroop Sriram, Matthew~J Muckley, Mary Bruno,
  Aaron Defazio, Marc Parente, Krzysztof~J Geras, Joe Katsnelson, Hersh
  Chandarana, et~al.
\newblock fastmri: A publicly available raw k-space and dicom dataset of knee
  images for accelerated mr image reconstruction using machine learning.
\newblock \emph{Radiology: Artificial Intelligence}, 2\penalty0 (1):\penalty0
  e190007, 2020.

\bibitem[LaMontagne et~al.(2019)LaMontagne, Benzinger, Morris, Keefe, Hornbeck,
  Xiong, Grant, Hassenstab, Moulder, Vlassenko, et~al.]{lamontagne2019oasis}
Pamela~J LaMontagne, Tammie~LS Benzinger, John~C Morris, Sarah Keefe, Russ
  Hornbeck, Chengjie Xiong, Elizabeth Grant, Jason Hassenstab, Krista Moulder,
  Andrei~G Vlassenko, et~al.
\newblock Oasis-3: longitudinal neuroimaging, clinical, and cognitive dataset
  for normal aging and alzheimer disease.
\newblock \emph{MedRxiv}, 2019.

\bibitem[Laparra et~al.(2011)Laparra, Camps-Valls, and
  Malo]{laparra2011iterative}
Valero Laparra, Gustavo Camps-Valls, and Jes{\'u}s Malo.
\newblock Iterative gaussianization: from ica to random rotations.
\newblock \emph{IEEE transactions on neural networks}, 22\penalty0
  (4):\penalty0 537--549, 2011.

\bibitem[Lauterbur(1973)]{lauterbur1973image}
Paul~C Lauterbur.
\newblock Image formation by induced local interactions: examples employing
  nuclear magnetic resonance.
\newblock \emph{nature}, 242\penalty0 (5394):\penalty0 190--191, 1973.

\bibitem[Li et~al.(2020)Li, Xu, Harris, and Darve]{li2020coupled}
Dongzhuo Li, Kailai Xu, Jerry~M Harris, and Eric Darve.
\newblock Coupled time-lapse full-waveform inversion for subsurface flow
  problems using intrusive automatic differentiation.
\newblock \emph{Water Resources Research}, 56\penalty0 (8):\penalty0
  e2019WR027032, 2020.

\bibitem[Li et~al.(2021)Li, Denli, MacDonald, Basler-Reeder, Baumstein, and
  Daves]{li2021multiparameter}
Dongzhuo Li, Huseyin Denli, Cody MacDonald, Kyle Basler-Reeder, Anatoly
  Baumstein, and Jacquelyn Daves.
\newblock Multiparameter geophysical reservoir characterization augmented by
  generative networks.
\newblock In \emph{First International Meeting for Applied Geoscience \&
  Energy}, pp.\  1364--1368. Society of Exploration Geophysicists, 2021.

\bibitem[Liang et~al.(2021)Liang, Zhang, Gu, Van~Gool, and
  Timofte]{liang2021flow}
Jingyun Liang, Kai Zhang, Shuhang Gu, Luc Van~Gool, and Radu Timofte.
\newblock Flow-based kernel prior with application to blind super-resolution.
\newblock In \emph{Proceedings of the IEEE/CVF Conference on Computer Vision
  and Pattern Recognition}, pp.\  10601--10610, 2021.

\bibitem[Lines \& Treitel(1984)Lines and Treitel]{lines1984review}
LR~Lines and S~Treitel.
\newblock A review of least-squares inversion and its application to
  geophysical problems.
\newblock \emph{Geophysical prospecting}, 32\penalty0 (2):\penalty0 159--186,
  1984.

\bibitem[Liu et~al.(2021)Liu, Lin, Liu, Rehg, Paull, Xiong, Song, and
  Weller]{liu2021orthogonal}
Weiyang Liu, Rongmei Lin, Zhen Liu, James~M Rehg, Liam Paull, Li~Xiong,
  Le~Song, and Adrian Weller.
\newblock Orthogonal over-parameterized training.
\newblock In \emph{Proceedings of the IEEE/CVF Conference on Computer Vision
  and Pattern Recognition}, pp.\  7251--7260, 2021.

\bibitem[Liu et~al.(2015)Liu, Luo, Wang, and Tang]{liu2015faceattributes}
Ziwei Liu, Ping Luo, Xiaogang Wang, and Xiaoou Tang.
\newblock Deep learning face attributes in the wild.
\newblock In \emph{Proceedings of International Conference on Computer Vision
  (ICCV)}, December 2015.

\bibitem[Lugmayr et~al.(2020)Lugmayr, Danelljan, Van~Gool, and
  Timofte]{lugmayr2020srflow}
Andreas Lugmayr, Martin Danelljan, Luc Van~Gool, and Radu Timofte.
\newblock Srflow: Learning the super-resolution space with normalizing flow.
\newblock In \emph{European Conference on Computer Vision}, pp.\  715--732.
  Springer, 2020.

\bibitem[Lustig et~al.(2007)Lustig, Donoho, and Pauly]{lustig2007sparse}
Michael Lustig, David Donoho, and John~M Pauly.
\newblock Sparse mri: The application of compressed sensing for rapid mr
  imaging.
\newblock \emph{Magnetic Resonance in Medicine: An Official Journal of the
  International Society for Magnetic Resonance in Medicine}, 58\penalty0
  (6):\penalty0 1182--1195, 2007.

\bibitem[Mardani et~al.(2018)Mardani, Gong, Cheng, Vasanawala, Zaharchuk, Xing,
  and Pauly]{mardani2018deep}
Morteza Mardani, Enhao Gong, Joseph~Y Cheng, Shreyas~S Vasanawala, Greg
  Zaharchuk, Lei Xing, and John~M Pauly.
\newblock Deep generative adversarial neural networks for compressive sensing
  mri.
\newblock \emph{IEEE transactions on medical imaging}, 38\penalty0
  (1):\penalty0 167--179, 2018.

\bibitem[Marinescu et~al.(2020)Marinescu, Moyer, and
  Golland]{marinescu2020bayesian}
Razvan~V Marinescu, Daniel Moyer, and Polina Golland.
\newblock Bayesian image reconstruction using deep generative models.
\newblock \emph{arXiv preprint arXiv:2012.04567}, 2020.

\bibitem[Meng et~al.(2020)Meng, Song, Song, and Ermon]{meng2020gaussianization}
Chenlin Meng, Yang Song, Jiaming Song, and Stefano Ermon.
\newblock Gaussianization flows.
\newblock In \emph{International Conference on Artificial Intelligence and
  Statistics}, pp.\  4336--4345. PMLR, 2020.

\bibitem[Mosser et~al.(2020)Mosser, Dubrule, and Blunt]{mosser2020stochastic}
Lukas Mosser, Olivier Dubrule, and Martin~J Blunt.
\newblock Stochastic seismic waveform inversion using generative adversarial
  networks as a geological prior.
\newblock \emph{Mathematical Geosciences}, 52\penalty0 (1):\penalty0 53--79,
  2020.

\bibitem[Nalisnick et~al.(2019)Nalisnick, Matsukawa, Teh, Gorur, and
  Lakshminarayanan]{nalisnick2018do}
Eric Nalisnick, Akihiro Matsukawa, Yee~Whye Teh, Dilan Gorur, and Balaji
  Lakshminarayanan.
\newblock Do deep generative models know what they don't know?
\newblock In \emph{ICLR}, 2019.
\newblock URL \url{https://openreview.net/forum?id=H1xwNhCcYm}.

\bibitem[Nocedal \& Wright(2006)Nocedal and Wright]{nocedal2006numerical}
Jorge Nocedal and Stephen Wright.
\newblock \emph{Numerical optimization}.
\newblock Springer Science \& Business Media, 2006.

\bibitem[Oja \& Yuan(2006)Oja and Yuan]{oja2006fastica}
Erkki Oja and Zhijian Yuan.
\newblock The fastica algorithm revisited: Convergence analysis.
\newblock \emph{IEEE transactions on Neural Networks}, 17\penalty0
  (6):\penalty0 1370--1381, 2006.

\bibitem[Ongie et~al.(2020)Ongie, Jalal, Metzler, Baraniuk, Dimakis, and
  Willett]{ongie2020deep}
Gregory Ongie, Ajil Jalal, Christopher~A Metzler, Richard~G Baraniuk,
  Alexandros~G Dimakis, and Rebecca Willett.
\newblock Deep learning techniques for inverse problems in imaging.
\newblock \emph{IEEE Journal on Selected Areas in Information Theory},
  1\penalty0 (1):\penalty0 39--56, 2020.

\bibitem[Papamakarios et~al.(2017)Papamakarios, Pavlakou, and
  Murray]{papamakarios2017masked}
George Papamakarios, Theo Pavlakou, and Iain Murray.
\newblock Masked autoregressive flow for density estimation.
\newblock \emph{arXiv preprint arXiv:1705.07057}, 2017.

\bibitem[Rabanser et~al.(2019)Rabanser, G{\"u}nnemann, and
  Lipton]{rabanser2019failing}
Stephan Rabanser, Stephan G{\"u}nnemann, and Zachary Lipton.
\newblock Failing loudly: An empirical study of methods for detecting dataset
  shift.
\newblock \emph{Advances in Neural Information Processing Systems}, 32, 2019.

\bibitem[Rombach et~al.(2022)Rombach, Blattmann, Lorenz, Esser, and
  Ommer]{Rombach_2022_CVPR}
Robin Rombach, Andreas Blattmann, Dominik Lorenz, Patrick Esser, and Bj\"orn
  Ommer.
\newblock High-resolution image synthesis with latent diffusion models.
\newblock In \emph{Proceedings of the IEEE/CVF Conference on Computer Vision
  and Pattern Recognition (CVPR)}, pp.\  10684--10695, June 2022.

\bibitem[Saad \& Schultz(1986)Saad and Schultz]{saad1986gmres}
Youcef Saad and Martin~H Schultz.
\newblock Gmres: A generalized minimal residual algorithm for solving
  nonsymmetric linear systems.
\newblock \emph{SIAM Journal on scientific and statistical computing},
  7\penalty0 (3):\penalty0 856--869, 1986.

\bibitem[Scarpace et~al.(2016)Scarpace, Mikkelsen, Cha, Rao, Tekchandani,
  Gutman, and Pierce]{scarpace2016radiology}
Lisa Scarpace, L~Mikkelsen, T~Cha, Sujaya Rao, Sangeeta Tekchandani, S~Gutman,
  and D~Pierce.
\newblock Radiology data from the cancer genome atlas glioblastoma multiforme
  [tcga-gbm] collection.
\newblock \emph{The Cancer Imaging Archive}, 11\penalty0 (4):\penalty0 1, 2016.

\bibitem[Shen et~al.(2020)Shen, Gu, Tang, and Zhou]{shen2020interpreting}
Yujun Shen, Jinjin Gu, Xiaoou Tang, and Bolei Zhou.
\newblock Interpreting the latent space of gans for semantic face editing.
\newblock In \emph{Proceedings of the IEEE/CVF conference on computer vision
  and pattern recognition}, pp.\  9243--9252, 2020.

\bibitem[Siahkoohi et~al.(2021)Siahkoohi, Rizzuti, Louboutin, Witte, and
  Herrmann]{siahkoohi2021preconditioned}
Ali Siahkoohi, Gabrio Rizzuti, Mathias Louboutin, Philipp~A Witte, and Felix~J
  Herrmann.
\newblock Preconditioned training of normalizing flows for variational
  inference in inverse problems.
\newblock \emph{arXiv preprint arXiv:2101.03709}, 2021.

\bibitem[Siarohin et~al.(2018)Siarohin, Sangineto, and
  Sebe]{siarohin2018whitening}
Aliaksandr Siarohin, Enver Sangineto, and Nicu Sebe.
\newblock Whitening and coloring batch transform for gans.
\newblock In \emph{International Conference on Learning Representations}, 2018.

\bibitem[Song et~al.(2021)Song, Shen, Xing, and Ermon]{song2021solving}
Yang Song, Liyue Shen, Lei Xing, and Stefano Ermon.
\newblock Solving inverse problems in medical imaging with score-based
  generative models.
\newblock \emph{arXiv preprint arXiv:2111.08005}, 2021.

\bibitem[Sriram et~al.(2020)Sriram, Zbontar, Murrell, Defazio, Zitnick,
  Yakubova, Knoll, and Johnson]{sriram2020end}
Anuroop Sriram, Jure Zbontar, Tullie Murrell, Aaron Defazio, C~Lawrence
  Zitnick, Nafissa Yakubova, Florian Knoll, and Patricia Johnson.
\newblock End-to-end variational networks for accelerated mri reconstruction.
\newblock In \emph{International Conference on Medical Image Computing and
  Computer-Assisted Intervention}, pp.\  64--73. Springer, 2020.

\bibitem[Starck et~al.(2002)Starck, Pantin, and
  Murtagh]{starck2002deconvolution}
Jean-Luc Starck, Eric Pantin, and Fionn Murtagh.
\newblock Deconvolution in astronomy: A review.
\newblock \emph{Publications of the Astronomical Society of the Pacific},
  114\penalty0 (800):\penalty0 1051, 2002.

\bibitem[Tarantola(1984)]{tarantola1984inversion}
Albert Tarantola.
\newblock Inversion of seismic reflection data in the acoustic approximation.
\newblock \emph{Geophysics}, 49\penalty0 (8):\penalty0 1259--1266, 1984.

\bibitem[Tromp et~al.(2005)Tromp, Tape, and Liu]{tromp2005seismic}
Jeroen Tromp, Carl Tape, and Qinya Liu.
\newblock Seismic tomography, adjoint methods, time reversal and
  banana-doughnut kernels.
\newblock \emph{Geophysical Journal International}, 160\penalty0 (1):\penalty0
  195--216, 2005.

\bibitem[Van~Veen et~al.(2018)Van~Veen, Jalal, Soltanolkotabi, Price,
  Vishwanath, and Dimakis]{van2018compressed}
Dave Van~Veen, Ajil Jalal, Mahdi Soltanolkotabi, Eric Price, Sriram Vishwanath,
  and Alexandros~G Dimakis.
\newblock Compressed sensing with deep image prior and learned regularization.
\newblock \emph{arXiv preprint arXiv:1806.06438}, 2018.

\bibitem[Virieux \& Operto(2009)Virieux and Operto]{virieux2009overview}
Jean Virieux and St{\'e}phane Operto.
\newblock An overview of full-waveform inversion in exploration geophysics.
\newblock \emph{Geophysics}, 74\penalty0 (6):\penalty0 WCC1--WCC26, 2009.

\bibitem[Wang et~al.(2018)Wang, Yu, Wu, Gu, Liu, Dong, Qiao, and
  Change~Loy]{wang2018esrgan}
Xintao Wang, Ke~Yu, Shixiang Wu, Jinjin Gu, Yihao Liu, Chao Dong, Yu~Qiao, and
  Chen Change~Loy.
\newblock Esrgan: Enhanced super-resolution generative adversarial networks.
\newblock In \emph{Proceedings of the European conference on computer vision
  (ECCV) workshops}, pp.\  0--0, 2018.

\bibitem[Wang et~al.(2004)Wang, Bovik, Sheikh, and Simoncelli]{wang2004image}
Zhou Wang, Alan~C Bovik, Hamid~R Sheikh, and Eero~P Simoncelli.
\newblock Image quality assessment: from error visibility to structural
  similarity.
\newblock \emph{IEEE transactions on image processing}, 13\penalty0
  (4):\penalty0 600--612, 2004.

\bibitem[Whang et~al.(2021)Whang, Lei, and Dimakis]{pmlr-v139-whang21a}
Jay Whang, Qi~Lei, and Alex Dimakis.
\newblock Solving inverse problems with a flow-based noise model.
\newblock In \emph{Proceedings of the 38th International Conference on Machine
  Learning}, volume 139 of \emph{Proceedings of Machine Learning Research},
  pp.\  11146--11157. PMLR, 18--24 Jul 2021.
\newblock URL \url{https://proceedings.mlr.press/v139/whang21a.html}.

\bibitem[White(2016)]{white2016sampling}
Tom White.
\newblock Sampling generative networks.
\newblock \emph{arXiv preprint arXiv:1609.04468}, 2016.

\bibitem[Wulff \& Torralba(2020)Wulff and Torralba]{wulff2020improving}
Jonas Wulff and Antonio Torralba.
\newblock Improving inversion and generation diversity in stylegan using a
  gaussianized latent space.
\newblock \emph{arXiv preprint arXiv:2009.06529}, 2020.

\bibitem[Yeo \& Johnson(2000)Yeo and Johnson]{yeo2000new}
In-Kwon Yeo and Richard~A Johnson.
\newblock A new family of power transformations to improve normality or
  symmetry.
\newblock \emph{Biometrika}, 87\penalty0 (4):\penalty0 954--959, 2000.

\bibitem[Yilmaz(2001)]{yilmaz2001seismic}
{\"O}zdo{\u{g}}an Yilmaz.
\newblock \emph{Seismic data analysis}, volume~1.
\newblock Society of exploration geophysicists Tulsa, 2001.

\bibitem[Zbontar et~al.(2018)Zbontar, Knoll, Sriram, Murrell, Huang, Muckley,
  Defazio, Stern, Johnson, Bruno, et~al.]{zbontar2018fastmri}
Jure Zbontar, Florian Knoll, Anuroop Sriram, Tullie Murrell, Zhengnan Huang,
  Matthew~J Muckley, Aaron Defazio, Ruben Stern, Patricia Johnson, Mary Bruno,
  et~al.
\newblock fastmri: An open dataset and benchmarks for accelerated mri.
\newblock \emph{arXiv preprint arXiv:1811.08839}, 2018.

\bibitem[Zhang et~al.(2018)Zhang, Isola, Efros, Shechtman, and
  Wang]{zhang2018perceptual}
Richard Zhang, Phillip Isola, Alexei~A Efros, Eli Shechtman, and Oliver Wang.
\newblock The unreasonable effectiveness of deep features as a perceptual
  metric.
\newblock 2018.

\bibitem[Zhang \& Castagna(2011)Zhang and Castagna]{zhang2011seismic}
Rui Zhang and John Castagna.
\newblock Seismic sparse-layer reflectivity inversion using basis pursuit
  decomposition.
\newblock \emph{Geophysics}, 76\penalty0 (6):\penalty0 R147--R158, 2011.

\bibitem[Zhao(2005)]{zhao2005fast}
Hongkai Zhao.
\newblock A fast sweeping method for eikonal equations.
\newblock \emph{Mathematics of computation}, 74\penalty0 (250):\penalty0
  603--627, 2005.

\bibitem[Zhu et~al.(1997)Zhu, Byrd, Lu, and Nocedal]{zhu1997algorithm}
Ciyou Zhu, Richard~H Byrd, Peihuang Lu, and Jorge Nocedal.
\newblock Algorithm 778: L-bfgs-b: Fortran subroutines for large-scale
  bound-constrained optimization.
\newblock \emph{ACM Transactions on mathematical software (TOMS)}, 23\penalty0
  (4):\penalty0 550--560, 1997.

\end{thebibliography}
\bibliographystyle{iclr2023_conference}

\newpage
\appendix

\section{Background of forward models}
\label{append:background_forward_models}

\subsection{Compressive sensing MRI}
\label{append:MRI}
The MRI process essentially samples the spatial frequency components of a target, following some trajectories in the spatial frequency space (k-space) according to the design of the physical system. For example, a system may sample the k-space line-by-line horizontally/vertically or in radial directions. If the k-space has been fully sampled on a Cartesian grid, one can directly use inverse FFT to reconstruct the image. For various practical reasons, however, it is necessary to speed up the data collection process, usually by skipping data points in the k-space, which can be mathematically represented by a masking operation. In addition, there can be multiple coils with different sensitivity maps collecting data simultaneously. The mathematical formulation reads
\begin{equation}
  \rvd_i = \mP \mF \mS_i \rvm + \bm{\epsilon},\quad i = 1, \cdots, N_{\text{coils}}
  \label{eqn:mri_eq_multicoil}
\end{equation}
where \(\rvd_i \in \mathbb{C}^N\) is the k-space data corresponding to the \(i\)-th coil, \(\rvm \in \mathbb{R}^N\) is the target object, \(\mP \in \mathbb{R}^{N\times N}\) is the mask, \(\mF \in \mathbb{C}^{N\times N}\) is the Fourier transform operator, \(\mS_i \in \mathbb{C}^{N\times N}\) is the point-wise sensitivity map (a diagonal matrix) corresponding to the \(i\)-th coil, and \(\bm{\epsilon}\) denotes noise.

\begin{figure}[htpb]
	\centering
		\setlength{\tabcolsep}{0.3cm}
		\begin{tabular}{ll}
			\small{(a)} & \small{(b)} \\
			\includegraphics[width=0.24\textwidth]{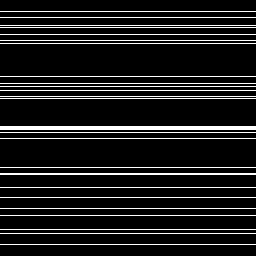} & \includegraphics[width=0.24\textwidth]{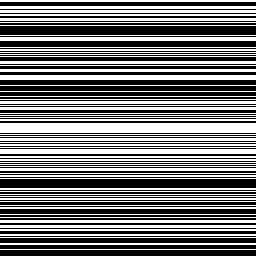}
		\end{tabular}
	\caption{Masks for compressive sensing MRI~\citep{kelkar2021prior}. White: 1, black: 0. (a) \(8\times\) acceleration; (b) \(2\times\) acceleration.}
	\label{fig:mri_masks}
\end{figure}

To ensure a fair comparison with prior work and reproducibility, we used the same masks from the repository of \citet{kelkar2021prior} (\figref{fig:mri_masks}). In addition, we also used the same single-coil setup as in \citet{kelkar2021prior}, where the sensitivity matrix is an identity matrix.

We condense the effects of all operators into a linear operator \(\mA \in \mathbb{C}^{M\times N}\) and arrive at the under-determined system
\begin{equation}
  \rvd = \mA \rvm + \bm{\epsilon} \tag{\ref{eqn:mri_eq}}
\end{equation}
from the main text, where we use \(\text{Accl}=N/M\) to denote the acceleration ratio.

\subsection{Deblurring}
\label{append:deblurring}
The mathematical model behind deblurring is
\begin{equation}
  \rvd = \mH \ast \rvm + \bm{\epsilon},
\end{equation}
where \(\mH\) is a smoothing filter, \(\ast\) denotes convolution, and \(\bm{\epsilon}\) is noise.

The purpose of deburring is to recover the original sharp image \(\rvm\) given a noisy blurred observation \(\rvd\). In this study, we showed deblurring examples for natural images. In scientific applications, deblurring or deconvolution is also a powerful tool. For example, in astronomy, \(\rvd\) is a blurred image from a telescope, \(\mH\) is a point-spread function (PSF) constructed from the physics model of the telescope, and we want to obtain a sharper image from the observation~\citep{starck2002deconvolution}. In geophysics, \(\rvd\) can be the seismic data, \(\mH\) is a calibrated wavelet, and we want to obtain sharp images of reflectivities defining the boundaries of subsurface strata~\citep{lines1984review,zhang2011seismic}. In general, \(\mH\) is a low-pass or band-pass filter, so certain frequency contents are lost in the forward process. The deblurring or deconvolution process needs to recover such missing information. In addition, the noise makes the inversion process unstable. The deblurring or deconvolution problem is thus ill-posed.

\subsection{Eikonal tomography}
\label{append:eikonal_tomography}
\begin{figure}[h]
	\centering
		\setlength{\tabcolsep}{0.1cm}
		\begin{tabular}{ll}
			\small{(a)} & \small{(b)} \\
			\includegraphics[width=0.42\textwidth]{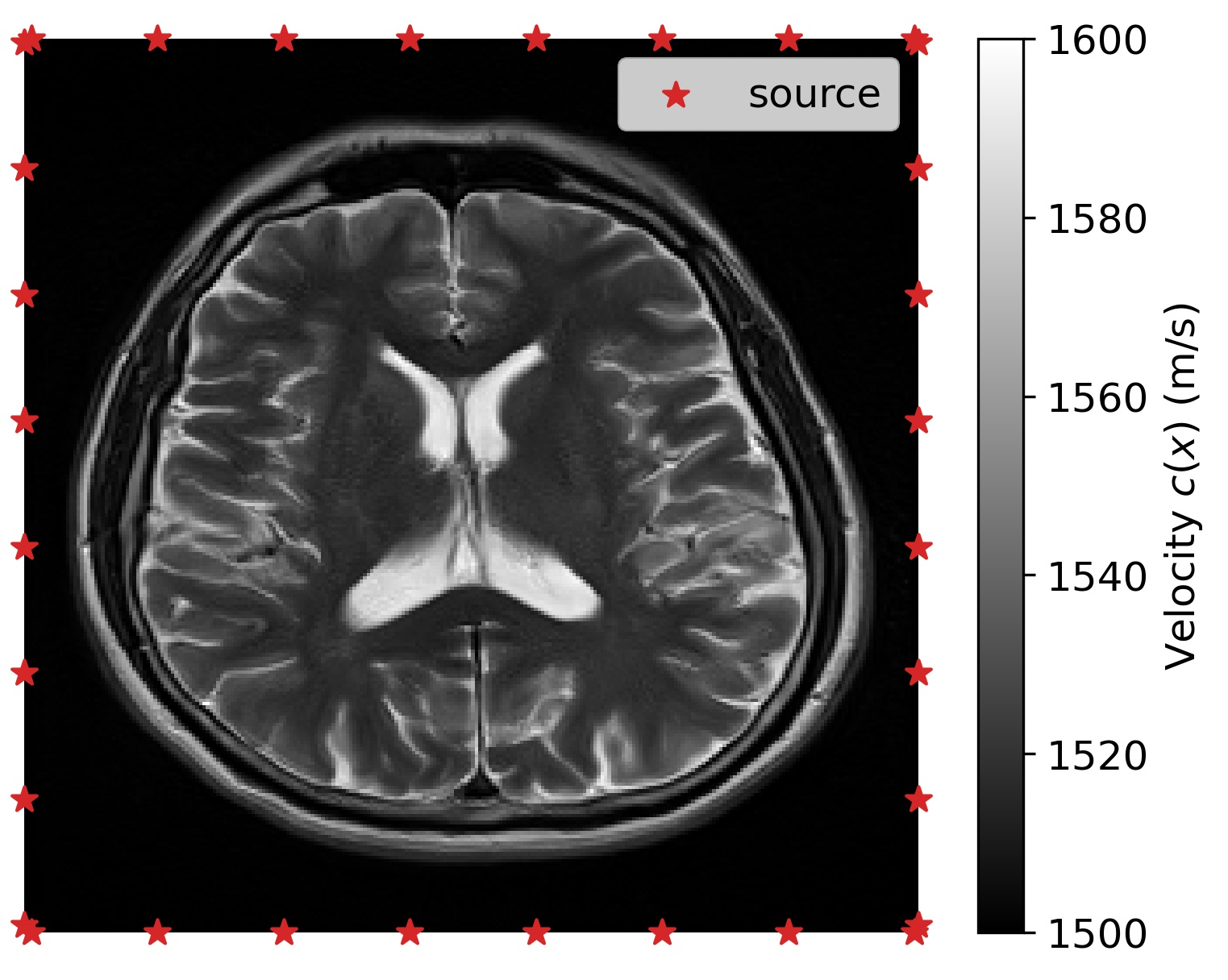} 
			& \includegraphics[width=0.42\textwidth]{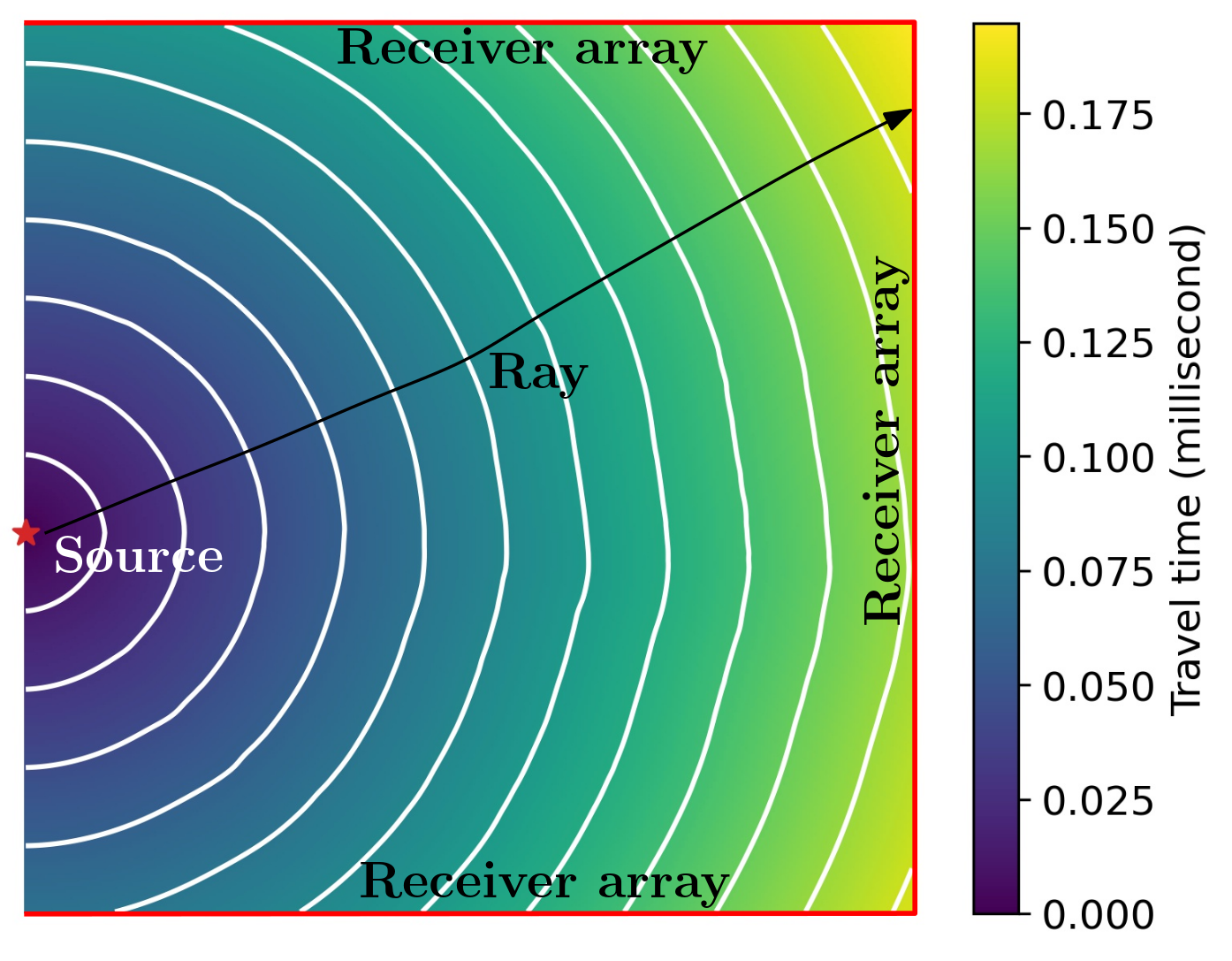} 
		\end{tabular}
		\begin{tabular}{l}
			\small{(c)}\\
			\includegraphics[width=0.42\textwidth]{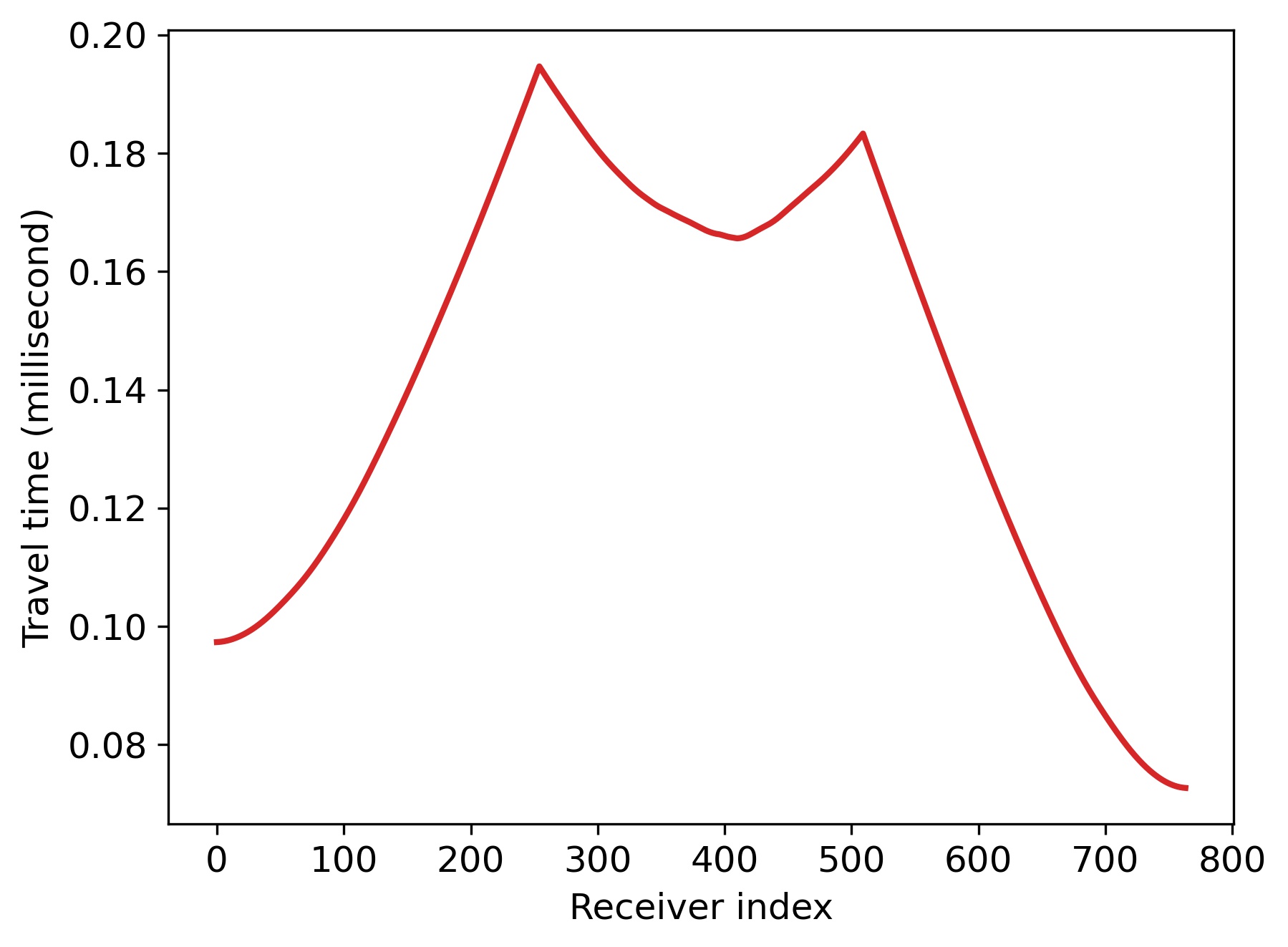} 
		\end{tabular}
	\caption{Experimental setup for eikonal tomography. (a) The target image and source locations; (b) Travel time field and the receiver array corresponding to the source. The contours are wavefronts. The propagation of waves can be viewed as curved rays traveling in directions perpendicular to such wavefronts; (c) The profile of the shortest wave travel time recorded at the receiver array. The receiver index starts from 0 in the top-left corner in subfigure (b) and increases in the clockwise direction.}
	\label{fig:tomo_config}
\end{figure}
In acoustic wave imaging (e.g., ultrasound tomography), we excite waves using sparsely distributed sources one at a time at the boundary of the object. Then we reconstruct its internal structures (the spatial distribution of wave speed) given the first-arrival travel time recorded on the boundary. The following eikonal equation describes the shortest travel time \(T(\vx;\vx_s)\) that the acoustic wave emerging from the source location \(\vx_s\) takes to reach location \(\vx\) inside the target object in the high-frequency limit~\citep{yilmaz2001seismic}:
\begin{equation}
	|\nabla T\left( \vx; \vx_s \right)| = 1/ c\left( \vx \right), \quad T(\vx_s;\vx_s) = 0,
\end{equation}
where \(c(\vx)\) is the wave propagation speed at each location. Both this eikonal PDE and the implicitly defined forward mapping \(c(\vx)\rightarrow T(\vx)\) are nonlinear.

To be more specific, we show the setup of our experiment in \figref{fig:tomo_config}(a). For the convenience of numerical testing, we put sources and receivers on the boundaries of a square box that contains the object, suggesting that the object is immersed in a square box filled with water. In reality, one can put sources and receivers directly on the target. The dimension of the square area is \(25.6 \text{ cm} \times 25.6 \text{ cm}\) with a grid interval of 0.001 m. There are eight sources located on each side, and receivers are located at each grid point on the boundary. \Figref{fig:tomo_config}(b) shows the shortest travel time field of the generated wave from the indicated source location. For each source, we only use receivers on the three other sides, indicated by the red lines, excluding the one on which the source is located. \Figref{fig:tomo_config}(c) shows the profile of the shortest wave travel time recorded at the receiver array. The receiver index starts from 0 in the top-left corner and increases in the clockwise direction.

We can interpret the nonlinearity and ill-posedness of eikonal tomography from another perspective. As shown in \Figref{fig:tomo_config}(b), we plot contours \(T(\vx) = \texttt{const}\) representing the wavefronts. Under the high-frequency approximation, the propagation of waves can be viewed as curved rays traveling in directions perpendicular to such wavefronts. Since the wavefronts depend on velocity field \(c(\vx)\), the rays are also functionals of the parameter \(c(\vx)\) to be estimated, contrary to straight-ray tomography such as CT. The eikonal inversion problem is thus nonlinear. Besides, the rays carry  information about the medium averaged along their paths. One property of curved-ray tomography is that the ray coverage is uneven inside the object. In fact, curved rays tend to avoid low-velocity areas, giving us little information about such regions, making the inverse problem intrinsically ill-posed.

We solved the eikonal equation using the fast sweeping method~\citep{zhao2005fast} and computed the gradient using the discrete adjoint-state method, using the code from the repository of \citet{li2020coupled}.

Our eikonoal tomography uses the same StyleGAN2 network as the compressive sensing MRI experiments. The value range of StyleGAN2 is \([-1, 1]\). In the forward model, we map the StyleGAN2 output to \(c(\vx)\) in two steps. First, we convert its values to the range of \([0, 1]\) by using \(\rvm \gets (\rvm + 1)/2\). Second, we convert the values to the range of acoustic wave velocity using \(\rvm \gets 100 \times \rvm + 1500\). This relationship is purely manufactured for our synthetic tests. One should use a more realistic relationship in practice.

\section{Additional motivating examples}
\label{append:add_latent_deviations}
\begin{figure}[htpb]
	\centering
	\includegraphics[width=0.8\textwidth]{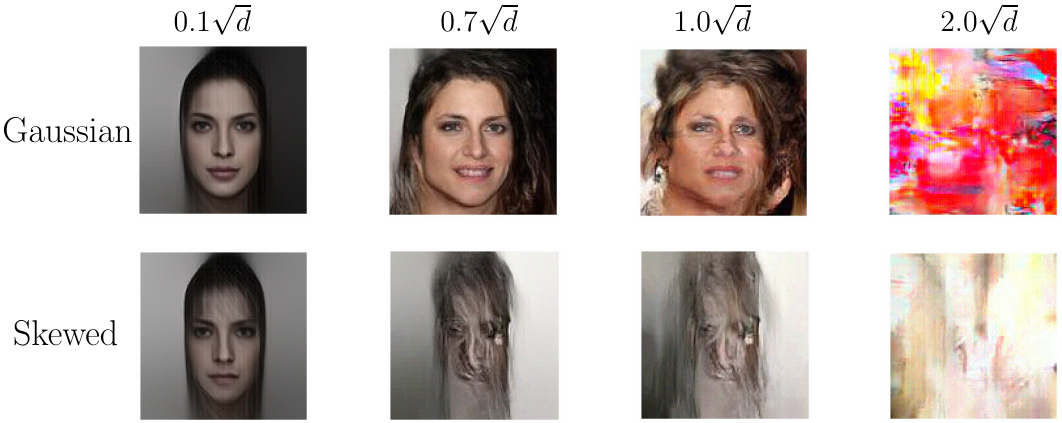}
	\caption{The visual effects of the \(\ell_2\) norm of latent tensors on Glow outputs. The top and bottom panels correspond to latent tensors drawn from a spherical Gaussian and a spherical skewed distribution, respectively. The two tensors are scaled to have various norms: 0.1, 0.7, 1.0, and 2.0 of \(\sqrt{d}\) (the square root of the latent tensor dimension). The images are getting increasingly unrealistic as we increase the norm from 1.0$\sqrt{d}$. This phenomenon demonstrates why we need to use a temperature \(<1\) for Glow.}
	\label{fig:various_radii}
\end{figure}

\paragraph{Glow}
We varied the $\ell_2$-norm of the latent tensors for Glow and reported the outputs in \figref{fig:various_radii}. The images are getting increasingly unrealistic as we increase the norm from 1.0$\sqrt{d}$, where \(d\) is the latent tensor dimension. This phenomenon demonstrates why we need to use a temperature \(<1\) for Glow.

\begin{figure}[htpb]
	\centering
	\includegraphics[width=0.6\textwidth]{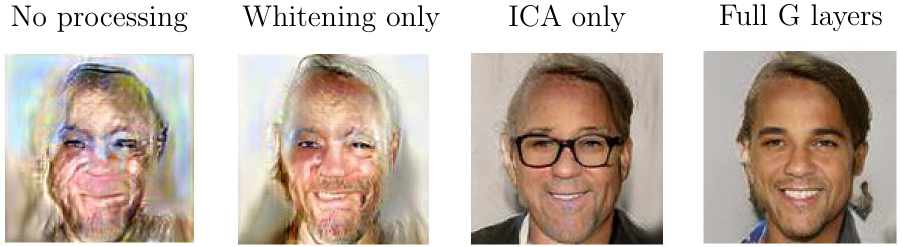}
	\caption{The visual effects of various components of Gaussianization layers on a pathologic style vector for StyleGAN2.}
	\label{fig:sg2_style_effects}
\end{figure}

\paragraph{StyleGAN2}
StyleGAN2 is a well-designed generator with a built-in spherical transformation for the style vector. It seems very robust to vectors drawn from the distributions similar to those in \figref{fig:effects_deviate_glow}. However, we were able to find challenging examples for it from inversion tasks. In \figref{fig:sg2_style_effects}, The first example is the direct output only using the pathologic style vector. The second example is the generator output using the style vector after whitening only. As we can see, whitening alone is ineffective in improving the image quality, which is confirmed by our ablation studies. On the contrary, if we use the ICA layer, we see a huge improvement in visual quality, and our interpretation is that getting rid of higher-order dependencies is very important. Note that there are eyeglasses in the third figure, meaning that if we relax the 1D Gaussian requirement, we can sample some rare examples. Finally, we use the full G layers to get another plausible image.

\paragraph{Stable diffusion}
\begingroup
\setlength{\tabcolsep}{1.2pt}
\begin{figure}[htpb]
  \centering
  \newcommand{\sizeI}{0.15}
  \begin{tabular}{c @{~~} c @{~~} c}

	\begin{tabular}{ccc}
	& Before & After\\
	\rotatebox[origin=c]{90}{Skewed} & \raisebox{-.5\height}{\includegraphics[width=\sizeI\linewidth]{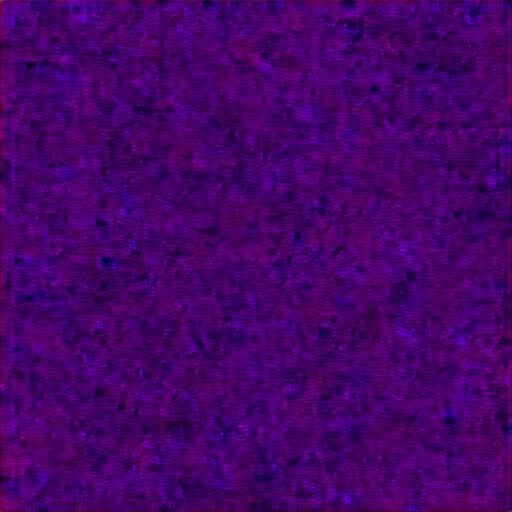}} & \raisebox{-.5\height}{\includegraphics[width=\sizeI\linewidth]{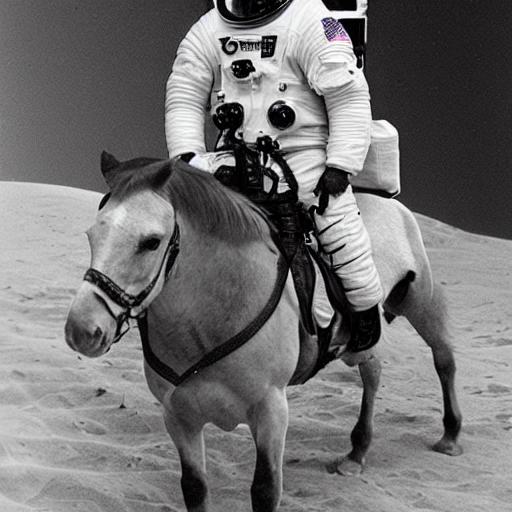}}\\

	\rotatebox[origin=c]{90}{Heavy-tailed} & \raisebox{-.5\height}{\includegraphics[width=\sizeI\linewidth]{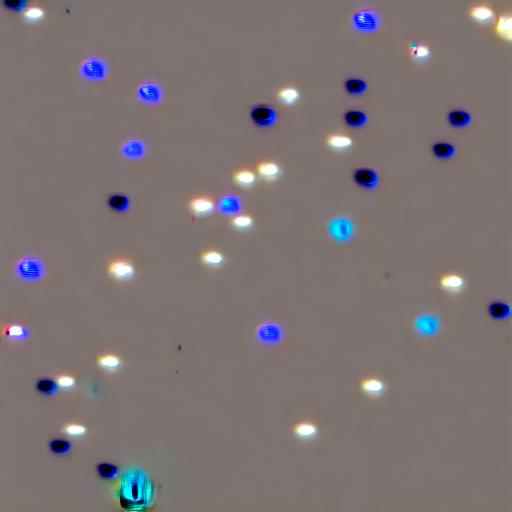}} & \raisebox{-.5\height}{\includegraphics[width=\sizeI\linewidth]{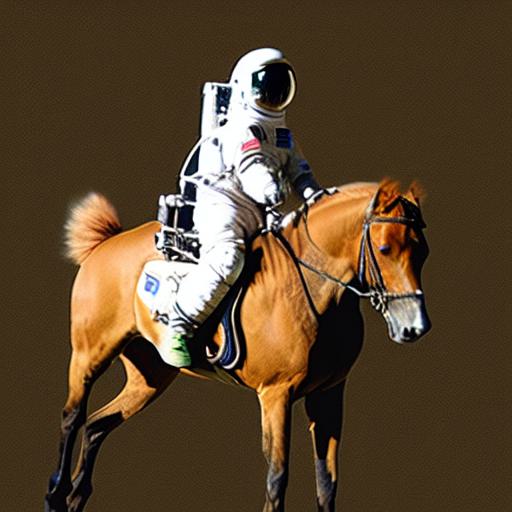}}\\

	\rotatebox[origin=c]{90}{Non-\iid} & \raisebox{-.5\height}{\includegraphics[width=\sizeI\linewidth]{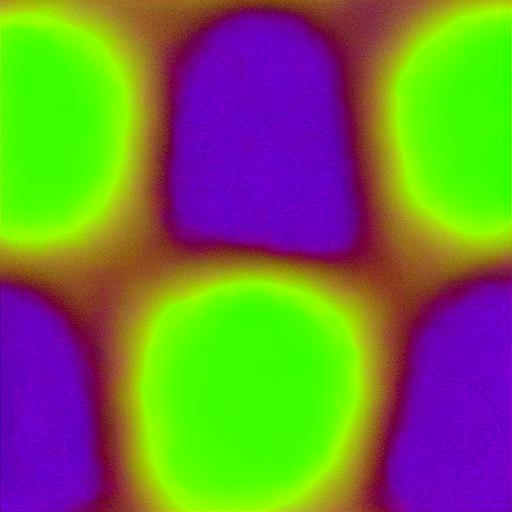}} & \raisebox{-.5\height}{\includegraphics[width=\sizeI\linewidth]{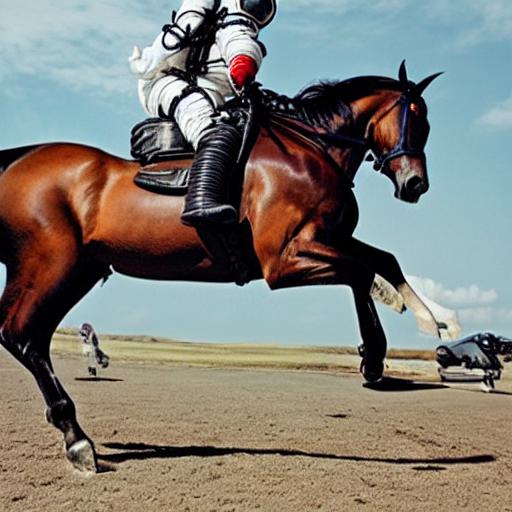}}\\
	
	\end{tabular} 

 	&
	\begin{tabular}{cc}
	Before & After\\
	\raisebox{-.5\height}{\includegraphics[width=\sizeI\linewidth]{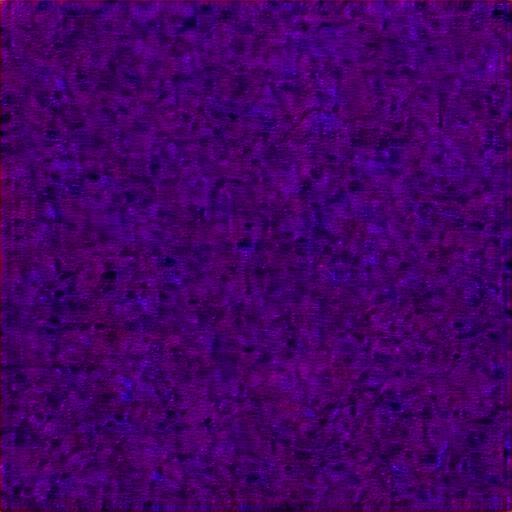}} & \raisebox{-.5\height}{\includegraphics[width=\sizeI\linewidth]{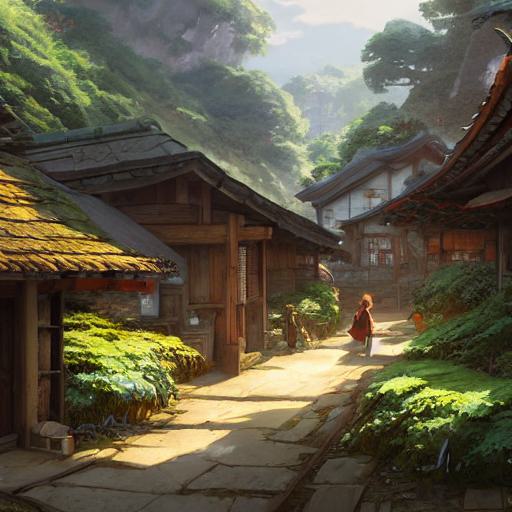}} \\
	
	\raisebox{-.5\height}{\includegraphics[width=\sizeI\linewidth]{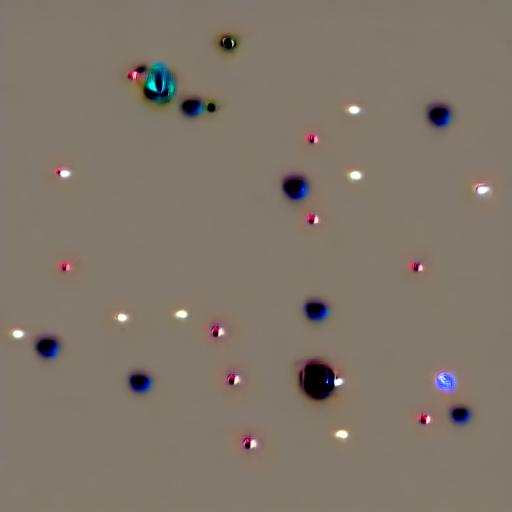}} & \raisebox{-.5\height}{\includegraphics[width=\sizeI\linewidth]{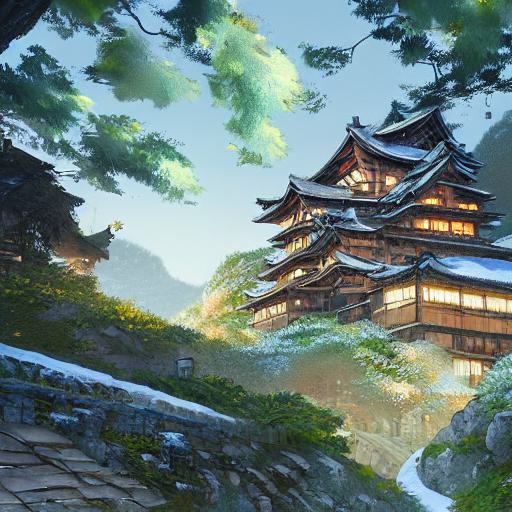}} \\
	
	\raisebox{-.5\height}{\includegraphics[width=\sizeI\linewidth]{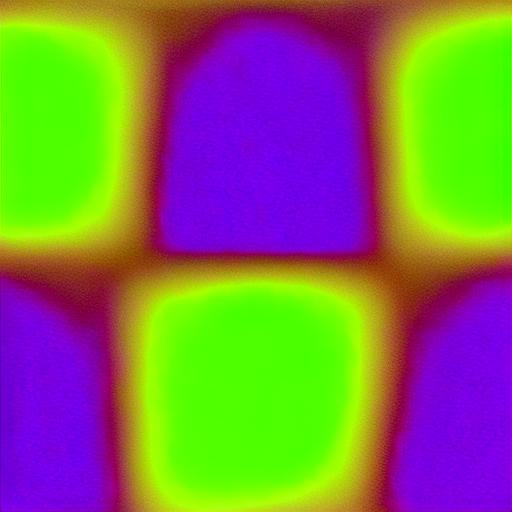}} & \raisebox{-.5\height}{\includegraphics[width=\sizeI\linewidth]{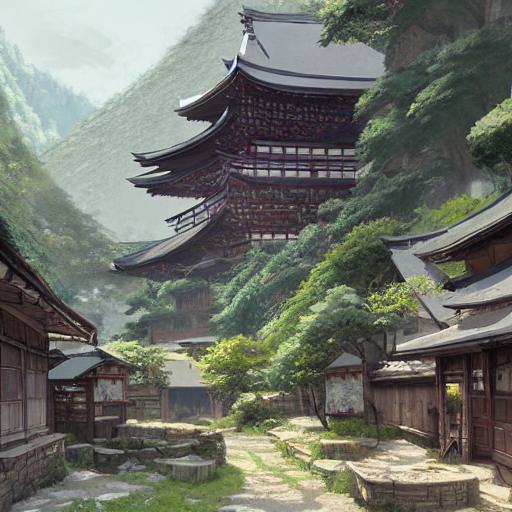}} \\
	\end{tabular}

	&
	\begin{tabular}{cc}
	Before & After\\
	\raisebox{-.5\height}{\includegraphics[width=\sizeI\linewidth]{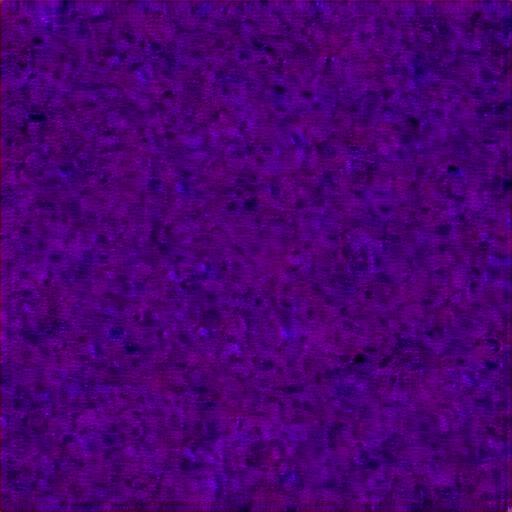}} & \raisebox{-.5\height}{\includegraphics[width=\sizeI\linewidth]{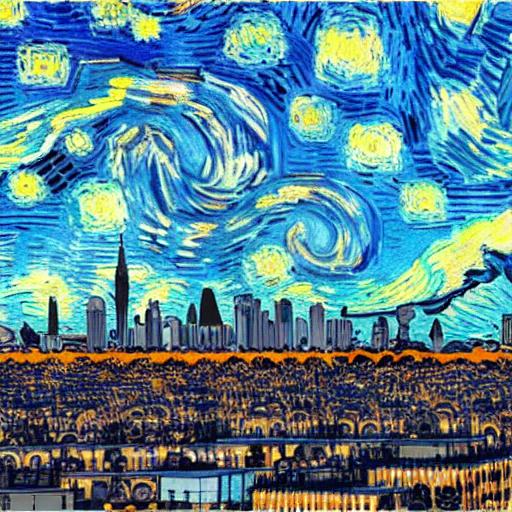}} \\
	
	\raisebox{-.5\height}{\includegraphics[width=\sizeI\linewidth]{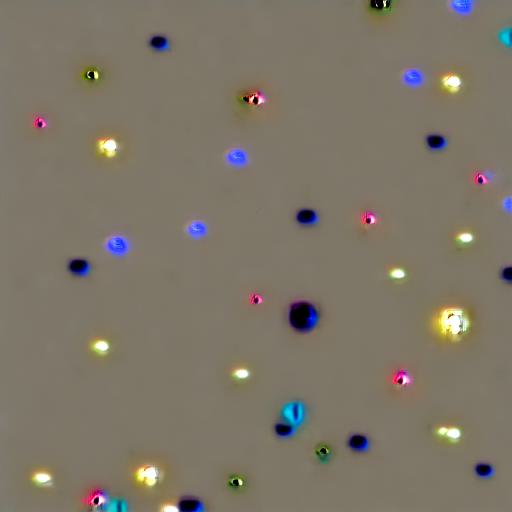}} & \raisebox{-.5\height}{\includegraphics[width=\sizeI\linewidth]{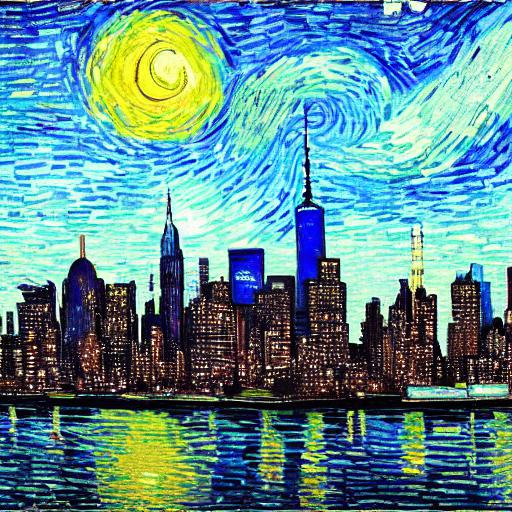}} \\
	
	\raisebox{-.5\height}{\includegraphics[width=\sizeI\linewidth]{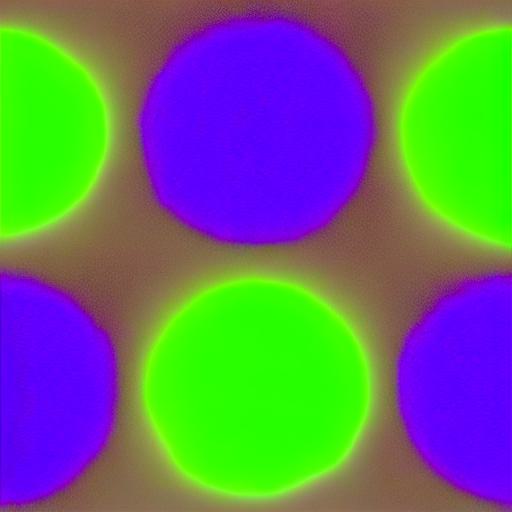}} & \raisebox{-.5\height}{\includegraphics[width=\sizeI\linewidth]{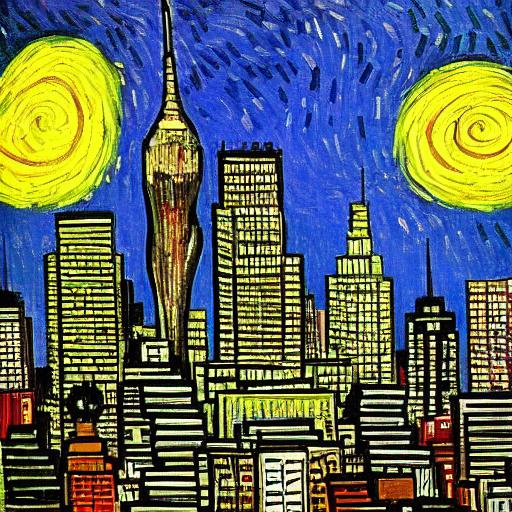}} \\
	\end{tabular}

  \end{tabular}

	\caption{Comparisons of Stable Diffusion outputs before and after applying the Gaussianization layers to latent tensors. The components of these latent tensors are \iid skewed, \iid heavy-tailed, and non-i.i.d., respectively. We scaled all latent tensors to be on the sphere with a radius of \(\sqrt{\text{tensor dimension}}\). Stable Diffusion picks up and amplifies the sparse large-amplitude points in the heavy-tailed case and the 2D sinusoidal pattern in the non-\iid case.}
	\label{fig:stable_difussion_g_layers}
\end{figure}
\endgroup

\begingroup
\setlength{\tabcolsep}{1.2pt}
\begin{figure}[htpb]
  \centering
  \newcommand{\sizeI}{0.15}
  \begin{tabular}{c @{~~} c}
  	\begin{tabular}{c c c}

	$0.7\sqrt{d}$ & $\sqrt{d}$ & $1.3\sqrt{d}$\\
	\includegraphics[width=\sizeI\linewidth]{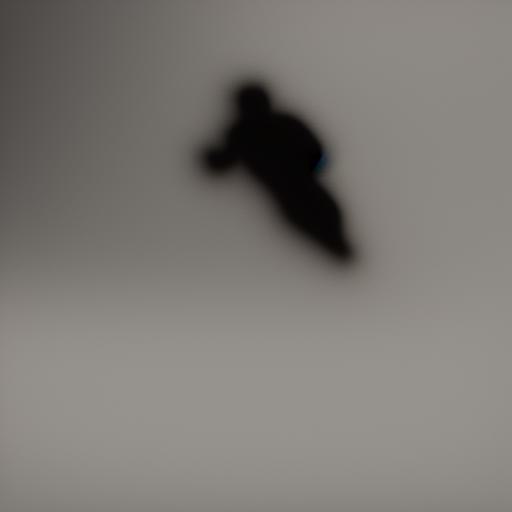} & \includegraphics[width=\sizeI\linewidth]{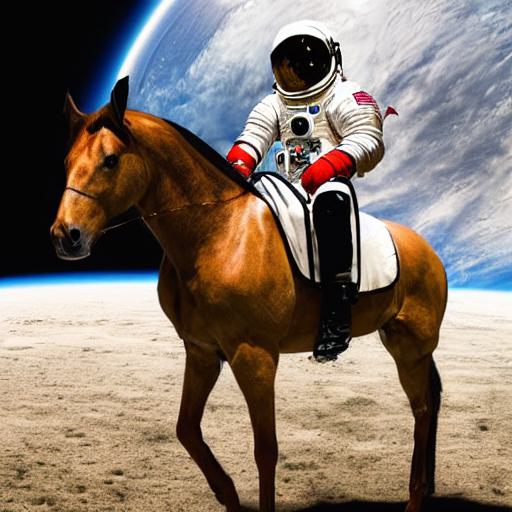} & \includegraphics[width=\sizeI\linewidth]{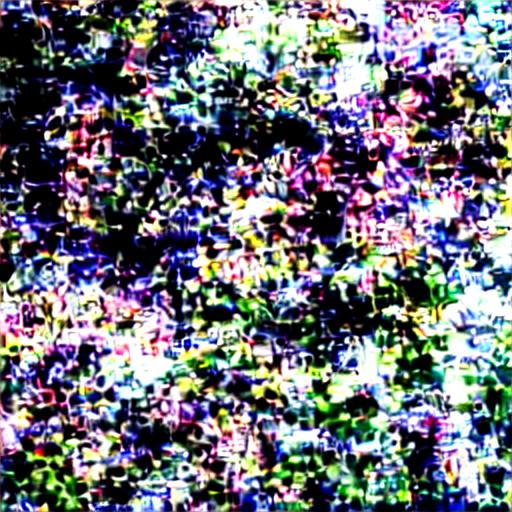}
	\end{tabular}
	&
	\begin{tabular}{c c c}
	$0.7\sqrt{d}$ & $\sqrt{d}$ & $1.3\sqrt{d}$\\
	\includegraphics[width=\sizeI\linewidth]{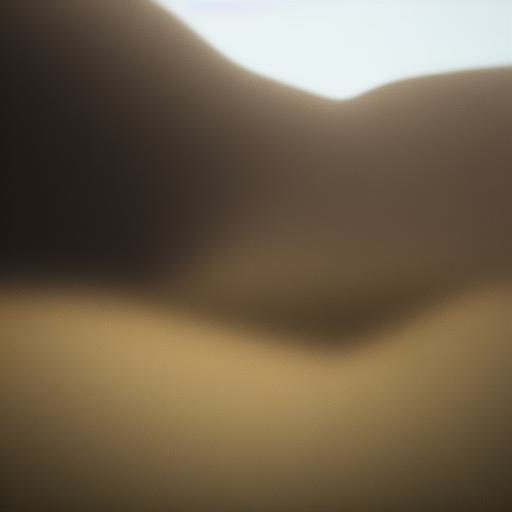} & \includegraphics[width=\sizeI\linewidth]{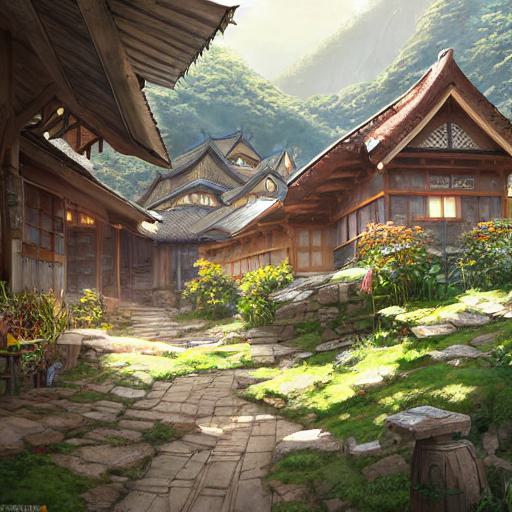} & \includegraphics[width=\sizeI\linewidth]{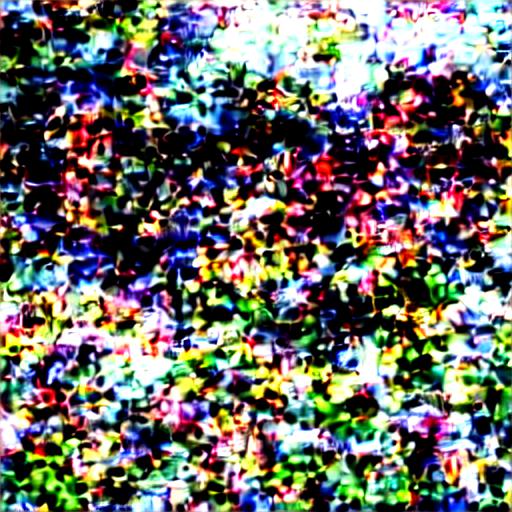}
	\end{tabular}

  \end{tabular}

	\caption{Comparison of Stable Diffusion outputs generated using spherical Gaussian latent tensors of different norms, where \(d = \sqrt{\text{tensor dimension}}\).}
	\label{fig:stable_difussion_norm}
\end{figure}
\endgroup
Stable Diffusion~\citep{Rombach_2022_CVPR} is a state-of-the-art deep generative model with text-to-image synthesis capability, which maps a Gaussian latent tensor to a high-resolution image. 

We repeated the tests for Glow and StyleGAN2 on Stable Diffusion and reported the results in \figref{fig:stable_difussion_g_layers}. The distribution for skewed case was the exponential-gamma distribution \(\log \Gamma(1, 1)\) (or the ``log-gamma distribution" in scipy), the distribution for the heavy-tailed case was a Lambert \(W\times F_X\) distribution with parameter \(\delta=0.5\) based on a standard Gaussian ~\citep{goerg2015lambert}. To generate the latent tensor with non-\iid entries, we first sampled \(\rvz_0 \in \mathbb{R}^{4\times 64 \times 64} \sim \mathcal{N}(\boldsymbol{0}, \rmI)\). Then we added \(0.5 * \sin(2 \pi/64\, x) \otimes \cos(2 \pi/64\, y)\) to each channel of \(\rvz_0\) for the latent tensor \(\rvz\), where \(\otimes\) stands for outer product. The number of denoising steps, classifier-free guidance, and the output dimensions for Stable Diffusion are 50, 7.5, and \(512\times 512\), respectively.

The outputs of Stable Diffusion are completely out-of-range if the latent tensors deviate from standard Gaussian, with much poorer quality than those from Glow and StyleGAN2. For example, Stable Diffusion picks up and amplifies the sparse large-amplitude points in the heavy-tailed case and the 2D sinusoidal pattern in the non-\iid case. However, with our Gaussianization layers, we are still able to constrain the outputs within the plausible range. Interestingly, we show in \figref{fig:stable_difussion_norm} that Stable Diffusion is much more sensitive to the norm of latent tensors than the other DGMs. Therefore, we anticipate that our Gaussianization layers are critical for potential inversion applications using Stable Diffusion such as text-guided inversion.

\section{Additional experiments}
\label{append:additional_exp}
First, we report in \tabref{tab:ablation_style_noise} an ablation study on the effects of various combinations of regularization techniques on style vectors and noise maps in StyleGAN2. The observations are (1) If we turn off the update of noise maps, the Gaussianization layers lead to the best result; (2) If we turn on the update of noise maps, applying the Gaussianization layers to both the style vectors and noise maps gives the best results.
\begin{table}[htpb]
\setlength{\extrarowheight}{2pt}
\caption{Ablation study on different combinations of regularization on the style vectors and noise maps in StyleGAN2 for compressive sensing MRI. In the parentheses, ``u'' stands for unconstrained, ``s'' stands for spherical constraint, ``g'' stands for reparameterization with the Gaussianization layers, ``o'' stands for orthogonal reparameterization, and ``\xmark'' means that the update is turned off. Note that StyleGAN2 has a built-in spherical transformation layer for the style vectors.}
\label{tab:ablation_style_noise}
\centering
\small
\begin{tabular}{l@{~~~~~}c@{~~~~~}c@{~~~~~}c@{~~~~~}c@{~~~~~}c@{~~~~~}c@{~~~~~}}
 \toprule
  \multirow{2}{3em}{Method} 
            & \multicolumn{2}{l}{Accl = 8x, SNR = 20 dB} & \multicolumn{2}{l}{Accl = 8x, SNR = 10 dB} \\
            \cmidrule(r){2-3} \cmidrule(r){4-5} 
            & PSNR$\uparrow$ & SSIM$\uparrow$ & PSNR$\uparrow$ & SSIM$\uparrow$ \\
 \hline
 Style (s), Noise (\xmark)   & 26.58\tiny{$\pm$5.90} & 0.80\tiny{$\pm$0.137}
            & 25.57\tiny{$\pm$5.24} & 0.78\tiny{$\pm$0.137} \\

 Style (g), Noise (\xmark)   & 27.39\tiny{$\pm$5.26} & 0.83\tiny{$\pm$0.124} 
            & 27.02\tiny{$\pm$4.57} & 0.82\tiny{$\pm$0.122} \\

 Style (o), Noise (\xmark)   & 25.88\tiny{$\pm$5.74} & 0.78\tiny{$\pm$0.134}
            & 25.39\tiny{$\pm$5.09} & 0.77\tiny{$\pm$0.131} \\

 Style (s), Noise (u)   & 26.51\tiny{$\pm$6.51} & 0.77\tiny{$\pm$0.163}
            & 22.49\tiny{$\pm$5.62} & 0.57\tiny{$\pm$0.266} \\

 Style (g), Noise (o)   & 27.43\tiny{$\pm$6.04} & 0.81\tiny{$\pm$0.134}
            & 24.93\tiny{$\pm$4.27} & 0.76\tiny{$\pm$0.128} \\

 Style (o), Noise (o)   & 26.14\tiny{$\pm$5.79} & 0.78\tiny{$\pm$0.134}
            & 25.10\tiny{$\pm$5.08} & 0.77\tiny{$\pm$0.132} \\

 Style (o), Noise (g)   & 26.04\tiny{$\pm$5.70} & 0.78\tiny{$\pm$0.134}
            & 25.22\tiny{$\pm$5.12} & 0.77\tiny{$\pm$0.133} \\

 Style (g), Noise (g)   & 27.99\tiny{$\pm$5.70} & 0.83\tiny{$\pm$0.128}
            & 25.48\tiny{$\pm$4.76} & 0.78\tiny{$\pm$0.149} \\

\bottomrule
\end{tabular}
\end{table}

Second, we studied the effect of patch size on the performance of Gaussianization layers. To be more specific, we tested the effects of various patch sizes on 1D style vectors. The largest patch size is 64 since the number of extracted patches should not be smaller than the dimension of the patches. We only turned on the ICA layer and the standardization layer in this experiment. One can observe that both the PSNR and the SSIM increase as the patch size increases. We would advise using the largest possible patch size in Gaussianization layers, although we only used a patch size of 32 for style vectors in all other experiments.

In all experiments, we fixed the patch size for Glow as \(3\times 8\times 8\) and the patch size for the noise maps in StyleGAN2 as \(1\times 8\times 8\). In the experiments with Glow, the image dimension was \(3\times 128\times 128\). If we chose a larger patch size, we could not obtain enough patch vectors. As for StyleGAN2, if we look at the parameterization illustrated in \figref{fig:style_gan_param}, there is already a \(4\times 4\) noise map unconstrained if we choose the patch size as \(8\times 8\). If we increase the patch size to \(1\times 16 \times 16\), there will be two additional \(8\times 8 \) noise maps unconstrained. To minimize the influence of unconstrained parameters, we chose the patch size for noise maps as \(1\times 8\times 8\).
\begin{table}[htpb]
\setlength{\extrarowheight}{1pt}
\caption{Ablation study on patch size of the style vectors in StyleGAN2.}
\label{tab:ablation_style_psize}
\vspace{0.2em}
\centering
\small
\begin{tabular}{l@{~~~~~}c@{~~~~~}c@{~~~~~}}
 \toprule
  Method      & PSNR$\uparrow$ & SSIM$\uparrow$ \\
 \hline
 
 Patch size = 8  	& 25.33\tiny{$\pm$6.42} & 0.76\tiny{$\pm$0.155} \\
 Patch size = 16  	& 26.83\tiny{$\pm$6.57} & 0.79\tiny{$\pm$0.144} \\
 Patch size = 32  	& 27.91\tiny{$\pm$5.77} & 0.82\tiny{$\pm$0.129} \\
 Patch size = 64  	& 28.03\tiny{$\pm$5.44} & 0.83\tiny{$\pm$0.130} \\

\bottomrule
\end{tabular}
\end{table}

Third, we did a parameter sweep on the weighting parameter \(\beta\) in the Glow-regularized deblurring problem (formulation \ref{eqn-augmented_inversion} or \ref{eqn:Glow_regularization_inv}). Consistent with \figref{fig:figs_intro}, the conventional Glow-based regularization is not as effective as our Gaussianization layers. If \(\beta\) is too large (\eg, 10 or 100), the inversion results become unrealistic with huge errors.
\begin{table}[htpb]
\setlength{\extrarowheight}{2pt}
\caption{Glow-regularized deblurring results using different $\beta$s.}
\label{tab:glow_deblur_betas}
\vspace{0.2em}
\centering
\small
\begin{tabular}{l@{~~~}c@{~~~}c@{~~~}c@{~~~}}
 \toprule
  Method      & LPIPS$\downarrow$ & PSNR$\uparrow$ & SSIM$\uparrow$ \\
 \hline
 
 $\beta=0.0$  & 0.172\tiny{$\pm$0.06} & 21.74\tiny{$\pm$1.27} & 0.58\tiny{$\pm$0.072} \\
 $\beta=0.01$  & 0.172\tiny{$\pm$0.06} & 21.75\tiny{$\pm$1.15} & 0.58\tiny{$\pm$0.067} \\
 $\beta=0.1$  & 0.174\tiny{$\pm$0.06} & 21.72\tiny{$\pm$1.15} & 0.57\tiny{$\pm$0.069} \\
 $\beta=1.0$  & 0.174\tiny{$\pm$0.06} & 21.95\tiny{$\pm$1.21} & 0.59\tiny{$\pm$0.067} \\
 $\beta=10.0$  & 0.247\tiny{$\pm$0.08} & 22.55\tiny{$\pm$1.10} & 0.58\tiny{$\pm$0.086} \\
 $\beta=100.0$  & 0.602\tiny{$\pm$0.15} & 12.56\tiny{$\pm$2.63} & 0.12\tiny{$\pm$0.101} \\

\bottomrule
\end{tabular}
\end{table}

Fourth, we investigated the performance of Gaussianization layers when the forward model is inaccurate (\tabref{tab:wrong_filter}). Our Gaussianization layers still outperform the conventional Glow-regularized inversion.
\begin{table}[htpb]
\setlength{\extrarowheight}{2pt}
\caption{Glow-regularized deblurring with an inaccurate filter. The ground-truth standard deviation of the Gaussian filter is 3, which is used to generate the observed data, but we used 5 for inversion.}
\label{tab:wrong_filter}
\vspace{0.2em}
\centering
\small
\begin{tabular}{l@{~~~}c@{~~~}c@{~~~}c@{~~~}}
 \toprule
  Method      & LPIPS$\downarrow$ & PSNR$\uparrow$ & SSIM$\uparrow$ \\
 \hline
 
 Conventional \tiny{($\beta=1.0$)} & 0.21\tiny{$\pm$0.06} & 17.72\tiny{$\pm$1.41} & 0.49\tiny{$\pm$0.059} \\
 G layers  					& \best{0.17\tiny{$\pm$0.06}} & \best{18.70\tiny{$\pm$1.40}} & \best{0.50\tiny{$\pm$0.060}} \\

\bottomrule
\end{tabular}
\end{table}

\begin{table}[htpb]
\setlength{\extrarowheight}{2pt}
\caption{Ablation study of the components of Gaussianization layers applied to Glow-regularized deblurring.}
\label{tab:ablation_glow_g_layers}
\vspace{0.2em}
\centering
\small
\begin{tabular}{l@{~~}l@{~~~}l@{~~~~}c@{~~~}c@{~~~}c@{~~~}}
 \toprule
  \multicolumn{3}{c}{Method}     & LPIPS$\downarrow$ & PSNR$\uparrow$ & SSIM$\uparrow$ \\
 \hline
 
 ICA (\xmark),\,& YJ (\xmark),& Lambt (\xmark) 
 	& 0.17\tiny{$\pm$0.06} & 21.79\tiny{$\pm$1.11} & 0.583\tiny{$\pm$0.065} \\

 ICA (\xmark),\,& YJ (\xmark),& Lambt (\checkmark) 
 	& 0.16\tiny{$\pm$0.06} & 21.96\tiny{$\pm$1.31} & 0.588\tiny{$\pm$0.065} \\

 ICA (\xmark),\,& YJ (\checkmark),& Lambt (\xmark) 
 	& 0.17\tiny{$\pm$0.06} & 21.75\tiny{$\pm$1.25} & 0.579\tiny{$\pm$0.067} \\

 ICA (\xmark),\,& YJ (\checkmark),& Lambt (\checkmark) 
 	& 0.16\tiny{$\pm$0.06} & 21.99\tiny{$\pm$1.20} & 0.587\tiny{$\pm$0.066} \\

 ZCA,\,& YJ (\xmark),& Lambt (\xmark) 
 	& 0.16\tiny{$\pm$0.05} & 21.66\tiny{$\pm$1.28} & 0.579\tiny{$\pm$0.070} \\
 
 ICA (\checkmark),\,& YJ (\xmark),& Lambt (\xmark) 
 	& 0.14\tiny{$\pm$0.06} & \best{22.52\tiny{$\pm$1.30}} & 0.586\tiny{$\pm$0.070} \\

 ICA (\checkmark),\,& YJ (\xmark),& Lambt (\checkmark) 
 	& \best{0.13\tiny{$\pm$0.05}} & 22.47\tiny{$\pm$1.27} & \best{0.590\tiny{$\pm$0.064}} \\
 
 ICA (\checkmark),\,& YJ (\checkmark),& Lambt (\xmark) 
 	& 0.14\tiny{$\pm$0.05} & 22.49\tiny{$\pm$1.35} & 0.586\tiny{$\pm$0.070} \\
 
 ICA (\checkmark),\,& YJ (\checkmark),& Lambt (\checkmark) 
 	& \best{0.13\tiny{$\pm$0.06}} & 22.40\tiny{$\pm$1.27} & 0.586\tiny{$\pm$0.066} \\
\bottomrule
\end{tabular}
\end{table}

In addition, we did an ablation study on the components of Gaussianization layers in Glow-regularized deblurring (\tabref{tab:ablation_glow_g_layers}). We adopted the second parameterization scheme (\appref{append:reparam_schemes}). The observations are similar to those from \tabref{tab:ablation_mri}: 1. The ICA layer is the most significant part in improving result scores, especially LPIPS -- the score that matches human perception the best; 2. The whitening/ZCA layer is not effective when used alone; 3. There is no clear winner among combinations of 1D Gaussianization layers, and their difference in performance is marginal. So we picked the Lambert \(W \times F_X\) layer, one of the best-performing options, in all other deblurring experiments as the 1D Gaussianization layer.

\paragraph{Inversion using out-of-distribution images} Finally, we tested the limit of DGM-regularized inversion by using real-world out-of-distribution target images. We randomly sampled 25 MRI images from the fastMRI DICOM dataset~\citep{zbontar2018fastmri,knoll2020fastmri}. The RSNA clinical trial processor was used to anonymize the whole dataset, and each image has been manually inspected to make sure that there is no protected information. According to the fastMRI paper~\citep{zbontar2018fastmri}, "This dataset represents a larger variety of machines and settings than are present in the raw data." Also, the image values are represented in \texttt{uint16} rather than float numbers, and their value range can be very large (0 to several thousand). So we normalized the image values to be between range $[-1, 1]$ using \texttt{img = (img - img.min())/(img.max()-img.min())*2 - 1}. However, the trained StyleGAN2 outputs are always slightly beyond $[-1, 1]$, and we clipped the values in both test image generation and inversion. Therefore, the new examples are out-of-distribution.

We kept the same experimental setup of $8\times$ acceleration and 20 dB SNR and tested inversion using the spherical constraint and Gaussianization layers on both only using the style vectors and using the additional noise maps. We report the inversion results in \tabref{tab:fastmri_dicom} and \figref{fig:fastmri_dicom}. To save experiment time, we only use one 1D Gaussianization layer (either Yeo-Johnson or Lambert \(W\times F_X\)).
\begin{table}[htpb]
\setlength{\extrarowheight}{2pt}
\caption{MRI inversion using real-world out-of-distribution images.}
\label{tab:fastmri_dicom}
\vspace{0.2em}
\centering
\small
\begin{tabular}{l@{~~~}c@{~~~}c@{~~~}}
 \toprule
  Method      & PSNR$\uparrow$ & SSIM$\uparrow$ \\
 \hline
 
 (a) Style only, spherical  				& 26.14\tiny{$\pm$3.59} & 0.784\tiny{$\pm$0.062} \\
 (b) Style+noise, spherical  				& 25.26\tiny{$\pm$3.37} & 0.755\tiny{$\pm$0.065} \\
 (c) Style only, G layers, YJ for 1D  		& 25.23\tiny{$\pm$3.30} & 0.765\tiny{$\pm$0.050} \\
 (d) Style + noise, G layers, YJ for 1D  	& 25.39\tiny{$\pm$3.28} & 0.769\tiny{$\pm$0.045} \\
 (e) Style only, G layers, Lambert for 1D  	& 25.39\tiny{$\pm$3.38} & 0.762\tiny{$\pm$0.064} \\
 (f) Style + noise, G layers, Lambert for 1D  & 25.45\tiny{$\pm$3.39} & 0.765\tiny{$\pm$0.064} \\

\bottomrule
\end{tabular}
\end{table}

\begingroup
\setlength{\tabcolsep}{1.2pt}
\begin{figure}[htpb]
    \centering%
    \newcommand{\sizeI}{0.11}
    \newcommand{\imgTrue}[1]{\includegraphics[width=\sizeI\linewidth]{figures/figures_inv_mri_dicom/style_sphere_accl8x_snr20.0/#1/True.jpg}}
    \newcommand{\imga}[1]{\includegraphics[width=\sizeI\linewidth]{figures/figures_inv_mri_dicom/style_sphere_accl8x_snr20.0/#1/Inv.jpg}}
    \newcommand{\imgb}[1]{\includegraphics[width=\sizeI\linewidth]{figures/figures_inv_mri_dicom/style_noise_sphere_accl8x_snr20.0/#1/Inv.jpg}}
    \newcommand{\imgc}[1]{\includegraphics[width=\sizeI\linewidth]{figures/figures_inv_mri_dicom/gaus_style_yj_accl8x_snr20.0/#1/Inv.jpg}}
    \newcommand{\imgd}[1]{\includegraphics[width=\sizeI\linewidth]{figures/figures_inv_mri_dicom/gaus_style_noise_yj_accl8x_snr20.0/#1/Inv.jpg}}
    \newcommand{\imge}[1]{\includegraphics[width=\sizeI\linewidth]{figures/figures_inv_mri_dicom/gaus_style_lmbt_accl8x_snr20.0/#1/Inv.jpg}}
    \newcommand{\imgf}[1]{\includegraphics[width=\sizeI\linewidth]{figures/figures_inv_mri_dicom/gaus_style_noise_lmbt_accl8x_snr20.0/#1/Inv.jpg}}
	\begin{tabular}{ccccccc}
	\small{True} & \small{(a)} & \small{(b)} & \small{(c)} & \small{(d)} & \small{(e)} & \small{(f)}\\
	\imgTrue{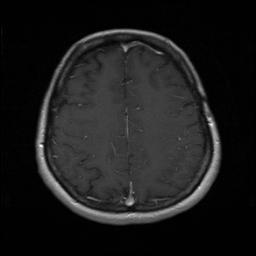} & \imga{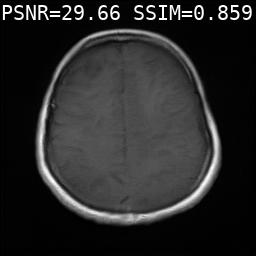} &\imgb{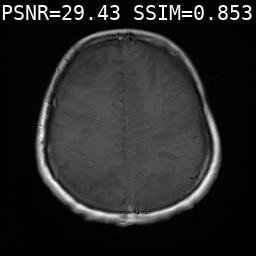}& \imgc{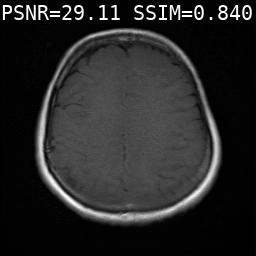} 
        & \imgd{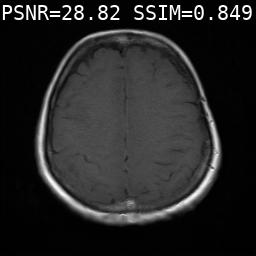} & \imge{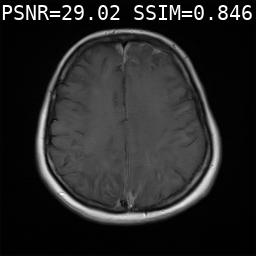} & \imgf{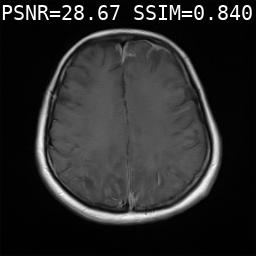} \\
	\imgTrue{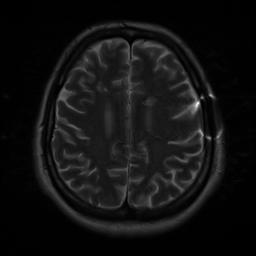} & \imga{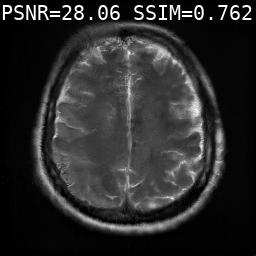} &\imgb{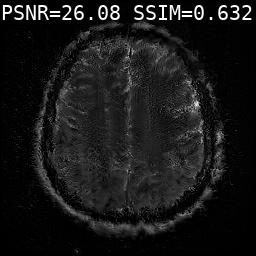}& \imgc{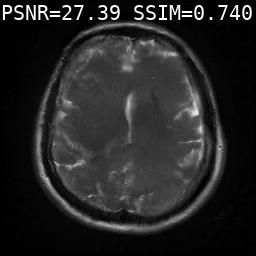} 
        & \imgd{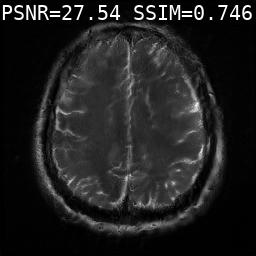} & \imge{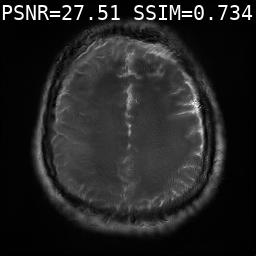} & \imgf{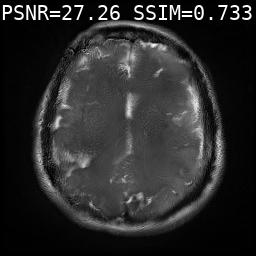} \\
	\imgTrue{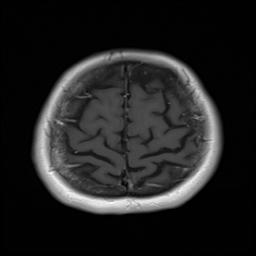} & \imga{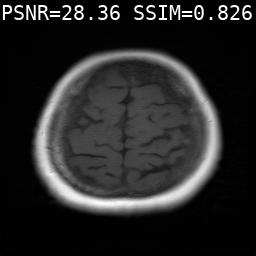} &\imgb{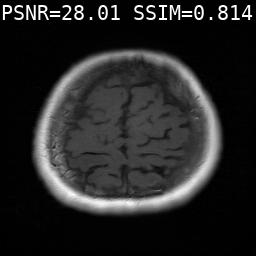}& \imgc{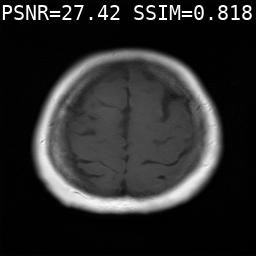} 
        & \imgd{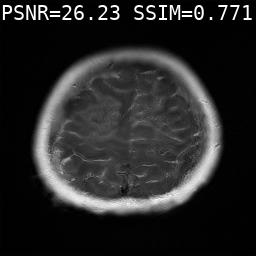} & \imge{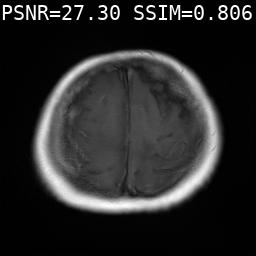} & \imgf{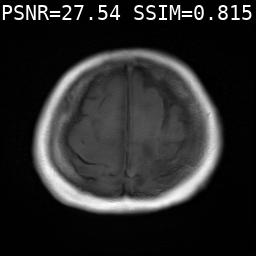} \\
	\end{tabular}
	\caption{Examples of MRI inversion using out-of-distribution images. (Accl=8x, SNR=20 dB). (a) Style vector only + spherical constraint; (b) style vector + noise maps + spherical constraint; (c) style vector only + Gaussianization (only the Yeo-Johnson layer for 1D Gaussianization); (d) style vector + noise maps + Gaussianization (only the Yeo-Johnson layer for 1D); (e) style vector only + Gaussianization (only the Lambert layer for 1D); (f) style vector + noise maps + Gaussianization (only the Lambert layer for 1D).}
	\label{fig:fastmri_dicom}
\end{figure}
\endgroup

First, none of the inversion results shown here are visually comparable to the ground truth, which shows the necessity of having a large training dataset that covers the target distribution. Also, it is vital to ensure that the pre-processing and value ranges of training and target images are the same. Second, in terms of metrics, the results are comparable to but slightly worse than those on synthetic test images reported previously. This is expected because these images are out of distribution. Besides, in cases where we applied Gaussianization layers, using both noise maps and style vectors gives better results than using style vectors only. In addition, Gaussianization layers improved results when we also optimized the noise maps in addition to the style vectors. However, we found that only using the style vector and the spherical constraint gives the best result on this test set, which contradicts our results reported in \tabref{tab:ablation_style_noise}. Our interpretation is that the dimension of style vectors is still relatively lower than noise maps or the latent tensor used in Glow, so our Gaussianization layers may be more effective in dealing with the latter two types of parameters.

\section{Reparameterization schemes for StyleGAN2 and Glow}
\label{append:reparam_schemes}
\begin{figure}[htpb]
	\centering
			\includegraphics[width=\textwidth]{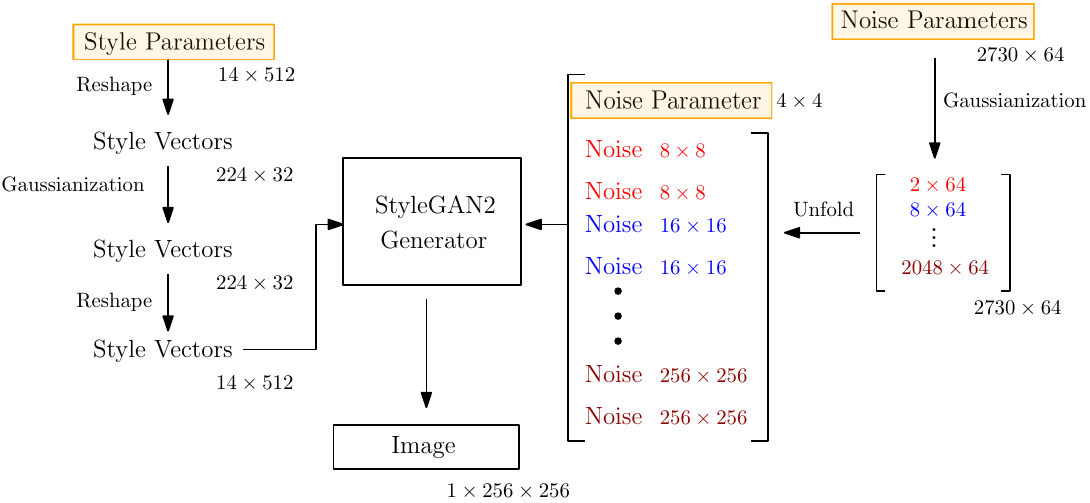}
	\caption{Reparameterization scheme for StyleGAN2. The dimensons of the latent parameters are [the number of vectors \(\times\) vector dimension], except the \(4\times 4\) one. Note that we transpose latent tensors in corresponding equations and algorithms. The style and noise parameters before the Gaussianization layers are \(\{\rvv_{i}\vert {}_{i = 1,\cdots,N}\}\), and the style and noise vectors after the Gaussianization layers are \(\{\rvz_{i}\vert {}_{i = 1,\cdots,N}\}\) mentioned in the main text.}
	\label{fig:style_gan_param}
\end{figure}

\begin{figure}[htpb]
	\centering
			\includegraphics[width=0.8\textwidth]{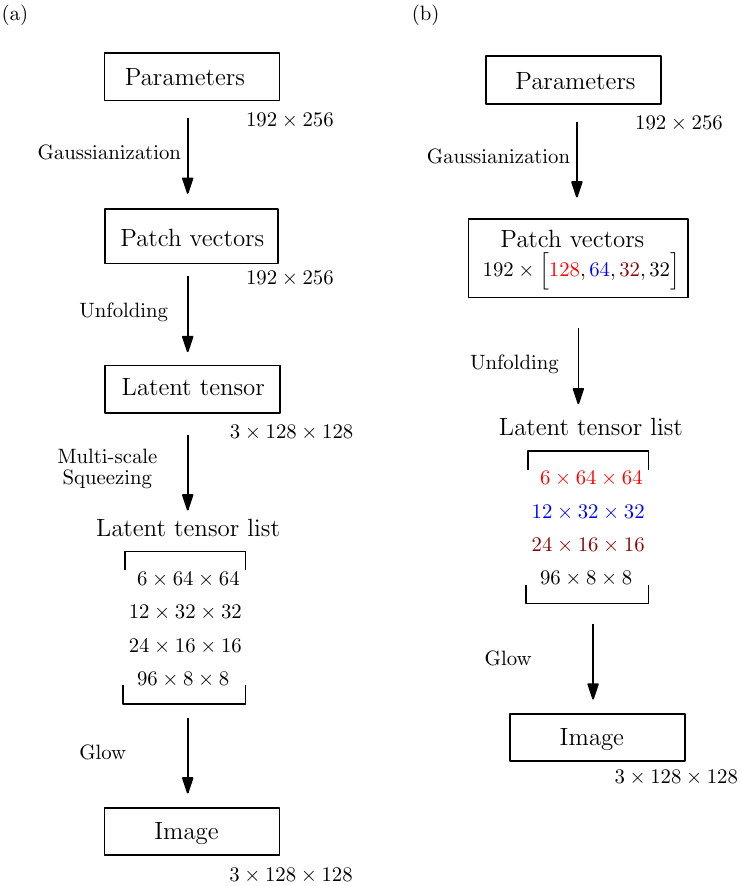}
	\caption{Two reparameterization schemes for Glow. (a) P1; (b) P2. The dimensions of the latent parameters are [vector dimension \(\times\) the number of vectors]. The parameters before the Gaussianization layers are \(\{\rvv_{i}\vert {}_{i = 1,\cdots,N}\}\), and the patch vectors after the Gaussianization layers are \(\{\rvz_{i}\vert {}_{i = 1,\cdots,N}\}\) mentioned in the main text.}
	\label{fig:glow_param}
\end{figure}

\begin{figure}[htpb]
	\centering
			\includegraphics[width=0.8\textwidth]{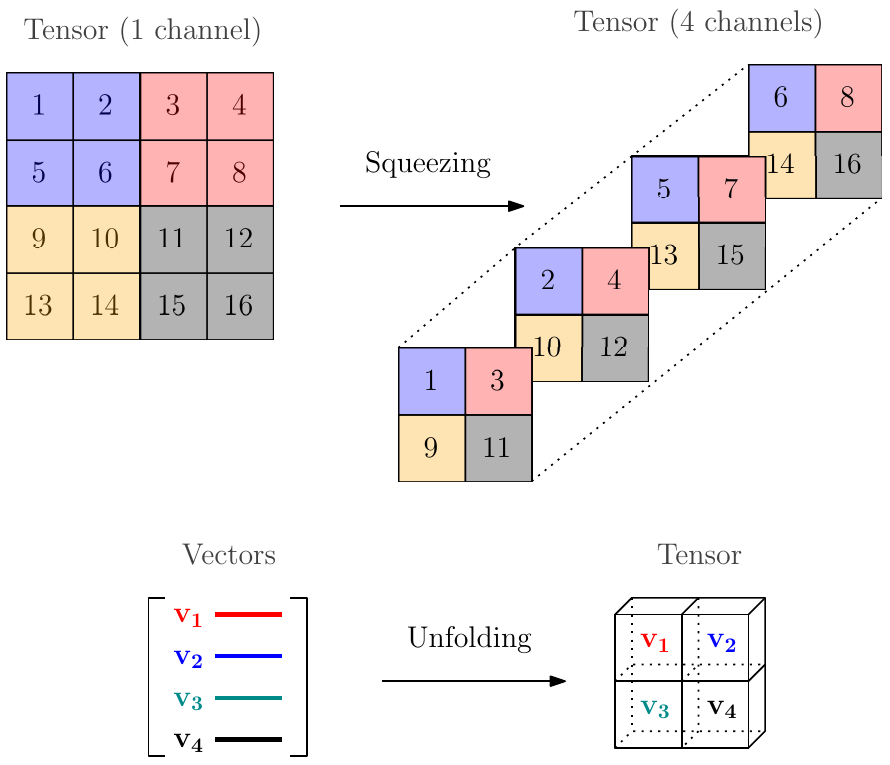}
	\caption{Illustration of the squeezeing and unfolding operations.}
	\label{fig:squeeze_unfold}
\end{figure}
\Figref{fig:style_gan_param} illustrates the reparameterization scheme for StylgeGAN2 using the Gaussianization layers. The patch size for the style vectors and noise maps are 32 and  \(1\times 8\times 8\), respectively. Note that we transpose the latent tensors in \figref{fig:style_gan_param} compared to the notations in corresponding equations and algorithms. Also, as pointed out in \citet{gu2020image}, we also find that a single latent code is insufficient for image reconstruction. Thus, we use multiple style vectors that were fed into different intermediate layers.

For Glow, we came up with two reparameterization schemes (\figref{fig:glow_param}). The first one (P1) reparameterizes patches from a latent tensor with the same dimension as the output image. The dimensions of the patches and the image are \(3\times 8 \times 8\) and \(3 \times 128 \times 128\), respectively. Then a multi-scale squeezing operation maps the latent tensor into a list of tensors corresponding to different scales of Glow. Glow uses the list of tensors as the input. The second scheme (P2) reparameterizes patches extracted directly from tensors in the above list. We also partition each tensor into \(3\times 8 \times 8\) patches. The unfolding and squeezing operations are illustrated in \figref{fig:squeeze_unfold}. In the multi-scale architecture of Glow, the squeezing operation is applied recursively on half of the output tensors cut in the channel direction~\citep{dinh2016density,kingma2018glow}.

\section{More details of the Gaussianization layers}
\label{append:G_layers_details}

\subsection{ICA layer}
\label{append:ICA_details}
The overall ICA layer is summarized in \algref{alg:ica}. We set a maximum number of the fixed-point iterations to reduce computational cost and ensure accurate gradient computation that can pass the finite-difference convergence test. The gradients are backpropagated through the loops without using the trick introduced in \appref{append:opt_layer_backward}.
\paragraph{Whitening} The FastICA algorithm typically requires that the data are pre-whitened. Let \(\mV\) be the data matrix whose columns are data vectors. We first subtract the mean from the data vector using \(\mV \gets \mV - \mathtt{mean}(\mV, \mathtt{dim}=1)\). Then we compute the data covariance matrix using \(\mC = (1 - \eta) \frac{1}{N-1}\mV\mV^\top + \eta \mI\), where we use a small constant (e.g., \(\eta = 0.001\)) to blend the empirical covariance matrix and an identity matrix to avoid ill-conditioning. 

After these data preparation steps, the ZCA whitening used throughout the study first computes the eigenvalues \(\mLambda\) and eigenvectors \(\mD\), and then output the whitened data using \(\mV \gets \mD\mLambda^{-1/2} \mD^\top \mV\).

Alternatively, we can use the following steps to whiten the data, which are also used later in ICA iterations to decorrelate column vectors in the orthogonal matrix~\citep{hyvarinen1999fast}:\\
\vspace{-0.5cm}
\begin{enumerate}
	\item Initialize \( \mW \gets \mI \); 
	\item Compute
	\begin{equation}
		\mW \gets \mW / \sqrt{\Vert \mW{^\top} \mC \mW \Vert_2}\\
	\end{equation}
	\item Repeat until convergence
	\begin{equation}
		\mW \gets \frac{3}{2}\mW - \frac{1}{2} \mW \mW^\top \mC \mW,
	\end{equation}
	\item Output \(\mV \gets \mW^\top \mV\)
\end{enumerate}

\begin{algorithm}[htpb!]
\SetKwComment{Comment}{// }
\SetAlgoLined
\SetKwInOut{Input}{Input}
\SetKwInOut{Output}{Output}
\Input{Data matrix \(\mV \in \mathbb{R}^{D\times N}\); error tolerance \(\epsilon\); damping parameters: \(\eta=10^{-4}\) and \(\alpha=0.8\); maximum iteration numbers \(J=10\) and \(K=100\).}
\Output{Matrix \(\mP \in \mathbb{R}^{D\times N}\) with \iid entries}
\Comment{Whitening stage}
\(\mV = \mV - \mathtt{mean}(\mV, \mathtt{dim}=1)\), \(\mC = (1-\eta)\frac{1}{N-1}\mV\mV^\top + \eta \mI\)\;
\(\mV \gets \text{\texttt{ZCA-whitening}} (\mV) \) \;
\Comment{ICA stage}
\(\mW = \mW^{\ast} \gets \rmI, \quad j\gets 1\) \;
\While{ \(j \leq J \)}{
  $\mW \gets \frac{1}{N} \left[\alpha \mV \phi\left(\mW^\top \mV\right)^\top - \mW \mathtt{diag}\left( \phi'\left( \mW^\top \mV \right) \mathbf{1} \right)\right]$ \eqref{eqn:mat_form_ica} \;
	\( \mW = \mW_0 \gets \mW / \sqrt{\Vert \mW^\top \mW \Vert_2}, \quad k\gets 1 \) \;
	\While{\(k < K\)}{
	\( \mW \gets \frac{3}{2}\mW - \frac{1}{2} \mW \mW^\top \mW \) \;
	\If{\(\|\mW - \mW_0\| < \epsilon\)}{break\;}
	\(\mW_0 \gets \mW, \quad k \gets k+1 \) \;
	}
	\If{\(\|\mW - \mW^{\ast}\| < \epsilon\)}{break\;}
	\(\mW^{\ast} \gets \mW, \quad j \gets j+1 \) \;
	}
\Return \(\mP \gets \mW^\top \mV\).
\caption{ICA Layer}
\label{alg:ica}
\end{algorithm}

\noindent\textbf{The modified FastICA iterations.}
As stated in \citet{hyvarinen1999fixed}, the objective function for one neural unit of the weight vector \(\vw_i\) and input \(\rvv\) is
\begin{equation}
	\argmax_{\vw_i}\, \E\left[ \Phi\left( \vw_i^\top \rvv \right)	 \right], \, \text{s.t., } \E\left[ \left( \vw_i^\top\rvv \right)^2 \right] = 1,
\end{equation}
where \(\Phi\) is the contrast function (\eg, logcosh).
The original derivations convert this constrained optimization to an unconstrained one using Lagrange multipliers. However, this procedure is unnecessary since the matrix \(\mW\) is orthogonalized after each iteration, and the input vectors have been pre-whitened. Therefore, we only need to solve the following equation
\begin{equation}
	\E{\left[\rvv \phi\left( \vw_i^\top \rvv \right)\right]} = 0,
\end{equation}
whose Jacobian is 
\begin{equation}
	\begin{split}
	J &= \E{\left[\rvv\rvv^\top \phi'(\vw_i^\top\rvv)\right]} \\
	&\approx \E{\left[\rvv\rvv^\top\right]} \E{\left[\phi'(\vw_i^\top\rvv)\right]} = \E{\left[\phi'(\vw_i^\top\rvv)\right]},
	\end{split}
\end{equation}
where \(\phi\) is the derivative of \(\Phi\).
The Newton iteration scheme is thus 
\begin{equation}
	\vw_i = \vw_i - \E{\left[\rvv \phi\left( \vw_i^\top \rvv \right)\right]}/\E{\left[\phi'(\vw_i^\top\rvv)\right]}.
\end{equation}
To improve the convergence, we damp the iterations by a parameter \(\alpha \in (0,1)\). Also, using the same technique to convert the Newton iterations to fixed-point iterations in \citet{hyvarinen1999fixed,hyvarinen1999fast}, we arrive at the modified fixed-point iteration scheme:
\begin{equation}
	\vw_i = \alpha \E\left[ \rvv \phi\left(\rvw_i^\top \rvv\right) \right] - \E\left[ \phi'\left(\vw_i^\top \rvv\right)\right] \vw_i,
\end{equation}
followed by the aforementioned decorrelation procedure after each step. The convergence of the modified FastICA iterations can be proved similarly as in \citet{oja2006fastica}. 

\subsection{Power transformation layer}
\begin{algorithm}[htpb]
\SetKwComment{Comment}{// }
\SetAlgoLined
\SetKwInOut{Input}{Input}
\SetKwInOut{Output}{Output}
\Input{Data vector \(\rvp\)}
\Output{Vector \(\rvs\) whose values are 1D Gaussianized.}
Estimate \(\lambda\) from \(\rvp\) using \eqref{eqn:yeo_johnson_ml} \;
Compute \( \rvs \) with the estimated \(\lambda\) and data \(\rvp\) using \eqref{eqn:yeo_johnson_mapping} \;
\Return \(\rvs \).
\caption{Power Transformation Layer}
\label{alg:yeo-johnson}
\end{algorithm}

We propose to use the power transformation or Yeo-Johnson transformation~\citep{yeo2000new} to reduce the skewness of distributions:
\begin{equation}
	s(\lambda, p)=\begin{cases}
	\left( \left( p + 1 \right)^{\lambda} - 1\right)/\lambda, & p\geq 0, \lambda \neq 0,\\
	\log \left( p + 1 \right), & p \geq 0, \lambda = 0, \\
	-(\left[ -p + 1 \right]^{2-\lambda} - 1 )/\left[\left( 2 - \lambda \right)\right], & p < 0, \lambda \neq 2, \\
	-\log(-p + 1), & p < 0, \lambda = 2,
	\end{cases}
\label{eqn:yeo_johnson_mapping}
\end{equation}
where \(p\) is an input value, \(s\) is an output value, and \(\lambda\) is the parameter to be estimated.
As shown in \figref{fig:lambert_yeojohnson_activation}(a), the form of the Yeo-Johnson activation function depends on parameter \(\lambda\). If \(\lambda = 1\), the mapping is an identity mapping. If \(\lambda > 1\), the activation function is convex, compressing the left tail and extending the right tail, reducing the left-skewness. If \(\lambda < 1\), the activation function is concave, which oppositely reduces the right-skewness. The only parameter \(\lambda\) is determined by solving an optimization problem that minimizes the negentropy:
\begin{equation}
\begin{split}
	\lambda &= \argmax_{\lambda} \, l\left( \lambda| \rvp \right) = \argmax_{\lambda} \, -\frac{n}{2} \log \Var\left( s\left(\lambda, p_i \right) \right) + (\lambda - 1) \sum^{n}_{i=1} \sign(p_i) \log(|p_i| + 1),
\end{split}
\label{eqn:yeo_johnson_ml}
\end{equation}
where \(\rvp\) is the input data vector with entries \(p_{i,\,i=1,\cdots,n}\).

We use a custom operator based on SciPy's implementation using Brent's algorithm~\citep{brent2013algorithms} to find an approximate minimum of Problem~\ref{eqn:yeo_johnson_ml}. Continuing from the approximate minimum, we use Brent's root finding algorithm~\citep{brent2013algorithms} to find the minimum where the gradient vanishes.

Since the parameter \(\lambda\) depends on input data, we need to back-propagate the gradient through the optimization process.

The power transformation layer is summarized in \algref{alg:yeo-johnson}.

\subsection{Lambert \(W \times F_X\) layer}
Due to noise and inaccurate forward models, we observe that the distribution of latent vector values tends to be shaped as a heavy-tailed distribution during the inversion process. To reduce the heavy-tailedness, we adopt the Lambert \(W\times F_X\) method detailed in~\citet{goerg2015lambert}. 

Let \(X\) be a random variable whose CDF is \(F_X\), with mean \(\mu_X\) and standard deviation \(\sigma_X\). The following transformation with a heavy-tail parameter \(\delta \geq 0\):
\begin{equation}
	S = \left( U \exp\left( \frac{\delta}{2} U^{2} \right) \right)\sigma_X + \mu_X,
	\label{eqn:lambert_w_fx}
\end{equation}
where \(U = \left( X - \mu_X  \right) / \sigma_X\), is a bijection and maps \(X\) to another random variable \(S\) with heavier tails.

The transformation \eqref{eqn:lambert_w_fx} is bijective if \(\delta \geq 0\), and we can use the Lambert W function to find its inverse.
The Lambert W function \(W\) is defined as the inverse of \(q = W^{-1}\left( t \right) = t\exp{(t)}\), where \(t\) and \(q\) are scalars. Given \(q\), Halley's method can be used to find $t = W(q)$~\citep{corless1996lambertw}. Hence, the inverse of \eqref{eqn:lambert_w_fx} is
\begin{equation}
	X = W_{\delta}\left( \frac{S - \mu_X}{\sigma_X} \right) \sigma_X + \mu_X,
	\label{eqn:inverse_lambert_w_fx}
\end{equation}
where
\begin{equation}
	W_{\delta}\left( u \right) = \sign{(u)} \sqrt{\frac{W\left( \delta u^2 \right)}{\delta}}.
\end{equation}

We use the parameterized Lambert \(W\times F_X\) distribution family to approximate a heavy-tailed input distribution and use \eqref{eqn:inverse_lambert_w_fx} to recover a distribution with lighter tails. In order to make the recovered distribution close to a Gaussian distribution, we compute the optimal heavy-tail parameter \(\delta\) by minimizing the difference between the kurtosis of the output distribution and 3 (Kurtosis is a common surrogate measure of negentropy~\citep{hyvarinen2000independent}): 
\begin{equation}
	\hat{\delta}_{\text{GMM}} = \argmin_{\delta > 0} \Big| \text{Kurt}\left( W_{\delta}\left( \frac{\rvs - \mu_X}{\sigma_X} \right) \right) - 3 \Big|^2,
\label{eqn:lambert_compute_delta}
\end{equation}
where \(\rvs\) is the data vector, and \(\text{Kurt}\) is the kurtosis. We constrain \(\delta > 0\), and solve \eqref{eqn:lambert_compute_delta} using the L-BFGS-B optimizer~\citep{zhu1997algorithm}.

In addition, we estimate the mean \(\mu_X\) and standard deviation \(\sigma_X\) along with \(\delta\) using the Iterative Generalized Method of Moments (IGMM)~\citep{goerg2015lambert}, which embeds an optimization problem for \(\delta\) in an outer loop of iterations to estimate \(\sigma_X\) and \(\mu_X\) (see \algref{alg:igmm}). If the kurtosis of input data vector is not greater than 3, we skip the whole Lambert \(W\times F_X\) layer by directly outputting the data vector.

\begin{algorithm}
\SetAlgoLined
\SetKwInOut{Input}{Input}
\SetKwInOut{Output}{Output}
\Input{Data vector $\rvs$, error tolerance $\epsilon=10^{-5}$, maximum iteration number \(K=100\).}
\Output{vector \(\rvx\), whose elements exhibit an empirical distribution with reduced heavy-tailedness compared to \(\rvs\), and the kurtosis \(\approx 3\).}
Initialize: \( \xi^{(0)} \gets \left( \hat{\mu}^{(0)}_X,  \hat{\sigma}^{(0)}_X,  \hat{\delta}^{(0)} \right) \), $k=0$\;
Compute initial kurtosis \(\beta_2 = \text{Kurt}(\rvs)\) \;
\If{\(\beta_2 \leq 3\)}{return \(\rvs\)\;}
\While{\(k < K\) and \(\|\xi^{(k)} - \xi^{(k-1)}\| \geq \epsilon\)}{
$\rvu^{(k)} \gets \left( \rvs - \mu^{(k)}_X \right) / \sigma^{(k)}_X $\;
Compute \(\delta^{(k+1)}\)\ using \eqref{eqn:lambert_compute_delta};
\( \rvu^{(k+1)} \gets \mW_{\delta^{(k+1)}}\left( \rvu^{(k)} \right) \); \( \rvx^{(k+1)} \gets \rvu^{(k+1)}\sigma^{(k)}_X + \mu^{(k)}_X \) \;
Update \( \mu^{(k+1)}_X \gets \mathtt{mean}\left(\rvx^{(k+1)}\right) \), and \( \sigma^{(k+1)}_X \gets \mathtt{std}\left( \rvx^{(k+1)} \right) \) \;
\( \xi^{(k+1)} \gets \left( \hat{\mu}^{(k+1)}_X,  \hat{\sigma}^{(k+1)}_X,  \hat{\delta}^{(k+1)} \right) \) \;
$k \gets k + 1$ \;
}
\Return \( \rvx = W_{\delta}\left( \frac{\rvs - \mu_X}{\sigma_X} \right) \sigma_X + \mu_X \),
\caption{Lambert \(W\times F_X \) Layer with the Iterative Generalized Method of Moments (IGMM)~\citep{goerg2015lambert}}
\label{alg:igmm}
\end{algorithm}

\section{Details of datasets and training}
\label{append:datasets_training}
For MRI and eikonal tomography, we generate synthetic brain images as inversion targets using the pre-trained StyleGAN2 weights from \citet{kelkar2021prior}, which are trained on data from the databases of fastMRI~\citep{zbontar2018fastmri,knoll2020fastmri}, TCIA-GBM~\citep{scarpace2016radiology}, and OASIS-3~\citep{lamontagne2019oasis}. As a result, there is no sensitive personal information in our target images. In the data generation process, we only used one style vector of a dimension of 512. We picked 100 images that are visually plausible, whose random seeds can be found in our code. These random seeds were never used to initialize inversion. In inversion experiments, we used 14 such style parameters: one for the lowest resolution of \(4\times 4\), one for the tRGB layer, and two for each resolution from \(8\times 8\) to \(256\times 256\), to avoid inversion crime. This choice of expanding latent space dimension is also justified by our observation that only one style parameter vector is generally insufficient for inversion tasks with data from datasets such as CelebA-HQ~\citep{karras2017progressive}.

We used the CelebA-HQ dataset \citep{karras2017progressive} (under the Creative Commons CC BY-NC 4.0 license) for the deblurring experiments. All images were downsampled to the resolution of 128\(\times\)128. We split the 30000 images from CelebA-HQ into the subsets of training (24183 images), validation (2993 images), and testing (2824 images) following the original splits from CelebA~\citep{liu2015faceattributes}. For the inversion tests, we randomly selected 100 images from the test set.

\begin{figure}[htpb]
	\centering
		\setlength{\tabcolsep}{0.1cm}
		\begin{tabular}{l}
			\includegraphics[width=0.45\textwidth]{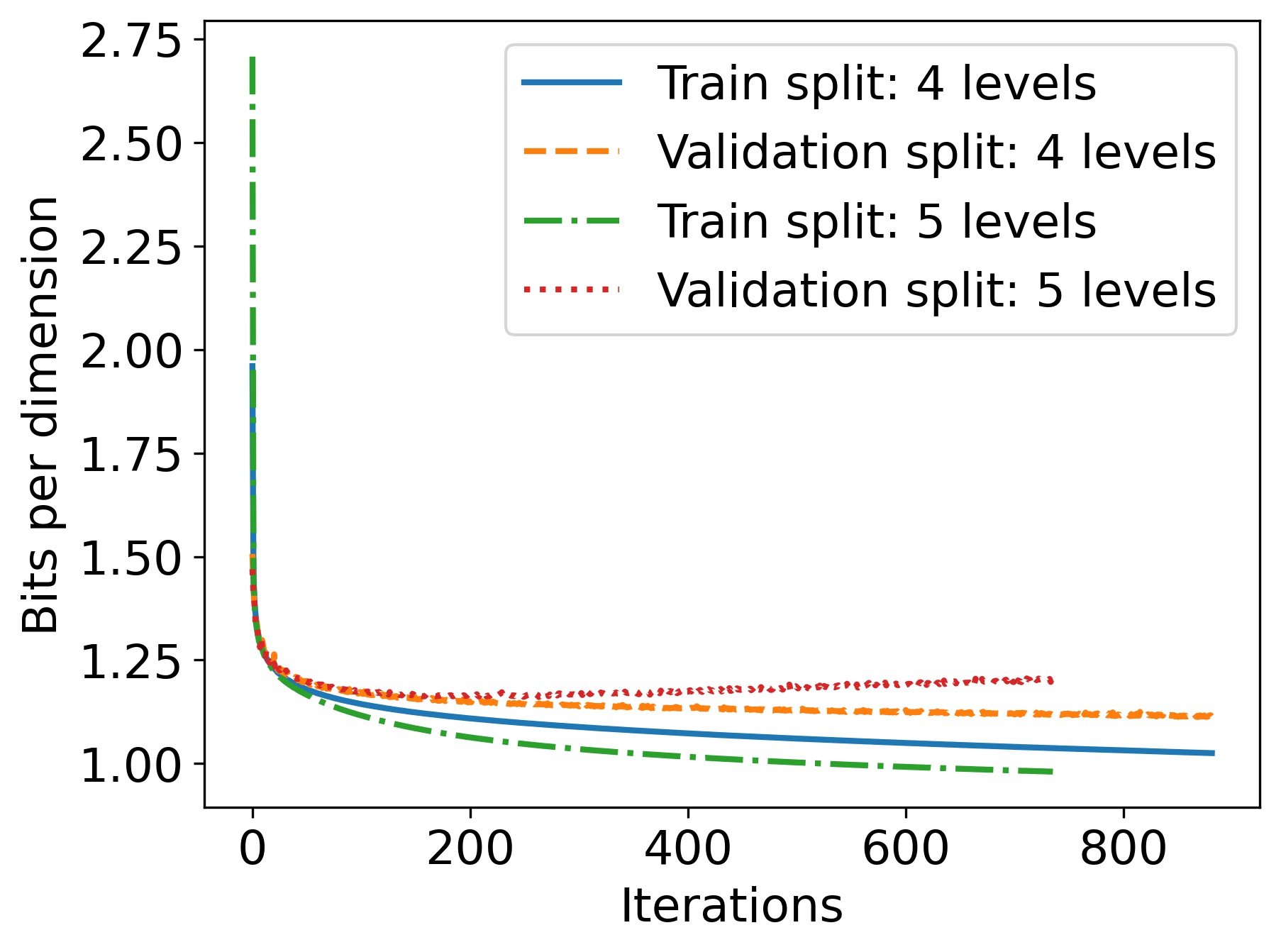} \\
		\end{tabular}
	\caption{The negative log-likelihood or NLL (reported in bits per dimension) on the training and validation splits of the CelebA-HQ dataset with different numbers of multi-scale levels.}
	\label{fig:train_val_losses}
\end{figure}

For the hyper-parameters of the Glow networks, we used 4 multi-scale levels and 32 flow-steps, and we only used additive coupling layers. \Figref{fig:train_val_losses} reports the training process. For each epoch, we computed the training negative log-likelihood (NLL) averaged throughout the epoch, and the validation NLL at the end of the epoch. The validation curves suggest that it is better to use 4 multi-scale levels. We chose the network weights from the epoch before the validation NLL stopped to decrease: 850 for the CelebA-HQ dataset. All training was conducted using \(8\times 32\text{ GB}\) Nvidia V100 GPUs with a batch size of 64. We used the Adam optimizer~\citep{kingma2015adam} with a learning rate of \(10^{-4}\), as well as \(\beta_1 = 0.9\) and \(\beta_2=0.99\).

\section{Gradient computation of the optimization-based differentiable layers}
\label{append:opt_layer_backward}
\begin{figure}[htpb]
	\centering
			\includegraphics[width=0.6\textwidth]{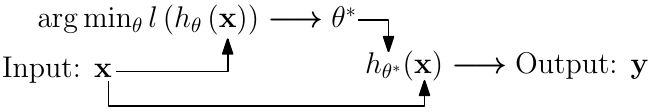}
	\caption{Illustration of the forward computation of optimization-based ICA, Yeo-Johnson, and Lambert $W\times F_X$ layers. We denote the input and output by \(\rvx\) and \(\rvy\), respectively. The layer is represented by \(h_\theta\), and the layer is defined by solving an optimization problem whose objective function is \(l\). The back-propagation of gradients in components defined by \eqref{eqn:yeo_johnson_ml} and \eqref{eqn:lambert_compute_delta} is enabled by the implicit function theorem and automatic differentiation detailed in \appref{append:opt_layer_backward}.}
	\label{fig:opt_layer}
\end{figure}
\begin{figure}[htpb]
	\centering
		\includegraphics[width=0.45\textwidth]{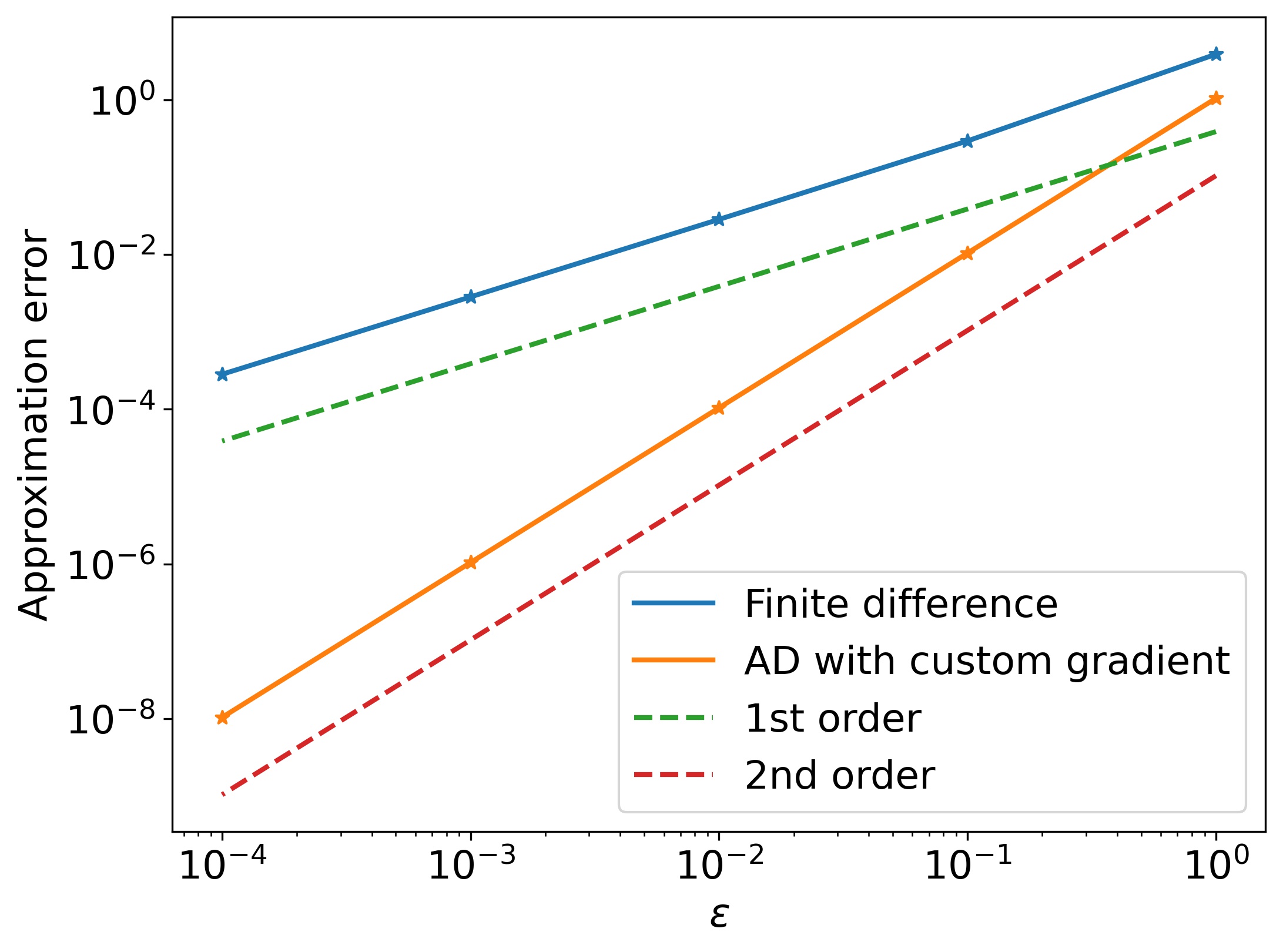}
	\caption{Gradient accuracy test for custom operators. An accurate gradient should make the error converge at a rate of the second order following the red dashed line.}
	\label{fig:grad_check}
\end{figure}

As shown in \figref{fig:opt_layer}, in the power transformation and Lambert \(W\times F_X\) layers, there are operators whose outputs are obtained by solving optimization problems formally described as
\begin{equation}
	\vtheta^\ast = \argmin_{\vtheta} l(\vx, \vtheta),
\end{equation}
where \(l\) denotes the objective function that defines the operator (combining the objective function and layer \(h_{\vtheta}\) in \figref{fig:opt_layer}), symbol \(\vtheta\) stands for the optimal output, a scalar in our cases but can also be a vector in general situations.
The optimal condition is
\begin{equation}
	l_{\vtheta}(\vx, \vtheta) \coloneqq L(\vx, \vtheta) = 0,
	\label{eqn:op_optimal_condition}
\end{equation}
where the subscript denotes partial derivatives.

The optimal condition implicitly defines a forward operator of the following form:
\begin{equation}
	\vtheta^\ast = \mathtt{op}_{\text{forward}} \left(\vx\right).
\end{equation}
The backward operator is
\begin{equation}
	\frac{\partial \chi}{\partial \vx} = \mathtt{op}_{\text{backward}} \left(\frac{\partial \chi}{\partial \vtheta}, \vtheta^\ast, \vx \right),
\end{equation}
where \(\chi\) is the objective function of an inverse problem.

Differentiating \eqref{eqn:op_optimal_condition} with respect to \(\vx\), we have
\begin{equation}
	L_{\vx} + L_{\vtheta}\vtheta_{\vx} =  0 \implies \vtheta_{\vx} = -L_{\vtheta}^{-1}L_{\vx},
\end{equation}
using the implicit function theorem. Then, to back-propagate the gradient from \(\frac{\partial \chi}{\partial \vtheta}\) to \(\frac{\partial \chi}{\partial \vx}\), we use
\begin{equation}
	\frac{\partial \chi}{\partial \vx} = \frac{\partial \chi}{\partial \vtheta} \vtheta_{\vx} = - \frac{\partial \chi}{\partial \vtheta}L^{-1}_{\vtheta}L_{\vx} = - \frac{\partial \chi}{\partial \vtheta}H^{-1}_{\vtheta}L_{\vx},
\end{equation}
where \(H_\vtheta\) is the Hessian matrix of \(\chi\) with respect to \(\vtheta\).

In our problems, the output \(\vtheta\) is a scalar, so it is easy to use automatic differentiation to compute \(L_{\vtheta}\) directly and hence \(L^{-1}_{\vtheta}\). Otherwise, if \(\vtheta\) has many parameters, we can first solve the following linear system with an auxiliary vector \(\bm{\lambda}\):
\begin{equation}
	\bm{\lambda} H_{\vtheta}  = - \frac{\partial \chi}{\partial \vtheta},
\end{equation}
and then compute the gradient using
\begin{equation}
	\frac{\partial \chi}{\partial \vx} = \bm{\lambda} L_{\vx},
\end{equation}
a technique also known as the adjoint-state method. Note that there is no need to compute the Hessian explicitly, but one can use automatic differentiation to compute the vector-Hessian product \(\bm{\lambda} H_{\vtheta}\) and utilize iterative linear solvers like GMRES~\citep{saad1986gmres} to solve the linear system.

As a final note, we check the accuracy of our gradient computation using the finite-difference convergence test based on Taylor expansion:
\begin{equation}
	f(\vx + \epsilon \delta\vx) = f(\vx) + \epsilon\nabla f(\vx)^\top \delta\vx + \mathcal{O}(\epsilon^2),
\end{equation}
where \(\delta\vx\) is a randomly sampled vector with a unit \(\ell_2\) norm, and \(\nabla f(\vx)\) is computed using our custom gradient. We here define \(f\) as a composite function that maps the (vector) output of an forward operator to a scalar, \eg, \(f(\vx) = \Vert \mathtt{op}_{\text{forward}} \left(\vx\right) \Vert^2_2\). Once we decrease \(\epsilon\), we should see that the error term \(f(\vx + \epsilon \delta\vx) - f(\vx) - \epsilon\nabla f(\vx)^\top \delta\vx\) decreases at a rate of the second order. All our layers passed this test, as the example shown in \figref{fig:grad_check}. This test should be conducted in double precision.

\section{Orthogonal reparameterization}
\label{append:ortho_param}
We also propose to reparameterize the latent vector \(\rvz\) using an orthogonal matrix \(\mR\):
\begin{equation}
	\rvz = \mR \rvv, \, \rvv \sim \mathcal{N}(\mathbf{0}, \rmI),
\end{equation}
where \(\rvv\) is fixed during an inversion, and we treat \(\mR\) as the parameter instead. There are various ways to parameterized an orthogonal matrix. We choose the Cayley parameterization:
\begin{equation}
	\mR = (\mI + \mW)(\mI - \mW)^{-1}, \text{ where } \mW = -\mW^\top,
\end{equation}
one of the best reported in \citet{liu2021orthogonal}. Therefore, the DGM-regularized inversion using orthogonal reparameterization is
\begin{equation}
	\argmin_{\mW} ~(1/2) \left\|\rvd - f \circ g \left(\mR \rvv \right)   \right\|^2_2.
	\label{eqn:dgm_inversion_orthogonal}
\end{equation}

The specific reparameterization schemes for StyleGAN2 and Glow are the same as in \appref{append:reparam_schemes}. For style vectors, we use the full dimension of 512 because of the observation described in \tabref{tab:ablation_style_psize}. For Glow and noise maps in StyleGAN2, we use patch sizes \(3\times 8\times 8\) and \(1\times 8\times 8\), respectively, to save computation time and memory.

\section{Gaussian typical set}
\label{append:gaussian_typical_set}
The Gaussian annulus theorem states that a high-dimensional standard Gaussian distribution has most of its probability mass concentrated within an annulus area around a high-dimensional sphere:
\begin{thm}[Gaussian Annulus Theorem~\citep{blum2020foundations}]
For an n-dimensional standard Gaussian, for any \(\beta\leq \sqrt{n}\), all but at most \(3e^{-c\beta^2}\) of the probability mass lies within the annulus \(\sqrt{n} - \beta \leq |x| \leq \sqrt{n} + \beta\), where \(c\) is a fixed positive constant.
\end{thm}
This theorem gives a necessary geometric condition of typical samples from a high-dimensional standard Gaussian. However, the converse is not true: not all vectors whose \(\ell_2\) norm is \(\sqrt{n} - \beta \leq |x| \leq \sqrt{n} + \beta\) are typical examples from the standard Gaussian distribution. As a result, a DGM does not necessarily map a latent vector staying within the Gaussian annulus geometrically to a plausible image, which is demonstrated in \figref{fig:effects_deviate_glow}.

The formal definition of a typical set is as follows.
\begin{mydef}[\citet{cover2012elements}]
\label{def:typical_set}
{Let \(p_X(x)\) be a distribution whose support is \(\mathcal{X}\). The typical set \(A^{(n)}_\epsilon\) is defined as the set of sequences \((x_1, x_2, \cdots, x_n) \in \mathcal{X}^n, \, x_i \sim p_X\), that satisfy
\begin{equation}
	\big\vert H\left[ X \right] + \frac{1}{n} \log p_X(x_1, \cdots, x_n)\big\vert \leq \epsilon,
\end{equation}
where \(H\left[ X \right]\) is the entropy of random variable \(X\).
}
\end{mydef}
Now a random vector \(\rvx \in \mathbb{R}^n \sim \mathcal{N}(\mathbf{0}, \sigma^2\rmI)\) can be factorized as \iid random variables that are distributed as \(\mathcal{N}(0,\sigma^2)\). Therefore, we can regard \(\rvx\) as an \iid sequence and give the following definition:
\begin{mydef}[Gaussian Typical Set]
\label{def:gaussian_typical_set}
	A Gaussian typical set is the typical set \(A^{(n)}_\epsilon\) of \(\rvx \in \mathbb{R}^n \sim \mathcal{N}(\mathbf{0}, \sigma^2\rmI)\).
\end{mydef}
The following theorem guarantees that a typical sample from \(\rvx \in \mathbb{R}^n \sim \mathcal{N}(\mathbf{0}, \sigma^2\rmI)\) resides in the Gaussian typical set with very high probability.
\begin{thm}[\citet{cover2012elements}]
	\label{thm:typical_set_probability}
	For every \(\epsilon > 0\), the typical set has probability \(P\left(A^{(n)}_\epsilon\right) > 1 - \epsilon\) with a sufficiently large dimension \(n\).
\end{thm}
This theorem is a direct application of the asymptotic equipartition property (AEP), which is based on the \iid assumption of sequence entries and the weak law of large numbers. Similar to the Gaussian annulus theorem, Theorem~\ref{thm:typical_set_probability} depicts a geometric property of the Gaussian typical set: it is concentrated in an annulus near a shell of radius \(\sigma\sqrt{n}\), which can be directly verified by definition. Of course, equivalently, if a vector whose \(\ell_2\) norm significantly deviates from \(\sqrt{n}\) cannot be a typical sample from a standard Gaussian. However, one cannot assert that a vector is sampled from the Gaussian typical set by only checking its norm.

\section{Miscellaneous topics}
\label{append:misc_topics}
\begingroup
\setlength{\tabcolsep}{1pt}
\begin{figure}[htpb]
    \centering%
    \newcommand{\sizeI}{0.11}
    \newcommand{\imgTrue}[1]{\includegraphics[width=\sizeI\linewidth]{figures/figures_inv_deblur/gaus_p2/#1/True.jpg}}
    \newcommand{\imgSmooth}[1]{\includegraphics[width=\sizeI\linewidth]{figures/figures_inv_deblur/gaus_p2/#1/Obs.jpg}}
    \newcommand{\imgCSBase}[2]{\includegraphics[width=\sizeI\linewidth]{figures/figures_inv_cs/base_beta#1/#2/Inv.jpg}}
    \newcommand{\imgCSGauss}[1]{\includegraphics[width=\sizeI\linewidth]{figures/figures_inv_cs/gaus/#1/Inv.jpg}}
    \newcommand{\imgDeblurBase}[2]{\includegraphics[width=\sizeI\linewidth]{figures/figures_inv_deblur/base_beta#1/#2/Inv.jpg}}
    \newcommand{\imgDeblurGauss}[1]{\includegraphics[width=\sizeI\linewidth]{figures/figures_inv_deblur/gaus_p1/#1/Inv.jpg}}

    \begin{tabular}{c@{~} c@{~} c@{~} c@{~} c@{~} c@{~} c@{~}}

    \small{Original} & \small{Observed} & \small{\(\beta=0.01\)} & \small{\(\beta=0.1\)} & \small{\(\beta=1.0\)} & \small{\(\beta=10.0\)} & \small{Ours} \\
    \imgTrue{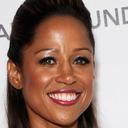} & \imgSmooth{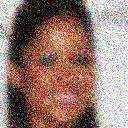} & \imgDeblurBase{0.01}{202591} &\imgDeblurBase{0.1}{202591} & \imgDeblurBase{1.0}{202591} & \imgDeblurBase{10.0}{202591} & \imgDeblurGauss{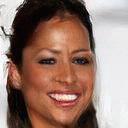} \\

    \end{tabular}

	\caption{Comparison of conventional DGM (Glow)-regularized deblurring results with different weighting factor \(\beta\)s and the result using our Gaussianization layers.}
	\label{fig:figs_intro}
\end{figure}
\endgroup

\subsection{Inversion using Glow}
\label{append:glow_inv_formulation}
In the Glow-regularization framework, we solve the inverse problem by finding the maximum \textit{a posteriori} (MAP) estimate from
\begin{equation}
	p_M(\rvm|\rvd) \propto p \left( \rvd | \rvm \right) p_M(\rvm),
\end{equation}
where \(M\) denotes the parameter space.
The probability density \(p_M\) introduces our \textit{a priori} knowledge and is represented by a normalizing flow (Glow) \(g_{\bm{\theta}}\), which is a differentiable invertible mapping between two distributions, parameterized by neural network parameters \(\vtheta\): \(\rvm = g_{\bm{\theta}}(\rvz)\), where \(\rvz\) is the latent vector.
After training, the log probability density of a given model \(\rvm\) is
\begin{equation}
      \log p_M(\rvm;\bm{\theta}) = \log p_Z\left(g^{-1}_{\bm{\theta}}\left( \rvm \right)\right) + \log \left \vert \det J_{g^{-1}_{\bm{\theta}}} (\rvm) \right \vert 
      = \log p_Z\left( \rvz \right) - \log \big| \det J_{g^{}_{\bm{\theta}}} (\rvz) \big|,
\end{equation}
where \(Z\) stands for the latent space. When we use the trained network in inversion, we freeze the network weights; hence, we drop \(\bm{\theta}\) in \(g_{\bm{\theta}}\) hereafter in our notation. Therefore, \eqref{eqn-augmented_inversion} for Glow-regularized inversion is
\begin{equation}
	   \argmin_{\rvz} ~(1/2)\left\|\rvd - f \circ g\left(\rvz\right) \right\|^2_2 - \beta \left( \log p_Z\left( \rvz \right) - \log \big| \det J_{g} (\rvz) \big| \right).
	   \label{eqn:Glow_regularization_inv}
\end{equation}
The new regularization term in \eqref{eqn-augmented_inversion} is \(\mathcal{R^\prime}(\rvz) = - \beta \left( \log p_Z\left( \rvz \right) - \log \big| \det J_{g} (\rvz) \big| \right)\), and we have shown the results with different \(\beta\)s in \figref{fig:figs_intro} and \tabref{tab:glow_deblur_betas}.

\subsection{Duality of KL divergence}
\label{append:KL_duality}
As also shown in \citet{papamakarios2017masked}, the KL-divergence between two distributions does not change under a differentiable invertible transformation, so
\begin{equation}
	D_{\text{KL}}\left[ p^{\ast}_M( \rvm ) \|  p_M( \rvm; \vtheta ) \right] = D_{\text{KL}}\left[ p^{\ast}_Z( \rvz; \vtheta) \|  p_Z( \rvz ) \right],
\label{eqn:DL_invariant}
\end{equation}
where \(p_M^{\ast}\) is the target distribution in the physical parameter space, and \(p_Z^{\ast}\) is the corresponding latent-space distribution under the normalizing flow. This means that minimizing the forward KL divergence in the \(M\) domain or physical parameter space is equivalent to minimizing the reverse KL-divergence in the \(Z\) domain or the latent space.

This fact and Theorem~\ref{thm:typical_set_probability} imply that a well-trained normalizing flow maps samples from the target distribution into the Gaussian typical set with very high probability and vice versa.

\subsection{Invariance of KL-divergence and multi-information}
\label{append:KL-MI-proofs}
For the completeness of this paper, we briefly prove the key properties of KL-divergence and multi-information used in the Gaussianization framework.

First, we prove that the KL-divergence is invariant under differentiable bijections. We write the definition of the KL-divergence between distribution \(p_\rvx\) and \(q_\rvx\) of random vector \(\rvx\) as
\begin{equation}
	\KL(p_\rvx \| q_\rvx) = \int p_\rvx(\vx) \log \frac{p_\rvx(\vx)}{q_\rvx(\vx)} \mathrm{d}\vx.
\end{equation}
Suppose there is a differentiable and invertible transformation \(T\) that maps \(\rvx\) to \(\rvu\): \(\rvu = T(\rvx)\). Then, we know the PDF under change of variable is
\begin{equation}
	p_\rvx(\vx) = p_\rvu(\vu)|\det J_T(\vx)|, \quad q_\rvx(\vx) = q_\rvu(\vu)|\det J_T(\vx)|,
\end{equation}
where we define the Jacobian matrix \(J_T(\vx) = \frac{\partial \rvu}{\partial \rvx} (\vx)\). Therefore, we have
\begin{eqnarray}
	\KL(p_\rvx \| q_\rvx) &=& \int p_\rvx(\vx) \log \frac{p_\rvx(\vx)}{q_\rvx(\vx)} \mathrm{d}\vx \\ \nonumber
	&=& \int p_\rvu(\vu)   |\det J_T(\vx)| \log \frac{p_\rvu(\vu)|\det J_T(\vx)|}{q_\rvu(\vu)|\det J_T(\vx)|} \mathrm{d}\vx \\
	&=& \KL(p_\rvu \| q_\rvu) \nonumber
\end{eqnarray}

Then, we show that the multi-information does not change under component-wise invertible differentiable transformations. The multi-information is defined by the KL-divergence between the joint distribution \(p(\rvx)\) and the marginals:
\begin{equation}
	I(\rvx) = I(\rx_1, \cdots, \rx_D) = \KL \left( p\left( \rvx \right) \Big\Vert \prod_j^D p^j_\rx(\rx_j) \right).
\end{equation}
We define the bijection \(\rvu = T(\rvx)\) specifically with invertible differentiable transformations for each component \(T_i(\rx_i) = \ru_i\). Then, using the KL-divergence invariance we just proved and that \(\prod_j^D p^j_\rx(\rx_j) = \prod_j^D p^j_\ru(\ru_j) |\det J_T(\vx)|\), we get \(I(\rvx) = I(\rvu)\).

\subsection{The rolling operation}
\label{append:rolling}
After applying one set of the Gaussianization layers to the latent tensor, it might be desirable to apply the layers to a different set of non-overlapping patches from the latent tensor again. The rolling operation shifts the latent tensor in the horizontal and vertical directions by half of the patch size \(w\) before patch extractions. The values at the boundaries are wrapped around to the opposite sides. One can, for example, use the \texttt{torch.roll} command to implement this functionality.

\section{Additional discussions} 
\label{append:additional_discussions}
\begingroup
\setlength{\tabcolsep}{1pt}
\begin{figure}[htpb]
    \centering%
    \newcommand{\sizeI}{0.1}
    \newcommand{\imgTrue}[1]{\includegraphics[width=\sizeI\linewidth]{figures/figures_inv_deblur/gaus_p2/#1/True.jpg}}
    \newcommand{\imgSmooth}[1]{\includegraphics[width=\sizeI\linewidth]{figures/figures_inv_deblur/gaus_p2/#1/Obs.jpg}}
    \newcommand{\imgOff}[1]{\includegraphics[width=\sizeI\linewidth]{figures/figures_inv_deblur/base_beta1.0/#1/Inv.jpg}}
    \newcommand{\imgOn}[1]{\includegraphics[width=\sizeI\linewidth]{figures/figures_inv_deblur/gaus_p2/#1/Inv.jpg}}
	\begin{tabular}{c c@{~~}|@{~~} c c}
	\small{Original} & \small{Observed} & \small{\(\beta=1.0\)} & \small{G layers} \\
	\imgTrue{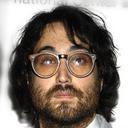} & \imgSmooth{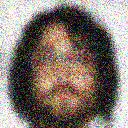} & \imgOff{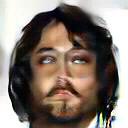} &\imgOn{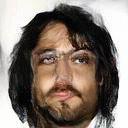} \\
	\end{tabular}
	\caption{A failure example where inversion is unable to restore the eyeglasses because of the strong constraint from traversing within the Gaussian typical set. The results are dominated by the bias of the training dataset.}
	\label{fig:comparison_celebA_large_biased}
\end{figure}
\endgroup
We point out the limitations and potential improvements of this work. First, requiring latent vectors to traverse within the Gaussian distribution/typical set means that the statistics of the training dataset will dominate the results. \Figref{fig:comparison_celebA_large_biased} shows that our method cannot restore the eyeglasses because it is not a significant feature within our training data. One needs to pay particular attention to training dataset construction to ensure that it represents the typical features to be investigated. One should also interpret their results with this caveat in mind. On the other hand, one may combine Gaussianization layers and latent space manipulation techniques~\citep{shen2020interpreting} for a guided inversion, \eg, requiring that the inverted images should have eyeglasses. Second, we only use one set of Gaussianization layers. One can further improve the performance by using additional sets of layers, with possible different partition schemes, such as using the rolling operation (\appref{append:rolling}) to shift the patch extraction locations. Third, since the deep generative models are highly nonlinear, the results may get stuck in local minima. We mitigate this problem by starting inversion from multiple randomly initialized latent vectors. However, this technique is computationally expensive. It is an important research direction to improve the initialization process.

The Gaussianization layers are based on optimization and fixed-point iterations, which can be computationally expensive. To study possible simplifications, we estimated the forward and backward time for various combinations of sub-layers using 100 repeated experiments (\tabref{tab:comp_time}). The computation was performed using one Nvidia V100 GPU and an Intel Xeon Platinum 8276 CPU. We always turn on the ICA layer (together with whitening) since the ablation studies show that it contributes the most to performance increase (\tabref{tab:ablation_mri} and \tabref{tab:ablation_glow_g_layers}). Combining results from those ablation studies, we can see that the option ICA+YJ strikes a good balance between performance and computation time. It has a shorter runtime than ICA+Lambert and ICA+YJ+Lambert. At the same time, it gives comparable and sometimes the best results. On the other hand, the additional computation time from Gaussianization layers is well justified by the performance increase, especially if the physics simulation is much more time-consuming. For example, the eikonal solver for the experimental setup in our study has a forward runtime of 0.293$\pm$0.013 seconds and a backward runtime of 5.933$\pm$0.151 seconds, respectively. In addition, our implementation can also be greatly improved by more efficient utilization of GPUs and rewriting bottleneck components using high-performance codes.  
\begin{table}[htpb]
\setlength{\extrarowheight}{2pt}
\caption{Forward and backward computation time of various combinations of Gaussianization sublayers. The times were computed using tensors of the same shape as latent tensors for Glow and StyleGAN2, respectively. Each computation time was estimated by 100 repeated experiments. We always turn on the whitening layer and the standardization layer.}
\label{tab:comp_time}
\vspace{0.2em}
\centering
\small
\begin{tabular}{l@{~~~}c@{~~~}c@{~~~}c@{~~~}c@{~~~}}
 \toprule
   (Note: the time is measured in seconds) & ICA    & ICA + YJ & ICA + Lambert & ICA + YJ + Lambert \\
 \hline
 Forward (Glow tensor)  & 0.030\tiny{$\pm$0.038} & 0.286\tiny{$\pm$0.052} & 1.911\tiny{$\pm$0.338} 	& 2.022\tiny{$\pm$0.304} \\
 Backward (Glow tensor) & 0.017\tiny{$\pm$3.8e-4} & 0.022\tiny{$\pm$3.3e-4} & 0.751\tiny{$\pm$0.110} & 0.693\tiny{$\pm$0.123}  \\
 Forward (StyleGAN2 tensor) & 0.033\tiny{$\pm$3.1e-4} & 0.280\tiny{$\pm$0.047} & 1.068\tiny{$\pm$0.141} & 1.212\tiny{$\pm$0.097} \\
 Backward (StyleGAN2 tensor) & 0.027\tiny{$\pm$2.6e-4} & 0.037\tiny{$\pm$5.4e-4} & 1.077\tiny{$\pm$0.100} & 0.964\tiny{$\pm$0.065}  \\

\bottomrule
\end{tabular}
\end{table}

We also provide insights into why Gaussianization layers perform better than other methods. First, the spherical constraint only relies on the geometric property of the Gaussian typical set (\appref{append:gaussian_typical_set}). However, as shown in \figref{fig:effects_deviate_glow}, it is insufficient to only constrain the norm of latent vectors in inversion. Gaussianization is necessary to ensure inverted results are in the range of DGMs. Second, the orthogonal reparameterization generally underperforms the Gaussianization layers. Our interpretation is as follows: rather than actively destroying latent tensor patterns like the Gaussianization layers do, orthogonal reparameterization cannot prevent permuting the fixed typical Gaussian latent tensors into ones with spatial patterns. In addition, we have shown in \figref{fig:sg2_style_effects} that the whitening transformation alone is ineffective in keeping the DGM outputs within the range. We thus interpret that reducing higher-order dependencies is essential. Therefore, the noise regularization and the whitening layers yield less satisfying results than the Gaussianization layers.

We believe this study can benefit both the broader scientific community and the general public. However, we point out that the MRI and eikonal tomography experiments are purely synthetic and numerical. They do not fully reflect realistic medical imaging configurations. One should be cautious about applying our techniques to real data and interpreting the results.

\end{document}